\documentclass[lettersize,journal]{IEEEtran}
\usepackage{amsmath,amsfonts}
\usepackage{array}
\usepackage[caption=false,font=normalsize,labelfont=sf,textfont=sf,font=footnotesize]{subfig}
\usepackage{textcomp}
\usepackage{stfloats}
\usepackage{url}
\usepackage{verbatim}
\usepackage{graphicx}
\def\BibTeX{{\rm B\kern-.05em{\sc i\kern-.025em b}\kern-.08em
    T\kern-.1667em\lower.7ex\hbox{E}\kern-.125emX}}

\usepackage{graphics} % for pdf, bitmapped graphics files
\usepackage{epsfig} % for postscript graphics files
\usepackage{balance} % balance the text of the last page
\usepackage{times} % assumes new font selection scheme installed

\usepackage{amssymb} % assumes amsmath package installed
\usepackage{bm}
\DeclareMathAlphabet{\mathpzc}{OT1}{pzc}{m}{it}

\usepackage{amsthm}
\usepackage[hidelinks]{hyperref} 
\usepackage[capitalise,nameinlink]{cleveref}
\usepackage{color}
\usepackage{tabularx}
\usepackage{siunitx}
\usepackage[hang]{footmisc}
\usepackage{float}

\usepackage{algorithm}
\usepackage[noend]{algorithmic}

\usepackage{xspace}
\newcommand{\eg}{\hbox{\emph{e.g.}}\xspace}
\newcommand{\ie}{\hbox{\emph{i.e.}}\xspace}
\newcommand{\etc}{\hbox{\emph{etc.}}\xspace}

\newtheoremstyle{mystyle}%                % Name
  {}%                                     % Space above
  {}%                                     % Space below
  {\itshape}%                                     % Body font
  {}%                                     % Indent amount
  {\bfseries}%                            % Theorem head font
  {.}%                                    % Punctuation after theorem head
  { }%                                    % Space after theorem head, ' ', or \newline
  {\thmname{#1}\thmnumber{ #2}\thmnote{ (#3)}}%                                     % Theorem head spec (can be left empty, meaning `normal')

\theoremstyle{mystyle}
\newtheorem{assumption}{Assumption}

\newtheorem{remark}{Remark}
\newtheorem{hypothesis}{Hypothesis}

\usepackage{xcolor}
\usepackage[export]{adjustbox}

\usepackage{tikz}
\usetikzlibrary{calc}
\ifdefined\figexternalized
    \usetikzlibrary{external}
\fi

\begin{document}
\title{Impact-Aware Bimanual Catching of Large-Momentum Objects}
\author{Lei Yan$^{*,1}$, Theodoros Stouraitis$^{*,2}$, Jo\~{a}o Moura$^{3}$, Wenfu Xu$^{1}$, Michael Gienger$^{2}$, and Sethu Vijayakumar$^{3}$
\thanks{$^{*}$Denotes equal contribution for both authors.} 
\thanks{
This work is supported by the National Natural Science Foundation of China (Grant No. 62203140), Shenzhen Excellent Scientific and Technological Innovation Talent Training Project (Grant No. RCBS20221008093122054 and RCJC20200714114436040), the EPSRC UK RAI Hub in Future AI and Robotics for Space (FAIR-SPACE, EP/R026092/1), The Alan Turing Institute, and the EU H2020 project Enhancing Healthcare with Assistive Robotic Mobile Manipulation (HARMONY, 101017008).
\textit{(Corresponding author: Wenfu Xu.)}
}
%\thanks{This work is supported by the National Natural Science Foundation of China (Grant No. 62203140), Shenzhen Excellent Scientific and Technological Innovation Talent Training Project (Grant No. RCBS20221008093122054 and RCJC20200714114436040 ). This work is also supported by EPSRC U.K. RAI Hub in Future AI and Robotics for Space (FAIR-SPACE: EP/R026092/1). \textit{(Corresponding author: Wenfu Xu.)}}
\thanks{$^{1}$Lei Yan, Wenfu Xu are with the School of Mechanical Engineering and Automation, Harbin Institute of Technology, Shenzhen, China, and also with Guangdong Provincial Key Laboratory of Intelligent Morphing Mechanisms Adaptive Robots, Key University Laboratory of Mechanism \& Machine Theory and Intelligent Unmanned Systems of Guangdong (e-mail: lei.yan@hit.edu.cn,  wfxu@hit.edu.cn).}
\thanks{$^{2}$Theodoros Stouraitis, Michael Gienger are with Honda Research Institute Europe (HRI-EU), Germany (e-mail: theostou@honda-ri.de, michael.gienger@honda-ri.de).}
\thanks{$^{3}$Jo\~{a}o Moura, Sethu Vijayakumar are with the School of Informatics, University of Edinburgh, Edinburgh, UK (e-mail: jpousad@ed.ac.uk, sethu.vijayakumar@ed.ac.uk).}
}

\markboth{IEEE TRANSACTIONS ON ROBOTICS,~Vol.~XX, No.~XX, March~2024}
{How to Use the IEEEtran \LaTeX \ Templates}
\maketitle

\begin{abstract}
This paper investigates one of the most challenging tasks in dynamic manipulation---catching large-momentum moving objects. 
Beyond the realm of quasi-static manipulation, dealing with highly dynamic objects can significantly improve the robot's capability of interacting with its surrounding environment. Yet, the inevitable motion mismatch between the fast moving object and the approaching robot will result in large impulsive forces, which lead to the unstable contacts and irreversible damage to both the object and the robot. To address the above problems, we propose an online optimization framework to: 1) estimate and predict the linear and angular motion of the object; 2) search and select the optimal contact locations across every surface of the object to mitigate impact through sequential quadratic programming (SQP); 3) simultaneously optimize the end-effector motion, stiffness, and contact force for both robots using multi-mode trajectory optimization (MMTO); and 4) realise the impact-aware catching motion on the compliant robotic system based on indirect force controller. We validate the impulse distribution, contact selection, and impact-aware MMTO algorithms in simulation and demonstrate the benefits of the 
proposed framework in real-world experiments including catching large-momentum moving objects with well-defined motion, constrained motion and free-flying motion.

\end{abstract}


% %===============================================================================
\section{Introduction}\label{sec:intro}

\begin{figure}[t]
\centering
	\def\svgwidth{0.94\linewidth}
        %% Creator: Inkscape 1.2.1 (9c6d41e410, 2022-07-14), www.inkscape.org
%% Accompanies image file 'illustration.pdf' (pdf, eps, ps)
\begingroup%
  \makeatletter%
  \providecommand\color[2][]{%
    \errmessage{(Inkscape) Color is used for the text in Inkscape, but the package 'color.sty' is not loaded}%
    \renewcommand\color[2][]{}%
  }%
  \providecommand\transparent[1]{%
    \errmessage{(Inkscape) Transparency is used (non-zero) for the text in Inkscape, but the package 'transparent.sty' is not loaded}%
    \renewcommand\transparent[1]{}%
  }%
  \providecommand\rotatebox[2]{#2}%
  \newcommand*\fsize{\dimexpr\f@size pt\relax}%
  \newcommand*\lineheight[1]{\fontsize{\fsize}{#1\fsize}\selectfont}%
  \ifx\svgwidth\undefined%
    \setlength{\unitlength}{788.38311768bp}%
    \ifx\svgscale\undefined%
      \relax%
    \else%
      \setlength{\unitlength}{\unitlength * \real{\svgscale}}%
    \fi%
  \else%
    \setlength{\unitlength}{\svgwidth}%
  \fi%
  \global\let\svgwidth\undefined%
  \global\let\svgscale\undefined%
  \makeatother%
  \begin{picture}(1,0.68420404)%
    \lineheight{1}%
    \setlength\tabcolsep{0pt}%
    \put(0,0){\includegraphics[width=\unitlength,page=1]{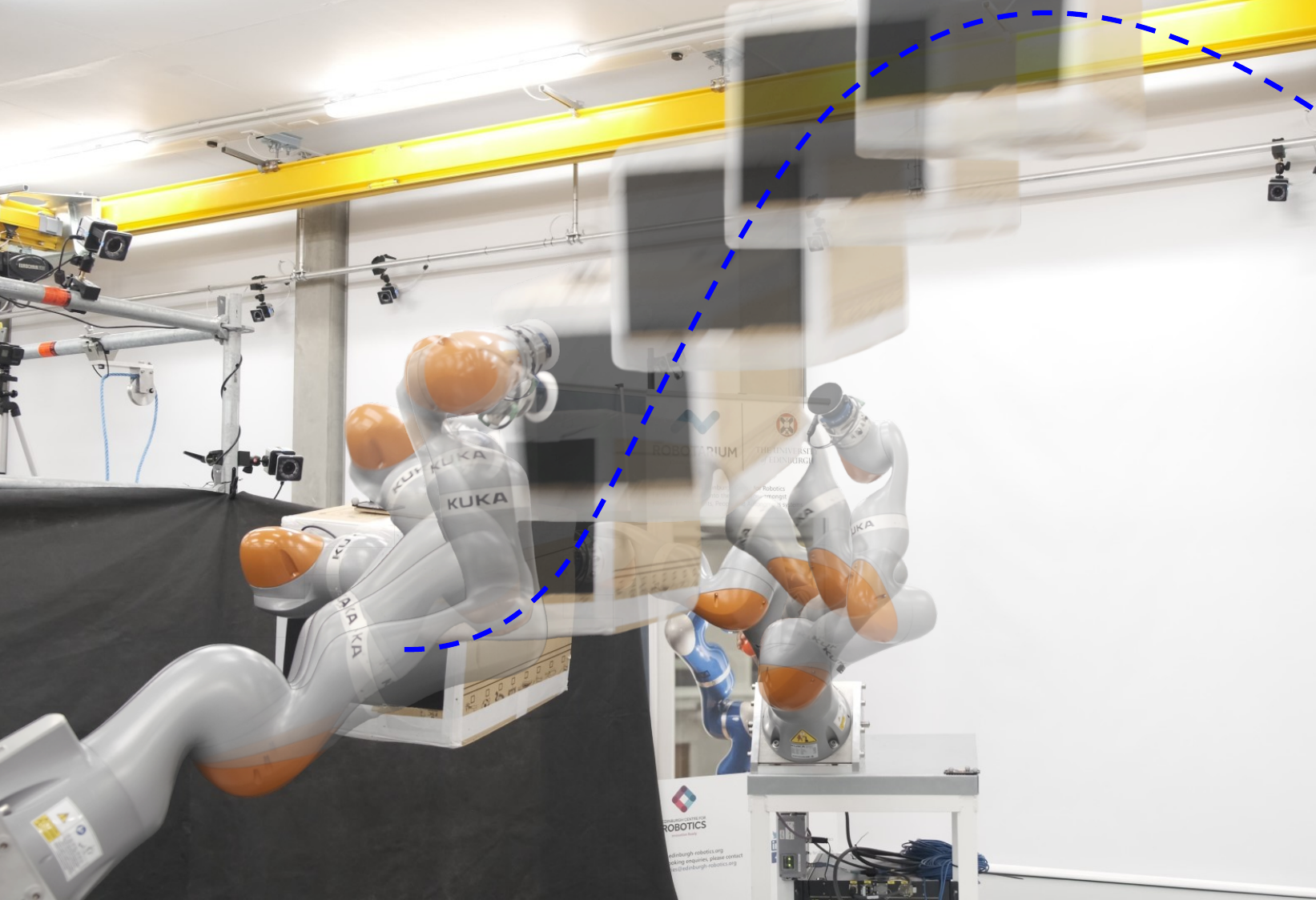}}%
    \put(0.76672967,0.49563547){\makebox(0,0)[lt]{\lineheight{1.25}\smash{\begin{tabular}[t]{l}Predicted \\Trajectory\end{tabular}}}}%
    \put(0,0){\includegraphics[width=\unitlength,page=2]{figures/illustration1.pdf}}%
    \put(0.76927253,0.2892003){\makebox(0,0)[lt]{\lineheight{1.25}\smash{\begin{tabular}[t]{l}Catching\\ Trajectory\end{tabular}}}}%
  \end{picture}%
\endgroup%

        \vspace{-2mm}
	\caption{Two KUKA-iiwa robots catching a flying large-momentum box that weighs~$\SI{4.2}{\kilogram}$ and travels with speed larger than~\SI{3.5}{\meter\per\second}. } 
	\label{fig:first_figure}
	\vspace{-5mm}
\end{figure}

\IEEEPARstart{I}{mpacts} are inevitable when performing efficient contact-rich motions such as running, kicking a ball, catching and tossing objects. They are also inherent to highly dynamic dexterous manipulation~\cite{dafle2014extrinsic} that goes beyond the realm of quasi-static problems~\cite{rodriguez2021unstable}.
Enabling robots with extra manipulation capabilities from prehensile throwing~\cite{braun2013robots} and tossing~\cite{zeng2020tossingbot, jongeneel2022model} to non-prehensile pushing~\cite{moura2022non} %~\cite{sleiman2019contact} 
and batting~\cite{jia2019batting} can boost task efficiency. Yet, when considering dynamic robot manipulation capabilities, such as prehensile
intercepting~\cite{kim2014catching} and soft catching~\cite{salehian2016dynamical}, for heavy and large objects (see~\cref{fig:first_figure}), \textcolor{black}{
% contact %/impact  modelling 
non-prehensile manipulation and impacts handling are essential---as it is not possible to cage those large objects, due to their size, momentum, and the fact that contacts made at speed can break or slide at any point and time.}

\vspace{0mm}

\subsection{The challenging problems}
\begin{figure}[t]
\centering
	\def\svgwidth{1.065\linewidth}
        \input{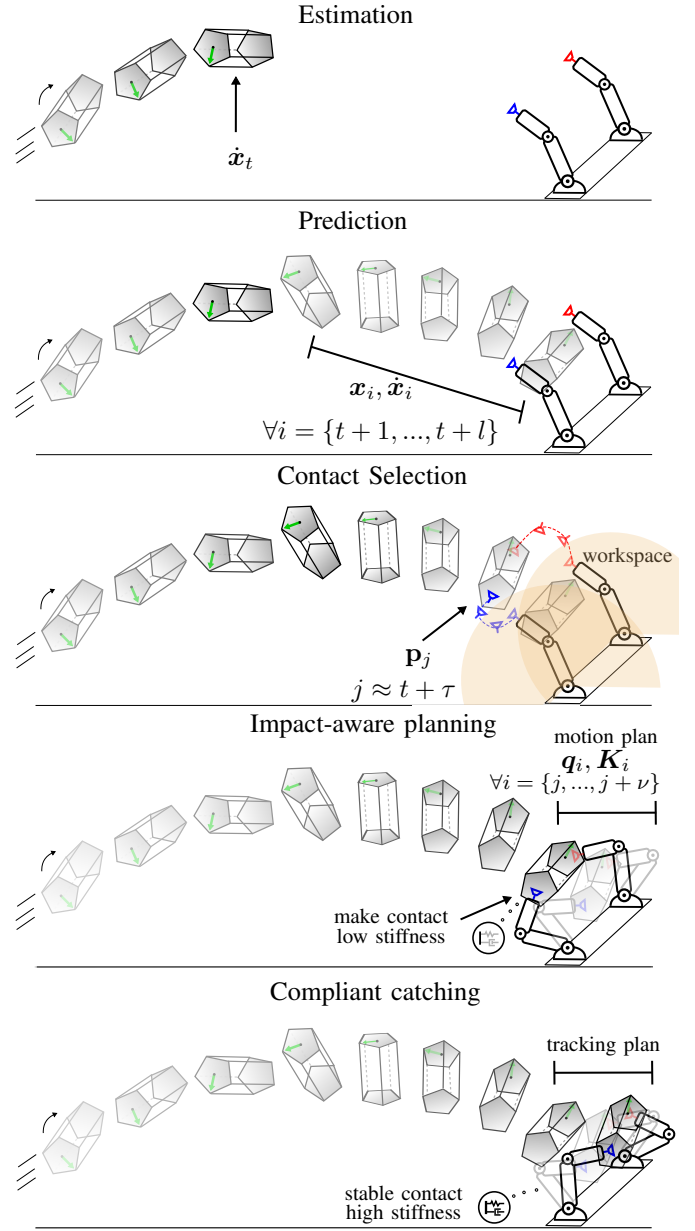}
        \vspace{-2mm}
	\caption{Step-by-step pictorial description of catching a tumbling-flying object.} 
	\label{fig:scenario}
	\vspace{-5mm}
\end{figure}

% \paraDraft{High-level idea of the studied problem}
\textcolor{black}{This work focuses on impact-aware dynamic manipulation, and particularly the challenging problem of coping with impacts to catch large, heavy, fast and tumbling objects in motion with a dual-arm robotic system.}
This goes beyond the state of art on dynamic object manipulation, where either light flying objects were caught  with a single arm \textcolor{black}{(negligible impulses)}~\cite{jia2019batting, salehian2016dynamical, lampariello2011trajectory} or heavy quasi-static objects were caught with a dual-arm robot \textcolor{black}{(negligible impulses)}~\cite{yan2016coordinated, yan2018dual,bombile2022dual}.  
For our problem, the objects travel with arbitrary orientation along projectile trajectories with a duration of less than one second \textcolor{black}{and contact making is prone to impacts}.
% Moreover, the large dimensions of the object in comparison to  the robots makes it impractical to simply intercept it~\cite{kim2014catching, bauml2010kinematically}---requiring the robots to make contacts by following the object motion for at least a short duration of time. 
\textcolor{black}{For these tasks, simply intercepting the object~\cite{kim2014catching, bauml2010kinematically} is impractical, as the object cannot be caged and it can bounce off. Thus, for the robots to make stable contact with the object they need to follow its motion for at least a short duration of time. }
Furthermore, this following motion needs to coincide with the short time window where the trajectory of the large object  and the dual-arm workspace intersect, as shown in~\cref{fig:scenario}.
% while the object trajectory just shortly intersects the dual-arm workspace, 
%The above conditions require the robots to make contacts by following the object motion---not simply intercepting it~\cite{kim2014catching, bauml2011catching}---for at least a short duration of time, while the object trajectory intersects the dual-arm workspace only briefly, as shown in~\cref{fig:scenario}.

% \paraDraft{Challenges of the task}
Catching large fast-moving objects is extremely challenging, its success depends on many factors and requires the solution of the following three main challenging problems.

\textit{\textbf{C1)}}~\textit{Estimating the object motion and predicting its future trajectory for $\approx$1 sec.}
    This translates into:
    (i) using a small fragment of position and orientation data of the object to estimate its momentary linear and angular velocities, and
    \textcolor{black}{(ii) using these estimates} to predict the trajectory of a moving object with nonlinear rotation.
    Both the estimation and the prediction computations need to happen in a fraction ($<$1/10 sec) of the object's total flying motion.  
    
    To address (i) we utilize prior work~\cite{jia2019batting, Prevost2007} and to address (ii) we solve an Initial Value Problem~\cite{lampariello2021optimal}, which predicts the motion of the object by integrating its forward dynamics model in time~\cite{kelly2017introduction}, while considering environment constraints.

\textit{\textbf{C2)}}~\textit{Planing impact-aware dual-arm motion to catch the moving object.} This involves the following considerations. (I) Establishing geometrically feasible and suitable contacts between the dual-arm robot and the moving, spinning object. This is far from trivial, as it depends on both  geometry and motion. Hence, our first research question is (Q1) \textit{how to optimally select contacts on the moving object?} \\
(II) Obtaining low impulsive forces while making contact at speed~\cite{yang2021impact} to avoid breaking the recently established contact~\cite{wang2022predicting}. We study the influence of contact selection on the magnitude of the impulsive forces, which leads to the following key questions: (Q2) \textit{how to minimize the impulsive force through contact selection?} and (Q3) \textit{how to analyze the distribution of impulse along different directions?}\\
(III) Coordinating motion and stiffness variation of the two arms with the object motion while respecting the limited workspace. %of a dual-arm robot. 
Intuitively \textcolor{black}{to minimize impulsive forces}, 
we aim to minimize
the relative contact velocity and be ``\textit{soft}". Hence, we can frame the scientific question as: (Q4) \textit{what is an appropriate model that captures the relation between stiffness, motion and impulsive force?}, along with the more practical question: (Q5) \textit{how to jointly optimize the motion, stiffness and force of a robot to mitigate impact, while taking into account environment and system constraints, such as workspace limits?}

\textcolor{black}{\textit{\textbf{C3)}}~\textit{Tracking motions and making contacts through impact in spite of estimation and execution errors.} Making contact with a large-momentum object at non-zero speed has a direct influence on the continuity of contact and the robot dynamics~\cite{wang2019impact}. This requires a  controller able to track motion and force profiles through impact and to accommodate substantial disturbances (\eg impulses exceeding the robot payload) resulting from inevitable estimation errors and motion mismatches.
Hence, the design of robot controller is subject of research, which we summarize with the last question; (Q6) \textit{what is an apt control scheme for tracking motions and making contact with fast-moving objects?}}  
        
\textcolor{black}{Failing to answer any of the above six questions will result in a large impulsive force when making contact. As a result, the object will be likely dropped, damaged or bounce off after contact. 
% Therefore, the aim of this paper is to mitigate impact and decrease impulsive force for dynamic manipulation of large-momentum object, improving the safety and success of dynamic manipulation tasks. 
% Therefore, the aim of this paper is to decrease impulsive forces during dynamic manipulation of large-momentum objects, and hence enable impact-aware dynamic manipulation tasks. }
Therefore, the aim of this paper is to decrease impulsive forces when making contact to enable impact-aware dynamic manipulation tasks with large-momentum objects. }

% To address these challenges we propose a holistic system that (a) estimates and predicts the object's motion, (b) plans and coordinates the motion of the dual-arm robot with respect to the moving object, and (c) controls the impact via tracking the planned motions and forces. Over the next few paragraphs we detail these three elements and highlight the relevant research questions addressed in this work.

\vspace{-2mm}
\subsection{Focus of our work and problem description}

\begin{figure}[t]
    \vspace{4mm}
	\def\svgwidth{1.0\linewidth}
  \input{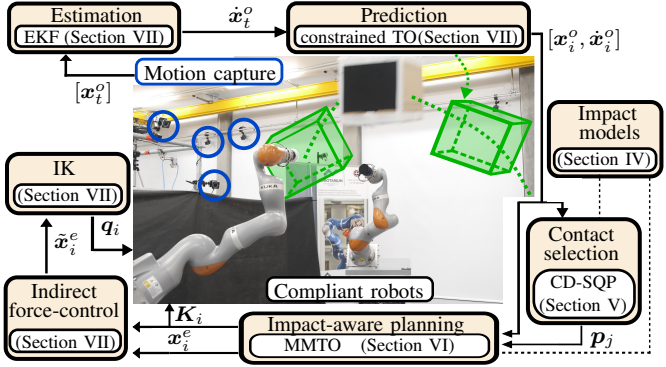}
    \vspace{-6mm}    
	\caption{Overview of the proposed system. The experimental setup is in the center. The flow of information starts from the motion capture and results into the input of the robots. Solid arrows denote information transmission   between modules and dotted lines denote functional dependency between components. }
 % \textcolor{blue}{I feel in this figure the image is too small and the text is a bit big}.}
	\label{fig:overview}
    \vspace{-5mm}    
\end{figure}

In this article, we investigate these six questions and propose a system with multiple modules.  These are: (i) the \textit{estimation} module where measurements of the object pose are collected and its velocity (linear and angular) is estimated using an EKF, (ii) the \textit{prediction} module where the future states of the object are predicted by solving an Initial Value Problem (IVP) with a constrained trajectory optimization (TO), (iii) \textit{impacts models}, a key ingredient of our framework, that are used in the two following modules, (iv) the \textit{contact selection} module where the optimal impact-wise contact locations on the object are found with our CD-SQP method, (v) the \textit{planning} module where our Multi-Mode Trajectory Optimization (MMTO) is used to generate the optimal dual-arm motions, stiffness and force profiles on-the-fly, (vi) the \textit{indirect force-control} module where the robots' set-points are adapted to track the desired force, and (vii) the \textit{IK} module (inverse kinematics) that computes the corresponding robots' configurations. \cref{fig:overview} illustrates the flow of information between these modules in our system. 

% a system that includes the following modules: (i) velocity estimation, (ii) trajectory prediction, (iii) contact selection mechanism, (iv) Multi-Mode Trajectory Optimization (MMTO)---which generates impact-aware motions---and (v) a variable stiffness controller.

% \subsubsection{Problem Description}
\label{sec:prob_desc}
% \paraDraft{Problem description }

% \begin{figure*}[t]
% 	\begin{center}
% 		\includegraphics[width=1.90\columnwidth]{figures/overview.png}
% 		\vspace{-10pt}
% 	\end{center}
% 	\caption{Overview of the proposed optimization framework .}
% 	\label{fig:overview}
% 	\vspace{-10pt}
% \end{figure*}

% \begin{figure}[t]
%     \vspace{4mm}
% 	\def\svgwidth{1.0\linewidth}
%         \input{figures/overview_diagram_reduced}
%     \vspace{-4mm}    
% 	\caption{Overview of the proposed system. The experimental setup is in the center. The flow of information starts from the motion capture of the setup and results into the input of the robots. Black arrows denote transmission of information between modules and dotted lines denote functional dependency between components. \textcolor{blue}{I feel in this figure the image is too small and the text is a bit big}.}
% 	\label{fig:overview}
%     \vspace{-4mm}    
% \end{figure} 

% \begin{figure*}[t]
% 	\def\svgwidth{1.0\linewidth}
%         \input{figures/overview_diagram_big2}
%     \vspace{-4mm}    
% 	\caption{Overview of the proposed system. The experimental setup is in the center. The flow of information starts from the motion capture of the setup and results into the input of the robots. Black arrows denote transmission of information between modules and dotted lines denote functional dependency between components. }
% 	\label{fig:overview}
%     \vspace{-4mm}    
% \end{figure*} 

\textcolor{black}{Our goal is to develop an optimisation-based approach that minimizes the impulsive force for bi-manual catching of large momentum objects.}
% \joao{Is the following the xxx a typo or variable name to desgin? There should be a color highlight here so you don't forget this.}
We denote with $\bm{x}_t \in \mathbb{R}^o$ the pose of an object and with $ \dot{\bm{x}}_t $ its velocity at time $t$. Assuming that the future trajectory of the object intersects the workspace of the robots, we denote with $\bm{q}_{\bm{t}} \in \mathbb{R}^m$ the trajectory of the robot arms in the configuration space, with $m$ Degrees of Freedom (DoF), at times $\bm{t} = \{t+1, ..., t+\tau+\nu\}$, where 
$t+\tau$ is the time it takes for the object to intersect the robot's workspace and $\nu$ is the planning horizon.
Further, $\bm{K}_{\bm{t}} \in \mathbb{R}^o$ denotes the stiffness of the robot end-effectors, with a task space of dimensionality $o$.
Given the known mass $M$, inertia $\bm{I}$ and shape $\mathcal{S}$ of the object, the catching problem at time $t$ can be described by:
\begin{equation}
    \label{eq:catching_prob}
    \bm{q}_{\bm{t}|t}, \bm{K}_{\bm{t}|t} = \mathtt{O}(\bm{x}_t, \dot{\bm{x}}_t; M, \bm{I}, \mathcal{S}).
\end{equation}
The notation $\bm{t}|t$ in the subscript indicates that the values of $\bm{q}$ and $\bm{K}$ are planned for the times $\bm{t}$ given the current information available at time $t$.

Formally, an optimisation problem for catching large momentum objects (see~\eqref{eq:catching_prob}) can be written as: \vspace{-0.5mm}
\begin{IEEEeqnarray}{CCC}
    \IEEEyesnumber\label{eq:cathing_opt} 
    \IEEEyessubnumber \label{eq:cathing_cost}
    \hspace{-13mm}\min_{\hspace{10mm}\bm{q}_{\bm{t}}, \bm{K}_{\bm t }, \bm{x}_{\bm{t}}, \dot{\bm{x}}_{\bm{t}}}  &  \mathtt{J}\left(\bm{q}_{\bm t }, \bm{K}_{\bm  t }, \bm{x}_t, \dot{\bm{x}}_t \right)~ \\
    \hspace{-14mm}~\text{s.t.}~  
    \IEEEyessubnumber \label{eq:catching_est} &
    \hspace{-10mm}\dot{\bm{x}}_i = \mathtt{e} (\bm{x}_{0,...,t}; \bm{I}),\hfill \forall i=[0,...,t],~~~\\
    \IEEEyessubnumber \label{eq:catching_pred} & 
    \hspace{-11mm}\left[\bm{x}_i, \dot{\bm{x}}_i\right] =  \mathtt{p}( \bm{x}_t, \dot{\bm{x}}_t; M, \bm{I}), \hfill \forall i=[t+1,...,t+l],~~~\\
    \IEEEnonumber & 
    \hspace{-10mm}\text{with } j=t+\tau,\text{ where }l>\tau,~\hfill\\
    \IEEEyessubnumber \label{eq:catching_grasp} & 
    \hspace{-6mm} \bm{p}_j = \mathtt{c}(\bm{x}_j, \dot{\bm{x}}_j; \mathcal{S}), \hfill \\
    \IEEEyessubnumber \label{eq:catching_plan} & 
    \hspace{-6mm}\bm{q}_i, \bm{K}_i= \mathtt{m}(\bm{x}_j, \dot{\bm{x}}_j, \bm{p}_j; M, \bm{I}), \hfill \forall i=[j,...,j+\nu],~~~
\end{IEEEeqnarray}
where in~\eqref{eq:cathing_cost} $\mathtt{J}(\cdot)$ is the cost function, in~\eqref{eq:catching_est} $\mathtt{e}(\cdot)$ denotes the \textit{estimation} process (see~\cref{fig:scenario}(a)) and in~\eqref{eq:catching_pred} $\mathtt{p}(\cdot)$ denotes the \textit{prediction} process, with $l$ being the prediction horizon (see~\cref{fig:scenario}(b)). Assuming that the object's predicted trajectory intersects the robot's workspace, the index $t+\tau$ indicates that moment in time the object enters the workspace. Based on the predicted state of the object at $t+\tau$, in~\eqref{eq:catching_grasp} $\mathtt{c}(\cdot)$ the \textit{contact selection} process  (see~\cref{fig:scenario}(c)) and 
in~\eqref{eq:catching_plan} $\mathtt{m}(\cdot)$ the \textit{impact-aware motion planning} process (see~\cref{fig:scenario}(d)) are completed.  \vspace{0.25mm}

\textbf{Problem Statement}:
\textit{Design an optimisation-based system able to smoothly catch large-momentum objects given only position measurements of the object till time $t$. This requires estimating and predicting the object's state, selecting contacts on its surface, planning the arms' motion, force and stiffness \textcolor{black}{to cope with impacts,} and realizing the catching motion.}

\vspace{-2mm}
\subsection{Contributions}
\textcolor{black}{As shown in Fig.~\ref{fig:overview}, we propose an optimization-based approach to minimize the impulsive force for bimanual catching of large-momentum objects}. Our framework enables the bimanual robots to generate on-the-fly impact-aware contacts, motion, force and stiffness plans that are conditioned on the position, speed, and shape of the object, as well as the environment that constraints the motion of the object. 
% Note that there is no scientific contribution in Estimation, Prediction, IK, and Indirect force-control modules in Fig.~\ref{fig:overview}.
\textcolor{black}{
Our scientific focus is on the impact models, contact selection and impact-aware planning modules, while for the estimation, prediction, IK, and indirect force-control modules we used existing approaches. Yet, to achieve the goals stated above, we implemented and integrated all these individual components.}
% \textcolor{blue}{Note that there is no scientific contribution in Estimation, Prediction, IK, and Indirect force-control modules in Fig.~\ref{fig:overview}.}
%  We propose an optimization-based approach for achieving non-prehensile catching of large momentum objects, which includes state estimation and motion prediction of the moving object, contact selection and force regulation, and impact-aware planning of robot motion.    
% The proposed system enables bimanual robots to generate on-the-fly impact-aware motion plans that are conditioned on the position, speed, and shape of the object, as well as the environment that constraints the motion of the object. 
The main contributions of this paper, under the assumptions that we have a good priori knowledge of the mass and inertia of the objects, are the following:

\begin{enumerate}

    \item The impulse along tangential and normal directions of the contact are analysed using a 3D compliant impact model, as a function of contact location, direction and duration. Based on that, the contact selection principles are proposed.

    \item According to the contact selection principles, an online impact-aware contact selection method  is developed based on Sequential Quadratic Programming (SQP). This allows online-searching of multiple contacts on arbitrary shaped objects, which minimize the impulse between the robots and the moving object.
 % which minimizes the impulse when the robot establishing contact with the object according to the predicted object motion.
 
	% This formulation uses three-dimensional (3D) compliant impact model to analyse the impulsive force along tangential and normal directions.
 % as a function of; (i) the magnitude and the direction of the pre-contact velocity and (ii) friction coefficient. 

     % \item An online impact-aware contact selection is developed based on Sequential Quadratic Programming (SQP) formulation.    for arbitrary shaped objects based on contact selection principles. These contacts minimize the impulse between the robot and the moving object.
	
	%Based on this analyses, we propose a Sequential Quadratic Programming (SQP) formulation that allows to online select impact-aware surfaces and locations on the object according to the predicted object motion. These are then used to establish contact with the object.
	
	\item Our contact force transmission model and corresponding parametric programming technique~\cite{stouraitis2020multi} are extended to encode both hybrid dynamics and hybrid control for multi-contact and 3D scenarios.
	Together with the proposed contact searching method, this enables simultaneous and online optimization of contact locations, motions, force and stiffness profiles in 3D for dual-arm robotic systems.

    \item The proposed methods and system are experimentally validated on hardware using two KUKA LBR iiwa robots, for the highly dynamic manipulation tasks of capturing swinging, spinning, tumbling, and flying large-momentum objects.
    
	% \item Experimental validation of the proposed methods using two KUKA LBR iiwa robots, for the highly dynamic tasks of capturing swinging, spinning, and flying large-momentum objects.
	% % showing the effectiveness of our system for handling large-momentum objects.
	% Especially, we propose an adaptive set point generator, via indirect force control, to track the planned forces with commercial compliant robot manipulators. 
    % \item Experimental validation of the proposed method in 
    
	%, showing the effectiveness of our system for handling large-momentum objects.
	% Especially, we propose an adaptive set point generator, via indirect force control, to track the planned forces with commercial compliant robot manipulators.
% 	\joao{The last sentence doesn't sound too impressive after referring the experiment validation. But i didn't really understand what does this sentence refers to.}
% 	Experimental validation of the proposed methods using two KUKA LBR iiwa robots, for the highly dynamic tasks of capturing swinging, spinning, and flying large-momentum objects. Especially, we propose an adaptive set point generator, via indirect force control, to track the planned forces with commercial compliant robot manipulators. }
\end{enumerate}

% \theo{Merge the two following paragraphs into one}
%The remainder of this paper is organized as follows.
We structured the paper as follows.
%The related work is presented in~\cref{sec:rel_work}.
\cref{sec:rel_work} contains the related work.
In~\cref{sec:impact_modelling}, we provide the preliminaries on impact modelling and a set of hypotheses for impact-aware contact selection. In~\cref{sec:contact_searching}, we describe the proposed impact-aware contact searching method and 
%we analyse the impulsive force along normal and tangential directions %when making contact along different directions and at different locations, 
% and further propose an autonomous impact-aware dual-contact selection algorithm based on the impact analysis in~\cref{sec:contact_searching}.
\cref{sec:multi_mode_optimization} presents the multi-mode trajectory optimization used for impact-aware planning. 
% which simultaneously optimizes the end-effector motion, contact force and stiffness. 
\cref{sec:sim_results} and~\cref{sec:exp_results} present several simulation and hardware experimental results of bi-manual catching of large swinging, spinning, and flying object, along with the implementation details on estimation and prediction. Sections VIII and IX 
% \cref{sec:discussion} and~\cref{sec:conclusion} 
discuss and conclude our work, respectively.
\vspace{-1mm}

% conclusions and future work.

% As this work is focused on dynamic manipulation and hence, impacts, we will first provide the preliminaries on impact modelling and then we will describe the proposed impact-aware contact and planning methods, respectively. The implementation details of estimation, prediction, indirect force-control and IK are provide in~\cref{sec:exp_results} along with the algorithmic flow and the description of the experimental setup.

% \subsection{Problem Statement and Contributions}
% In this paper, we investigate the problem of non-prehensile bimanual catching of large-momentum objects as shown in Fig.~\ref{fig:scenario}. 

% To capture an object in flight, a bimanual robot needs to first estimate the current state (pose and velocity) of the object and predict the future trajectory of the object.

% %===============================================================================
\section{Related Work}\label{sec:rel_work}

\subsection{Object Motion Estimation and Prediction}

In order to catch a fast-moving object, such as swinging or flying object, we strive for methods that are able to estimate both the linear and angular velocities of the object in a fraction of a second. Hence, here we review methods towards this goal.

\subsubsection{Model-based methods}
Towards capturing a small spherical object~\cite{bauml2010kinematically} where the orientation of the object can be ignored, both Kalman Filter (KF) without air drag and with air drag~\cite{dong2020catch} have been used to estimate the state of the object.
% \joao{This is not a complete sentence:}
Given these estimates, an Initial Value Problem (IVP)~\cite{lampariello2021optimal} can be solved
to predict the future trajectory of the object by numerically integrating its dynamics equation. \textit{These works only consider free-flying objects that are not constrained by the environment and only the linear part of their motion.} 
% \textit{These works only consider the objects in free-flying mode that are not constrained by the environment and only their linear motion.} 
% \textit{These works only consider the linear part of the motion and objects only in free-flying that are not constrained by the environment.} 

A hybrid estimator was proposed~\cite{jia2019batting} for in-flight objects in the case of 2D motion, where an Extended Kalman Filter (EKF) was used to estimate the linear position and velocity, and a least-square fitting method to estimate a single axis angular motion. Yet, to estimate and predict an object's motion in 3D, its inertia characteristics are also required.
Masutani~\textit{et al.}~\cite{Masutani1994} first studied the problem of inertial ratio parameters identification in 3D,  given the angular velocity. Similar works motivated by space applications proposed EKF-based methods~\cite{aghili2012prediction} to estimate the state and the inertial ratio parameters of objects.  
% constrained nonlinear least squares~\cite{hillenbrand2005motion} approaches to simultaneously estimate the state and the inertial ratio parameters of the objects. 
Yet, identifying the inertial parameters of an unperturbed free-flying object is a lagging process, due to the observability of the problem. \textit{These approaches require tens or hundreds of seconds to converge, hence they are unsuitable for catching fast-moving objects.} 

% In the contact domain impact ~\cite{jongeneel2021model}

\subsubsection{Model-free methods}
In addition to the model-based approaches, recent works~\cite{kim2014catching, salehian2016dynamical, yu2021neural} proposed model-free estimation methods using  learned models. These approaches are particularly useful in cases where the dynamics model of the object are not available, \eg half-filled water bottle~\cite{kim2014catching}. 
%Recently, a recurrent neural network model~\cite{yu2021neural} has been used for estimating and predicting the motion for in-flight uneven and unseen object. 
\textit{However, their prediction accuracy is highly dependent on the collected data~\cite{kim2014catching}, while predictions of unseen objects relies on datasets~\cite{yu2021neural} that are only feasible for small and light objects. For heavy and large objects, the data collection process can be impractical and strenuous.} 

\textcolor{black}{In this work, first we use a EKF~\cite{MooreStouchKeneralizedEkf2014} to estimate the current linear and angular velocities of the object given its inertia parameters. Secondly, we adopt a nonlinear program~\cite{kelly2017introduction} to predict the future trajectory of the object
%. The latter is an IVP, 
which can be generalized to a variety of environments that may constrain the motion of the object, \eg an object sliding on a table or being suspended from the ceiling. }\vspace{-2mm}

% There are a number of successful tasks using RL that are related to the task in this paper. For example, in [1] a robot learns how to launch and catch a pancake on a frying pan, which is also a non-prehensile grasp. In [2], a robot learns to decelerate an incoming ball and later push it with two paddles. In [3], a ball in a cup is solved, and in [4] juggling is solved. 

% In common, while [3] and [4] could be seen as a
% "semi-prehensile" (due to the concave shape of the
% end-effector), what these references have in common with the submitted paper is the fact that a smooth, compliant deceleration is necessary to avoid bouncing.

% [1] Kormushev, P., S. Calinon, and D.G. Caldwell. “Robot Motor Skill Coordination with EM-Based Reinforcement Learning.”  In Proceedings of the IEEE/RSJ International Conference on Intelligent Robots and Systems, 3232–37, 2010.
% [2] Maeda, G., Okan K, and Morimoto, J. “Phase Portraits as Movement Primitives for Fast Humanoid Robot Control.” Neural Networks 129 (2020): 109–22.
% [3] Celemin, C., G. Maeda, J Ruiz-del-Solar, J. Peters, and J. Kober. “Reinforcement Learning of Motor Skills Using Policy Search and Human Corrective Advice.” International
% Journal of Robotics Research (IJRR) 38, no. 14 (2019): 1560--1580.
% [4] Ploeger, Kai, Michael Lutter, and Jan Peters. “High Acceleration Reinforcement Learning for Real-World Juggling with Binary Rewards.” 

\subsection{Impact Mechanics}

\textit{When catching large momentum objects, contacts may stick, slide, or bounce---due to impact.} Hence, to analyze impacts under various conditions we study works on the mechanics of impact.

% \paraDraft{Analysis and modelling of impact}
% \subsubsection*{Computat}
A key concept for analysing impacts is the \textit{angle of incidence}. This is defined as the angle between the pre-impact relative velocity vector and the normal vector of the contact plane, and is used to differentiate oblique impacts from direct impacts~\cite{stronge2018impact}, \ie whether there is a tangential impact component. 
Brach~\cite{brach1989rigid} studied the 2D collisions and showed that the coefficient of restitution and the ratio of tangential impulse to normal impulse are related to energy loss and the existence of sliding. Keller~\cite{keller1986impact} and Stewart et al.~\cite{stewart2000rigid} analysed impact of two rigid bodies with friction and showed that frictional impulse can be calculated based on the sliding velocity. More recently, Jia~\cite{jia2013three} studied an energy-based 3D impact model with both tangential compliance and friction, which considers both sliding and sticking contact modes.
% And the impact analysis was carried out by using the normal impulse rather than time as the independent variable.
A comprehensive survey %of contact and on 
of impact models can be found in~\cite{gilardi2002literature}.

In this paper, \textcolor{black}{we aim to inform the contact selection with impact-specific optimization criteria, such as \textit{angle of incidence}. Thus, we use the 3D impact model with tangential compliance~\cite{jia2013three} to investigate a variety of impact scenarios and identify impact-specific metrics for contact selection.} 
Also, we use a simpler model~\cite{nagurka2004mass} of impact that have been used to \textcolor{black}{relate impact properties such as  duration and restitution coefficient to the mass, damping and stiffness parameters of the mechanical system.} This model may be less accurate, but they are particularly useful when used for online planning and control. An extensive study on hybrid models and their use on robotics can be found in~\cite{johnson2016hybrid}. \vspace{-2mm}

\subsection{Contact Selection in Manipulation}

Manipulation can be split into two paradigms~\cite{mason1999progress, roa2015grasp}. The prehensile, \ie grasping, where the goal is to constrain the object within a ``hand" such that the manipulation problem can be treated as a motion planning problem~\cite{lamiraux2021prehensile}. The non-prehensile---with which dexterity is exhibited---where the relative pose between the object and the end-effector or ``finger" can change at anytime.

% In both paradigms contacts are crucial. In the former, contact forces on the object are deemed useful or not according to conditions such as form-closure and force-closure~\cite{markenscoff1990geometry, bicchi1995closure}.
In both paradigms contacts are crucial. In the former, contact forces on the object are considered as passive or active according to the grasp properties such as form-closure or force-closure~\cite{markenscoff1990geometry, bicchi1995closure}. 
These grasp conditions are quasi-static and applied in problems where the object is graspable. \textit{As a result this paradigm is only relevant if the object is moving slowly or if the object can get attached to the robot ``hand" immediately~\cite{kim2014catching, salehian2016dynamical}. }

For non-prehensile manipulation, objects' state is altered by applying unilateral contact forces at the selected contact locations. This allows manipulation of large and heavy objects with as few as a single contact at a time and in dynamic setups (subject to impact handling). The main charm of this paradigm~\cite{lynch1999dynamic, ruggiero2018nonprehensile} is that it can be used to solve any type of manipulation problem~\cite{toussaint2018differentiable}. Hence, it has become increasingly used in tasks such as planar pushing~\cite{moura2022non,Hogan2020}, pivoting objects~\cite{hou2019robust}, large object manipulation under gravity~\cite{yan2018dual, stouraitis2020online}, reactive use of tools~\cite{toussaint2022sequence}, and small object stopping on an inclined air-hockey table~\cite{pekarovskiy2014hierarchical}. 

% intended contact mode switches need to be planned, and unexpected contact mode transitions need to controlled---all of which make it very challenging problem.

In non-prehensile manipulation, to manipulate an object to a desired position, contact locations need to be particularly selected. The selection process is achieved either via sampling~\cite{zhou2017pushing} or optimization~\cite{toussaint2020describing, stouraitis2018dyadic, murooka2020optimization} of the contact locations.  
In the former, the contact locations augment the configuration space (object pose), which are sampled such that the object can be pushed (quasi-statically) to the goal location. In dynamic setups, 
the object state is high-dimensional as it includes position, velocity, and acceleration, \etc \textit{Hence, finding optimal solution via sampling in such configuration spaces is time-consuming.}
In the latter, contact searching is achieved by following contact-informative gradients. In terms of feasibility, the point-of-attack~\cite{toussaint2020describing} was introduced to find the wrench exchange between two rigid bodies which is consistent with both contact geometry and dynamics equation. A sequence of QPs were bridged via a projection mechanism~\cite{murooka2020optimization} such that a sequence of local optima leads to global optima. These approaches require a smooth surface representation, \eg convex sphere-swept shapes~\cite{toussaint2018differentiable},  splines~\cite{stouraitis2018dyadic} or learned signed-distance-fields for nonconvex objects~\cite{driess2022learning}. 

\textcolor{black}{Here, we represent objects with 3D meshes and we propose a novel impact-informed optimization method based on SQP. This allow us to find the (impact-wise) optimal contact locations for catching fast-moving objects of arbitrary shape (even non-convex) with minimal impulsive forces.} 

\vspace{-2mm}
\subsection{Hybrid Motion Planning} 

In robotics, there is a variety of Hybrid Trajectory Optimization methods~\cite{Hogan2020, toussaint2018differentiable} that are able to synthesize multi-contact manipulation plans. Typically, planning-focused methods study problems where impacts are not likely to occur. However, there has been increasing interest~\cite{dehio2022dual, khurana2021learning} in transferring multi-contact manipulation behaviours to the hardware while simultaneously increasing the operation speed of robots.
\textit{These setups generate increasingly strong impacts}, therefore a variety of techniques were adopted to cope with them, such as inertia matrix shaping~\cite{wang2022inverse}, stiffness regulation~\cite{stouraitis2020multi}, motion regulation and impact-invariant projection~\cite{wang2019impact}.

In our previous work~\cite{yan2018dual}, we studied catching large  objects by separating the motion planning into three phases; (i) pre-contact, (ii) contact, (iii) post-contact. This provides sub-optimal solutions and requires re-designing these distinct but consecutive phases in every scenario, \eg when changing the workspace, the object shape or velocity. Further, to eliminate impacts, we induced a common heuristic~\cite{amanhoud2019dynamical} that constrains the relative velocity at contact to zero.

In this paper, \textcolor{black}{to cope with impacts in very dynamic tasks and to relax the above assumption, we extend our multi-mode trajectory optimization framework (MMTO)~\cite{stouraitis2020multi}
from 1D to 3D, from one to two arms and to scenarios with gravity. This enables us to generate optimal solutions online} that involve synchronous motion of the two arms with respect to the object motion in 3D as well as the internal %intrapersonal\footnote{Intrapersonal refers to processes that occur within one agent.} % 
coordination of the force and stiffness profiles of the bi-manual robot. 

\vspace{-2mm}
\subsection{Hybrid Motion Control} 

Control-focused methods have also been proposed towards regulating the transitions between free motion and motion in contact, \ie impacts. \textit{Although, dedicated controllers for impacts are essential, given the inherent limitations of the hardware~\cite{haddadin2009requirements}, the impacts that a 
stand-alone controller is capable of dealing with are limited~\cite{stouraitis2020multi}.}

To cope with this, the reference spreading~\cite{saccon2014sensitivity} concept was introduced. According to this concept more than one reference are provided in the neighborhood of impact, such that the appropriate jump can be triggered at the moment of impact. This has been used in hybrid controllers~\cite{rijnen2015optimal}, %extensions of those with linear quadratic regulator~\cite{rijnen2015optimal} 
and QP controllers~\cite{van2022robot}.  
On another line of work, to avoid explicitly dealing with the combinatorial aspects of hybrid motions, provably stable control policies were synthesized 
into optimization algorithms 
based on complementary contact constraints~\cite{aydinoglu2020contact}. 

A variety of compliance controllers have also been used to deal with the imprecise timing of the transition in the reference motions.
Depending on the conditions, whether the robot is in free-motion or in contact, a hybrid force controller that switches between impedance and admittance control was proposed~\cite{ott2010unified}. Similarly, in~\cite{roveda2015optimal} the gains of an adaptive impedance controller were optimized for each conditions.

\textcolor{black}{In this paper, we adopt an indirect force control mechanism based on an impedance controller~\cite{lutscher2017hierarchical}, which allows us to track the desired force when in contact and concurrently modulate the stiffness of the robot to deal with the imprecision during and after the contact transition.}

\section{Preliminaries}\label{sec:impact_modelling}
% \textcolor{red}{The key reason why we use the cd DS: is to have forces that are both according to the impact model and smooth!!!! Relevant slides :
% \url{https://www.me.utexas.edu/~dsclab/leks/DSC_Vibration_Modeling.pdf} , slide 8}\\

% Generally speaking, the impact comes from the velocity mismatch at contact location between the object and the robot, because of the uncertainties resulted from perception, planning and control. 
% Impulsive forces result from contact events that happen when the velocity of object and the robot
% During impact events  impulsive forces are generated to resolve the velocity mismatch between the two bodies. % , in this section the evolution of an impact event along with the respective impulse are analyzed. 

% According to this analyses 
% several principles used to select contact are established along with a second-order contact force transmission for non-prehensile manipulation.
% to achieve smooth transition from free motion to contact.
% \theo{I tried to re-write the previous paragraph, but it is repeated in the next subsection, so I  merge them.}
% and impact dynamics.
% In scenarios where two objects collide with each other, such as a robot makes contact with a moving object as shown in Fig.~\ref{fig:contact_selection} (1), the non-zero relative velocity between them will result in an impulsive force. 

\begin{figure}[t]
	\begin{center}
		\includegraphics[width=0.90\columnwidth]{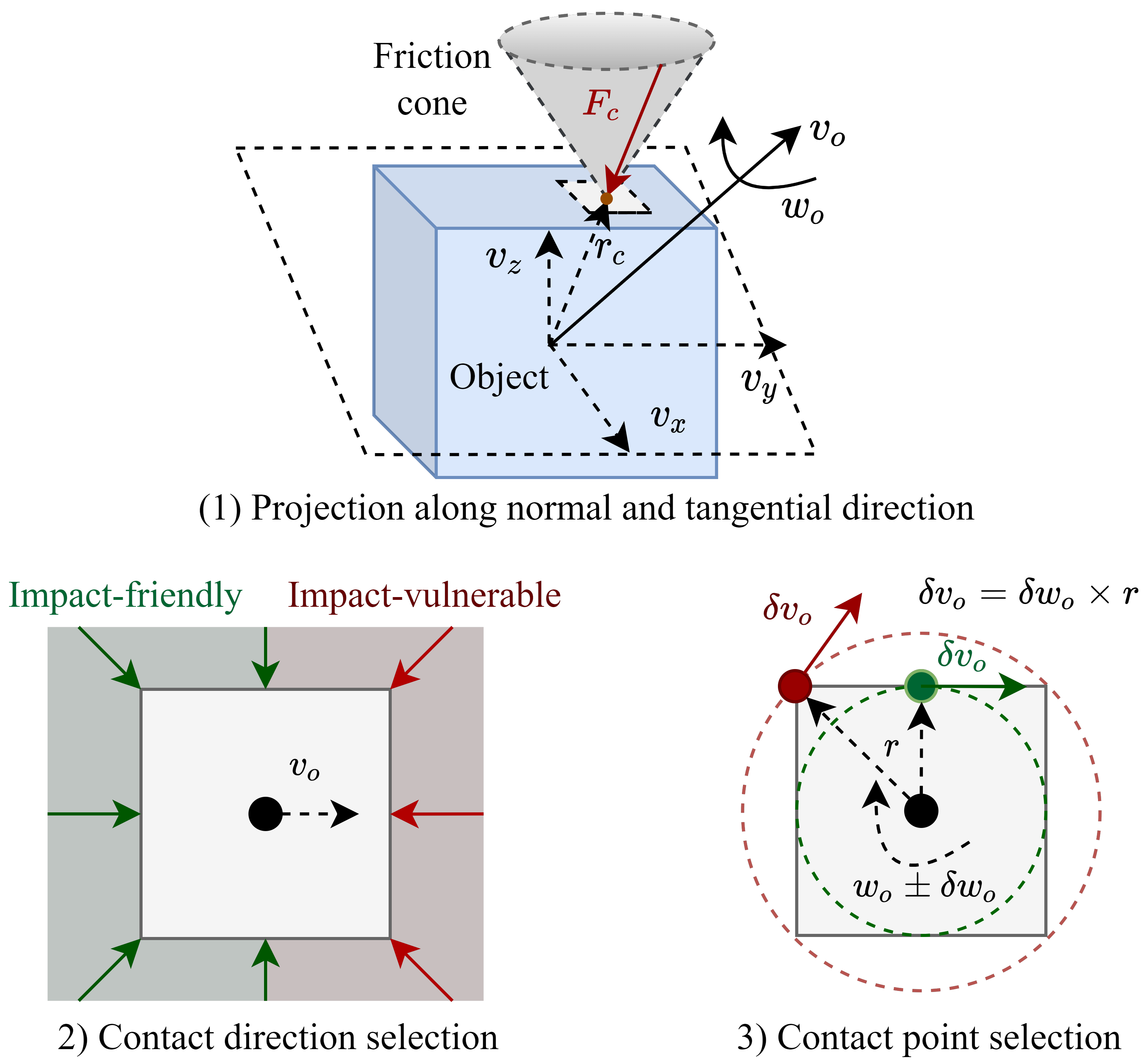}
		\vspace{-10pt}
	\end{center}
	\caption{Projection of contact velocity and force along normal and tangential directions (1) and 2D illustration for contact selection (2-3).}
	\label{fig:contact_selection}
	\vspace{-5mm}
\end{figure}

% For contact transition of multi-body system,
Impact events occur when two rigid bodies~(\eg an object and a robot end-effector, see~\cref{fig:contact_selection}(1)) collide at non-zero relative velocity. 
The impulse resulting from these events is analyzed here based on a 3D compliance impact model. The analysis led to a set of hypotheses that are used for the contact selection (see \cref{sec:contact_searching}), while a simpler compliance impact model is used to determine the optimal contact force profile in trajectory optimization (see \cref{subsec:cntforceTrans}). 

% \theo{Let's discussion notation, my suggestion is to drop the "o" subscript for the object}
\vspace{-2mm}
\subsection{{Impacts}}\label{subsec:impacts}

%Without loss of generality, we assume that the velocity of the robot is zero at the moment of contact \textcolor{red}{ and the impact/contact between the robot and the object is modelled as point-contact}. 
The relationship between the impulse and the velocity jump of the object before and after impact can be described as:
% \begin{equation}
%     M \left(\bm{\mathrm{v}}^+ - \bm{\mathrm{v}}^- \right) = \bm \Lambda \delta t,
% \end{equation}
% where $M$ is the mass of the system, $\bm{\mathrm{v}}^-$ and $\bm{\mathrm{v}}^+$ are the pre-impact and post-impact relative velocities, respectively. $\bm \Lambda$ is the impact force and $\delta t \simeq 0$ is the impact's duration.
% First, we write the impact dynamics as follows:
% \begin{equation}\label{eqn:impact_dynamics}
% \begin{split}
%     m_o \left(\bm v_o^+ + \bm r_c \times \bm \omega_o^+  -\bm v_o^- -\bm r_c \times \bm \omega_o^- \right) &=  \bm \Lambda \\
%     \bm I_o \left(\bm \omega_o^+ - \bm \omega_o^- \right) = \bm r_c \times \bm \Lambda & ,
% \end{split}
% \end{equation}
\begin{equation}\label{eqn:impact_dynamics_linear}
    M \Delta \bm v = M \left(\bm v^+   -\bm v^-  \right) = \int_0^{\Delta t}\bm F \, dt =  \bm \Lambda,
\end{equation}
\begin{equation}\label{eqn:impact_dynamics_angular}
    \bm I \Delta \bm \omega = \bm I \left(\bm \omega^+ - \bm \omega^- \right) =\int_0^{\Delta t} \bm r \times \bm F \, dt  = \bm r \times \bm \Lambda ,
\end{equation}
where $M$, $\bm I \in \mathbb{R}^{3 \times 3} $ are the mass and inertia matrix of the object, respectively; \textcolor{black}{$\bm v \in \mathbb{R}^3$ and $\bm \omega \in \mathbb{R}^3$ are the linear and angular velocity of the object with respect to the robot}; $\bm r \in \mathbb{R}^3$ is the position vector  of the contact location with respect to the center of mass of the object.
\textcolor{black}{We assume that the contact between the robot and the object is modelled as point-contact.} The superscript ``$-$" and ``$+$" respectively represent the states before and after impact. The impulse $\bm \Lambda \in \mathbb{R}^3$ can be calculated by the integral of contact force $\bm F \in \mathbb{R}^3$. Assuming small enough impact duration $\Delta t$, the contact force $\bm F$ can be approximated to be constant. Hence, we can rewrite \eqref{eqn:impact_dynamics_linear} and \eqref{eqn:impact_dynamics_angular} as:
\begin{equation}\label{eqn:impact_model}
\begin{split}
\left[ \begin{array}{cc} 
M \bm E_3 & \bm O_3 \\ \bm O_3 & \bm I \end{array} \right] & \left[ \begin{array}{c} 
\bm v^+ - \bm v^-  \\ \bm \omega^+ - \bm \omega^- \end{array} \right]   \\ =& \underset{(i)} {\underbrace{ \left[ \begin{array}{cc} \bm E_3 & \bm O_3 \\ \bm r^\times & \bm O_3 \end{array} \right]}}  \underset{(ii)} {\underbrace{\left[ \begin{array}{c} 
\bm u_{\bm F} \\ \bm O_{3\times1} \end{array} \right]}} \underset{(iii)}{\underbrace{\|\bm F\|}}~ \underset{(iv)}{\underbrace{\Delta t}},
\end{split}
\end{equation}
where $\bm E_n$ and $\bm O_n$ are respectively $n \times n$ identity and zero matrices; $\bm r^\times$ is the skew symmetric matrix of $\bm r$; $\bm u_{\bm F}$ is the unit vector of $\bm F$ and $\|\bm F\|$ is 2-norm of $\bm F$. It can be seen from~\eqref{eqn:impact_model} that the impact is related to four elements; (i) the contact location, (ii) the contact force direction, (iii) the contact force magnitude, and (iv) the impact event duration.
 
In this paper, we present a model-based optimization approach to obtain the optimal combination of these four quantities, \ie our system determines on-the-fly the impact-resistant contact location $\bm r$ and contact direction $\bm u_{\bm F}$, as well as the magnitude $\|\bm F\|$ and duration $\Delta t$ of contact force $\bm F$.
% of impact.
% \theo{let's construct the flow of info for this section here and together. What are the main key points for the flow (milestones in thought process), what are the main take away points(new knowledge provided), what is the utility of these take away points in algorithmic implementations or where are they used.}
We will present principles for determining the impact-resistant contact location $\bm r$ and direction $\bm u_{\bm F}$ based on impulse analysis in this section, while the optimization of the magnitude $\|\bm F\|$ and time duration $\Delta t$ of contact force will be explained in Section~\ref{sec:multi_mode_optimization}.

% \subsection{{Impulse Analysis based on 3D Compliance Impact Model}}
 \vspace{-2mm}
\subsection{\textcolor{black}{{Impulse Analysis using Compliance Model of Impact}}}\label{subsec:3DcomplianceModel}
% Theo's comments, please keep for now 
% Use for the analysis of normal and tangential impulse...
% Assume that velocity before contact --- we don't know $\bm u_{\bm F_c}$
% $  impact = f(\bm u_{\bm F_c}, \bm r_c) $, 
% previous models have only normal impulse
% conclusion the tangential impulse is smaller than the normal
% Maybe move results of the analysis here
% we make contact perpendicular to the motion of the objectand when this happens we need to control the force along the normal of the surface (which perpendicular to the motion) --- hence, we can only control the force along this direction --> this assumption 1
% then we can use this to select contact

% \paraDraft{Motivate use of complex 3D compliance model}
\textcolor{black}{To analyse the impulse using~\eqref{eqn:impact_model}}, we need to know $\Delta \bm v$ and $\Delta \bm \omega$ in advance \textcolor{black}{(see in~\eqref{eqn:impact_dynamics_linear} and~\eqref{eqn:impact_dynamics_angular})}, which implies that the velocities ($\bm v^+$ and $\bm \omega^+$) after the impact need to be known. \textcolor{black}{Consequently,~\eqref{eqn:impact_dynamics_linear},~\eqref{eqn:impact_dynamics_angular} and~\eqref{eqn:impact_model} are only suitable for calculating the impulse after impact, which is not useful for simulating forward the impulse, nor analysing the effects of its four elements (see~\eqref{eqn:impact_model}).} 

% \paraDraft{Introduce here the 3D compliance model that we use and with what goal in mind}
Our goal here is to investigate the effect of the %four quantities
\textcolor{black}{contact location, direction and duration}, highlighted in~\eqref{eqn:impact_model}, on the resulting impulse. 
%\paraDraft{Maybe here; why is a spring damper suitable for modelling impact ... there is a relation via the energy dissipation}
%\theo{Can we relate as simplification the mass-spring-damper model to the 3D compliance model mentioned above?}
\textcolor{black}{To model impact and analyse the corresponding impulses, mass-spring-damper systems have been adopted~\cite{ jia2013three, nagurka2004mass}.}
% To simulate the impact process and analyse the corresponding impulse, a mass-spring-damper system is adopted to model real-world collision behavior~\cite{nagurka2004mass}. 
\textcolor{black}{Accordingly, we model the dynamic interaction between the object and the robot as well as the relationship between the Cartesian position and the contact force of the robot as: 
\begin{equation}\label{eqn:impact_force_model}
% \tag{6}
    M \Delta \ddot{\bm x} +B \Delta \dot{\bm x} + K \Delta \bm x = \bm F.
\end{equation}
where \textcolor{black}{$M$ is set to the desired mass (as the effective Cartesian inertia of the robot is configuration-dependent and variant) while $B, K$ are the desired damping and stiffness of the robot. $\bm x$, $\dot{\bm x}$ and $\ddot{\bm x} \in \mathbb{R}^3$ are respectively the position, velocity and acceleration of the robot end-effector;} and $\bm F$ is the contact force.
The Laplace transform of \eqref{eqn:impact_force_model} can be written as
% we can obtain the following transfer functions
\begin{equation}\label{eqn:admittance_transfer_function}
% \tag{7}
{\Delta \bm x} = \frac{1}{ M s^2+  B s+  K} {\bm F}  = \frac{\frac{K}{M}}{ s^2+  \frac{B}{M} s+ \frac{K}{M}} \frac{{\bm F}}{K},
\end{equation}
where the damping ratio is set to $\mathrm{\zeta}=1$, i.e. $B = 2\sqrt{M K}$, to guarantee the desired critically damped behavior of the dynamic contact interaction.}

The energy dissipation of such a system is caused by the damper and can be calculated as follows:
\begin{equation}
    \label{eq:mass_spring_damperEnergy}
    E_d = \int_0^{\Delta t} B\Delta \dot{x}dx = \int_0^{\Delta t} B \Delta \dot{x}^2 dt.
\end{equation}

%For a moving object that experiences an impact, 
% \textcolor{red}{the Newton's coefficient of restitution $c_r$ is calculated by }
% \begin{equation}
%     \epsilon_r  = \left|\frac{ v_+}{ v_-} \right| = \sqrt{\frac{\frac{1}{2}M{v^+}^2}{\frac{1}{2}M{v^-}^2}},
% \end{equation}
% \textcolor{red}{where $\epsilon_r=1$ represents a perfectly elastic collision,~$0\leq\epsilon_r\leq 1$ represents a real-world inelastic collision. ~$\epsilon_r=0$ represents a perfectly inelastic collision, where all kinetic energy is converted to deformation, and is often used to impose a no-rebound condition.}
% the dissipated energy---due to the impulsive force---during the collision is calculated as 
% \begin{equation}
%     \label{eq:impactEnergy}
%     E_{\Lambda} = \frac{1}{2}M{{v}^-}^2 - \frac{1}{2}M{{v}^+}^2.
% \end{equation} 

% In the perfectly inelastic collision scenario 
%

% \begin{figure}[t]
% % \includegraphics[width=1.0\linewidth]{figures/background/impact_model.png}
% \includegraphics[width=1.0\linewidth]{figures/background/impact_model_TheoEdit.pdf}
% % \vspace{-12pt}
% \caption{Equivalent Newton restitution model with spring-damper system}
% \label{fig:impact_model}
% % \vspace{-12pt}
% \end{figure}

\begin{figure}[t]
    \centering
    \def\svgwidth{0.95\linewidth}
    \input{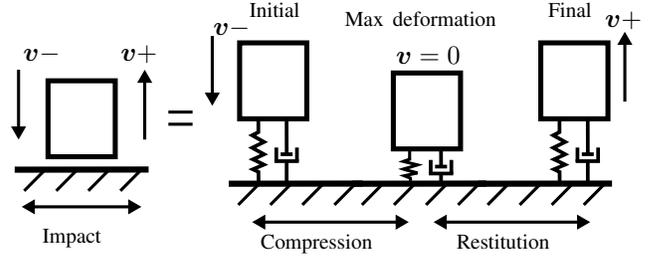}
    \caption{Correspondence between Newton's restitution model for impacts and the mass-spring-damper system.}
    \label{fig:impact_model}
    \vspace{-5mm}
\end{figure}

% By equating \eqref{eq:impactEnergy} with \eqref{eq:mass_spring_damperEnergy}, 
\noindent \textcolor{black}{Following~\cite{nagurka2004mass}}, the energy loss during impact can be modelled with a mass-spring-damper system, where the dissipated energy during compression and restitution stages (see \cref{fig:impact_model}) is related to both stiffness and damping parameters.
% where $\delta t$ is the duration time of impact.

\textcolor{black}{Here, we adopt a 3D compliance impact model with three spatial springs~\cite{jia2013three} which has been used to simulate forward the compression and restitution impulses} and compute the bodies' velocities after impact. %By updating 
Utilizing the energies stored in the springs, the impulse and the velocity changes during impact can be updated as follows:  
% \theo{remember to fix notation for iterations - has to be subscript}

\begin{equation}\label{eqn:impulse_iteration}
    \bm \Lambda_{i+1} = \bm \Lambda_{i} + \left(\frac{\partial \Lambda_x}{\partial \Lambda_z} \hat{\bm x} + \frac{\partial \Lambda_y}{\partial \Lambda_z} \hat{\bm y} + \hat{\bm z}\right) \Delta \Lambda_z
\end{equation}

% \begin{equation}\label{eqn:linear_velocity_variation}
% \Delta \bm v_o = \left(\frac{\bm E_3}{m_o}- \bm r_c^\times \bm I_o^{-1} \bm r_c^\times \right) \bm \Lambda,
% \end{equation}

\begin{equation}\label{eqn:linear_velocity_variation}
\Delta \bm v = \frac{\bm \Lambda}{M}, \ \Delta \bm \omega= \bm I^{-1}\bm r^\times \bm \Lambda,
\end{equation}
% \begin{equation}\label{eqn:angular_velocity_variation}
% \end{equation}
where $\Lambda_x$, $\Lambda_y$ and $\Lambda_z$ are respectively the three components of the impulse $\bm \Lambda$; $\Delta \Lambda_z$ is the incremental impulse along the normal direction at each iteration.
% $\Delta \bm v$ and $\Delta \bm \omega$ are the linear and angular velocity changes of the object after impact.

This model allows us to simulate the evolution of both the normal and tangential components of the impulse, while considering friction and the fact that the \textcolor{black}{instantaneous contact} between the two bodies can be sticking or sliding. %\footnote{Sliding and sticking contacts are the crux of manipulation, and they can be both intentional and unintentional~\cite{mason1999progress}.\lei{do we need this footnote?}}%.  
The 3D impact model is briefly introduced in~Appendix\ref{appendix:impact_modelling}. % and more details about it can be found in~\cite{jia2019batting}.
% More details about the 3D impact model can be found in~\cite{jia2019batting}, which is briefly introduced in~Appendix\ref{appendix:impact_modelling}. 
Based on the forward simulation of impulse, we can decide the best contact direction and  location by analysing the normal and tangential impulse components.

\vspace{-2mm}
\subsection{Contact Selection based on Impulse Analysis}
\label{subsec:impactProps}
% \paraDraft{Observation based on equations above}
In the 3D compliance impact model, the contact force is decomposed into the normal contact force $\bm F_z$ that is unlimited\footnote{Normal contact force $\bm F_z$ can be infinite unless the mechanical damage is reached.} and the tangential contact force $\bm F_{tan}={\bm F_{x}} + \bm F_{y}$ that is constrained by the friction cone. 
Further, the included angle between the object velocity and the contact force affects the distribution of impulse along normal and tangential directions.  
Based on these two observations we make the following hypotheses. These, along with the two observations above, are both validated with a variety of simulations in~\cref{sec:sim_results}. 

\begin{hypothesis}\label{hypo:orientation}
To minimize the magnitude of the impulse, the ideal contact direction(see~\cref{fig:contact_selection}) is defined to be perpendicular to the linear velocity of the object at the contact point, i.e. $\bm u_{\bm F}\perp \bm v$. This implies that the linear velocity at the contact point lies in the tangential plane of the contact surface. Hence, the normal impulse is minimal and the tangential impulse will be limited by the friction cone of the contact surface. 
% To minimize the magnitude of the impulse, the ideal contact location is defined to be the one with normal vector perpendicular to the linear velocity of the object at the contact point, i.e. $\bm u_{\bm F}\perp \bm v$. This implies that the linear velocity at the contact point lies in the tangential plane of the contact surface. Hence, the normal impulse is minimal and the tangential impulse will be limited by the friction cone of the contact surface. 
% We can decompose an impulse along the normal and tangential directions at the contact location.
% Make contact along the normal direction of relative motion is preferable because the corresponding tangential impulse is limited by the friction force in friction cone of normal contact force.
\end{hypothesis}
By selecting the contact surface according to~\textit{\cref{hypo:orientation}}, the unconstrained contact force lies in the orthogonal space of the object motion, and hence it will not be directly used for halting the object motion. On the other hand, the constrained contact force along tangential direction can used to absorb all the momentum of the object.
The algorithmic details on how to select optimal contact surfaces according to~\textit{\cref{hypo:orientation}} are given in~\cref{sec:contact_searching}.

\begin{remark}
As shown in~\cref{fig:contact_selection} (2), the contact surface that allows to apply normal forces in the same direction with object's velocity vector would also minimize the magnitude of the impulse. However, we are only interested in contacts that are able to reduce the velocity of the moving object after the impact event. Thus, such contact surfaces are deemed not useful and are not considered in this paper.
\end{remark}
Additionally, similar to the grasp quality measurement in~\cite{roa2015grasp}, we introduce the following assumption for selecting optimal contact locations  which minimize the effect of the motion uncertainty on the impact.

\begin{hypothesis}\label{hypo:position}
% \theo{This seems to be the same as in the grasping literature. See section 3.2.3 in \url{https://link.springer.com/article/10.1007/s10514-014-9402-3}}
Given the determined optimal contact surface, the optimal contact location should be selected such that the effect of motion uncertainty on the impact is minimized, i.e. minimizing the contact velocity $\delta \bm v = \delta \bm \omega \times \bm r$ resulting from the angular motion uncertainty $\delta \bm \omega$. Without loss of generality, in the case of isotropic uncertainty ($\delta \omega_x = \delta \omega_y = \delta \omega_z$), this condition is equal to minimizing $\| \bm r\|$ (see~\eqref{eqn:impact_model}).

% \theo{for Theo: I like it, can we be more clear about the 1st sentence? and can we mathematically prove the 2nd?}
\end{hypothesis}

The algorithmic details on how to select optimal contact locations according to~\textit{\cref{hypo:position}} are given in~\cref{sec:contact_searching}.
The numerical simulation results of impulse with different contact locations and directions are given in Section~\ref{sec:sim_results}. 

 \begin{remark}
In this paper, the optimal contact locations are selected to be impact-resistant. However, for tasks that require impact, such as batting~\cite{jia2019batting}, we can also select contact locations that maximise impact.
\end{remark}

\section{Impact-Aware Contact Selection}\label{sec:contact_searching}

% Similar to human beings, in order to intercept and catch a moving object, the robot should first estimate the current state of the object and predict its future trajectory, then further decide whether itself has the ability to catch the moving object and find the corresponding optimal contact time and contact locations. Finally, the optimal catching motion will be generated based on the predicted trajectory of the object and the selected optimal contact locations.

% \textcolor{blue}{In this paper, the unknown state (pose and velocity) of the moving object is estimated by Extended Kalman Filter~\cite{MooreStouchKeneralizedEkf2014} using the \textit{robot localization} repository\footnote{https://github.com/cra-ros-pkg/robot\_localization}. At the same time, the future trajectory of the object is further predicted by using trajectory fitting and solved by nonlinear programming.}
% With the predicted trajectory of the moving object and all 

Based on the predicted object trajectory and the contact selection principles described in Section~\ref{sec:impact_modelling}, here we describe in detail the computational procedure to select optimal contacts (including the contact direction and  location) for catching the moving object, \ie \textit{contact selection} process (see~\cref{sec:prob_desc}).

% or can be measured by Lidar or RGBD camera, 
% the corresponding 3D  mesh can be respectively generated by using PyMesh~\cite{zhou2019pymesh} or Open3D~\cite{zhou2018open3d} libraries. 
\vspace{-2mm}
\subsection{Problem Formulation of Contact Selection}
Assuming that the shape of the object (3D model) is known in advance, we can generate the corresponding 3D  mesh of the object using \textit{PyMesh}~\cite{zhou2019pymesh} or \textit{Open3D}~\cite{zhou2018open3d}. 
The  mesh file includes all the vertices and faces of the object that form the contact surfaces and allows us to obtain the normal vector for each candidate contact location. Given the object mesh, we can search for the suitable contact points $\{\bm p^k\}$ along the object surfaces by solving the following nonlinear-programming problem (NLP):
\textcolor{black}{
\begin{equation}\label{eqn:nonlinear_contact_optimization}
\begin{aligned}
\min_{\bm p^k} \ & \bm e(\bm p^k, \bm n^k, \bm v^o)\\
\textrm{s.t. } \  & \bm p_k \in \mathcal{S} 
\end{aligned}~,~~\forall k \in \{0,...,\mathcal{K}\},
\end{equation}
where $\mathcal{K}$ is the number of robots;} $\bm p^k \in \mathbb{R}^3$; $\mathcal{S}$ represents the face set of the object mesh (surface); $\bm e(\cdot)$ is the cost function that depends on the contact location $\bm p^k$ of the end-effector of robot $k$, and the normal vector $\bm n^k \in \mathbb{R}^3$ at the closest contact location corresponding to $\bm p^k$.  
% \theo{Keep this here as a reminder for potential "verbal" extension of the concept, what about workspace limits ? and time indexing ?}\lei{I think we can not achieve this in our algorithm. Reminder: we have too many reference, like a review paper. I would prefer delete some reference papers from unknown conference and journal.}

% \begin{figure}[t]
% 	\begin{center}
% 		\includegraphics[width=0.95\columnwidth]{figures/contact_searching.png}
% 		\vspace{-10pt}
% 	\end{center}
% 	\caption{Normal vector smoothing and dual-contact point searching.}
% 	\label{fig:contact_searching}
% 	\vspace{-5pt}
% \end{figure}

\begin{figure}[t]
	\begin{center}
		\includegraphics[width=0.90\columnwidth]{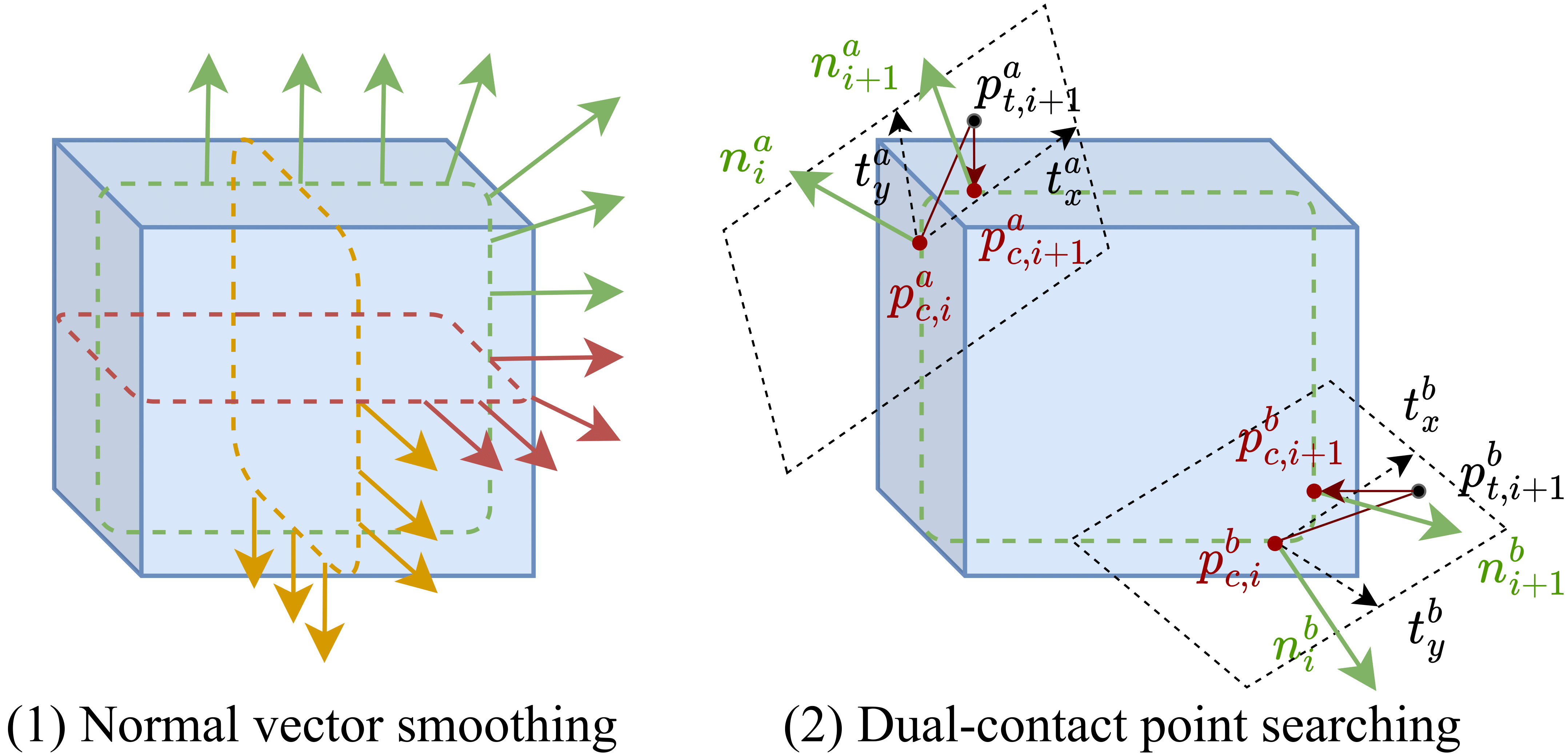}
		\vspace{-10pt}
	\end{center}
	\caption{Normal vector smoothing and dual-contact point searching.}
	\label{fig:contact_searching}
	\vspace{-10pt}
\end{figure}

\paragraph*{Cost function} Our objective is to find the \textcolor{black}{$\mathcal{K}$} contact locations on the surface of the object that minimize impulse and simultaneously provide contact forces such that the motion of the object can be halted. Hence, the proposed cost function is $\bm e(\cdot) = \bm e^\top  \bm e$ with
% \begin{equation}\label{eqn:contact_opt_cost}
% % \bm e(\cdot) = \omega_1 \|\bm n_k\cdot \bm v_o \| + \omega_2 (\| \bm p_k - \bm p_o \| + \| \bm n_k \cdot \bm n_j -1 \|),\\
% \bm e(\cdot) = \omega_1 \|\bm n_k\cdot \bm v_o \|^2 + \omega_2 \| \bm p_k - \bm p_o \|^2 + \omega_3 \sum \| \bm n_k \cdot \bm n_j + cos(\frac{2\pi}{K}) \|^2,
% \end{equation}
\textcolor{black}{
\begin{equation}\label{eqn:contact_opt_cost}
    \bm e^k = \left[ \begin{array}{c} 
w_1 (\bm n^k\cdot \bm v^o )  \\ w_2 ( \bm p^k - \bm p^o )  \\ w_3 \sum_{j} ( \bm n^k \cdot \bm n^j - cos(\frac{2\pi}{\mathcal{K}})) \end{array} \right]^\top 
\end{equation}
for $\forall k \in \{0,...,\mathcal{K}\} $ and $ j \in \{0,...,\mathcal{K}\ | j$ is adjacent to $k\}$.} $\bm p^o, \bm v^o$ are the position and velocity vectors of the center of mass (CoM) of the object. The first and second terms in~\eqref{eqn:contact_opt_cost} 
are respectively designed according to\textit{~\cref{hypo:orientation}} and\textit{~\cref{hypo:position}} in Section~\ref{sec:impact_modelling}, while the third term is designed for improving the catching quality which is typically used in grasping~\cite{roa2015grasp}. 
% \theo{The third term is designed according to the condition of force closure for non-prehensile dual-arm catching---non-prehensile and force-closure are opposing concepts... this terms is kind of a coordination term... apparently we use the same metrics as in quasi-static grasping. Is this good or bad? }
% \lei{I think it is good. Dual-arm catching is a special case of grasping, i.e. two finger grasping. However, here, we are handling the impact and momentum of the object.}

% \begin{figure}[t]
% 	\begin{center}
% 		\includegraphics[width=0.95\columnwidth]{figures/contact_searching.png}
% 		\vspace{-10pt}
% 	\end{center}
% 	\caption{Normal vector smoothing and dual-contact point searching.}
% 	\label{fig:contact_searching}
% 	\vspace{-5pt}
% \end{figure}
\vspace{-2mm}
\subsection{Gradient-based Contact Searching}
The contact point search problem~\eqref{eqn:nonlinear_contact_optimization} (with cost given in~\eqref{eqn:contact_opt_cost}) is a nonlinear constrained optimization problem, which can be solved efficiently via Sequential Quadratic Programming (SQP) following Coordinate Descent (CD)~\cite{wright2015coordinate}. The outline of the method is provided in~\cref{alg:bilevel_sqp}. Given the pose $\bm x^o$ and velocity $\bm v^o$ of the object and the end-effector pose $\{\bm x^{e,k}\}$ of \textcolor{black}{$\mathcal{K}$} robots, the gradient-based contact searching algorithm can find the optimal contact locations $\{\bm p^k\}$.

\paragraph*{Coordinate Descent SQP} 
Optimizing multiple interacting contact points concurrently makes the problem non-convex due to cross-coupling factors, such as the coordination (third) term in~\eqref{eqn:contact_opt_cost}. To accommodate for that, we optimize convex subproblems in a loop (see lines 5-8 in~\textit{\cref{alg:bilevel_sqp})}. Within each iteration we optimize the location of a single contact point $\bm p^k$ and we fix all other contact points $\bm p^j$ -- referred to as \textit{Single Contact Searching QP}. The scheduling can be seen as a cyclic CD where every time we change the contact point $\bm p^k$ that we optimize for, we alter the direction of descent.  

\setlength{\textfloatsep}{10pt}
\begin{algorithm}[t]
\caption{CD-SQP for Multi-Contact Selection}
\label{alg:bilevel_sqp}
\begin{algorithmic}[1]
	\STATE {Initialize: \small{${\bm x}^o$, ${\bm v}^o$, $\{\bm x^{e,k}\}$; $i =1$}}
	% $\bm p_t^a=\bm x_e^a$, $\bm p_t^b=\bm x_e^b$;
    \vspace{1mm}
	\FOR{ \small{$ \left( k=0; k<\textcolor{black}{\mathcal{K}}; ++k \right)$} }
     	\STATE {$\bm p_0^k \leftarrow \bm x^{e,k}$}; {$\bm p_1^k \leftarrow $\small{Proj}$(\bm x^{e,k})$}
    \ENDFOR
    \vspace{1mm}
	\WHILE{ $\bigwedge_{k=\{0,..\textcolor{black}{\mathcal{K}}\}} \left\{\|\bm p^k_i -\bm p^k_{i-1} \| \geqslant \tau \right\}$}
    \vspace{1mm}
    \STATE {\small{$i=i+1$}}
	    	\FOR{ \small{$ \left( k=0; k<\textcolor{black}{\mathcal{K}}; ++k \right)$} }
	    	    \STATE \small{$j = \{0,...,\textcolor{black}{\mathcal{K}}\ | j \neq k\}$}
                \vspace{1mm}
	    	    \STATE {$\bm{\hat{p}}_i^k \leftarrow $\small{SCS-QP}$(\bm p^k_{i-1},\bm n^k_{i-1}, {\bm x}^o, \bm p^j_{i-1},\bm n^j_{i-1})$}
			 %   \hspace{-2mm}
			    \rlap{\raisebox{-2.5mm}{\smash{$\left.\begin{array}{@{}c@{}}\\{}\end{array}\color{black}\right\}%
			    \color{black}\begin{tabular}{c}\footnotesize{Single}\\\footnotesize{Contact}\\\footnotesize{Searching}\end{tabular}$}}}
             	\STATE {$\bm p_i^k \leftarrow$ \small{Proj}$(\bm{\hat{p}}_i^k)$} 
            \ENDFOR
            \vspace{1mm}
			\IF{\small{$(i >  max\_iter)$}}
				\STATE {\small{$break$}}
			\ENDIF
		
	\ENDWHILE
\RETURN {$\{\bm p_i^k\}, \forall k \in \{0,...,\textcolor{black}{\mathcal{K}}\}$} 
\end{algorithmic}
\end{algorithm}

% \begin{algorithm}[t]
% \caption{CD-SQP for Dual Contact Searching}
% \label{alg:bilevel_sqp_bk}
% \begin{algorithmic}[1]
% 	\STATE {Initialize: $\hat{\bm x}_o$, $\hat{\bm v}_o$, $\bm x_e^k$; $\bm p_t^a=\bm x_e^a$, $\bm p_t^b=\bm x_e^b$; $i =1$}
% 	\STATE {$\bm p_c^a(0) \leftarrow Projection(\bm p_t^a)$; $\bm p_c^b(0) \leftarrow Projection(\bm p_t^b)$}
% 	\WHILE{$(i\leqslant max\_iterations)$ }
% 			\STATE {$\bm p_t^a \leftarrow DualCntQP(\bm p_c^a,\bm n_c^a, \hat{\bm x}_o, \bm p_c^b,\bm n_c^b)$}
%   			\STATE {$\bm p_c^a(i) \leftarrow Projection(\bm p_t^a)$}
% 			\STATE {$\bm p_t^b \leftarrow DualCntQP(\bm p_c^b,\bm n_c^b, \hat{\bm x}_o, \bm p_c^a,\bm n_c^a)$}
% 			\STATE {$\bm p_c^b(i) \leftarrow Projection(\bm p_t^b)$}
% 			\IF{$\bigwedge_{k=a,b} \left\{\|\bm p_c^k(i) -\bm p_c^k(i-1) \| \leqslant threshold \right\}$}
% 				\STATE {$break$}
% 			\ELSE{}   \STATE {$i=i+1$}
% 			\ENDIF
% 	\ENDWHILE
% \RETURN {$\bm p_c^a(i)$, $\bm p_c^b(i)$}
% \end{algorithmic}
% \end{algorithm}
% The contact searching problem~\eqref{eqn:nonlinear_contact_optimization} is a nonlinear constrained optimization problem, which could be solved using Nonlinear-Programming (NLP). 
% % However, the computation time can not guarantee online optimization of contact location.
% Alternatively, in this paper Sequential Quadratic Programming (SQP) is adopted to solve this dual contact point optimization problem.
% generalized voxels information, 
% which can be used to calculate the contact surface, contact location and normal vector at the contact location. 
\paragraph*{Single Contact Searching QP} Each subproblem is responsible for optimizing the location of a contact point, while keeping it on the surface ($\bm p^k \in \mathcal{S}$). This can be expressed via non-linear equality constraints, \eg ${\mathcal{S}} (\bm p^k) = 0$. The challenge here is that for closed surfaces the constraint ${\mathcal{S}}(\cdot) $ is non-linear and should to be differentiable (for a gradient-based solver). Although, there are potential approaches using non-linear solvers~\cite{stouraitis2018dyadic, driess2022learning}, we adopt ideas from~\cite{murooka2020optimization} to perform a series of linear steps along consecutive tangential planes of the surface based on smoothed normal vectors, %~\cite{jones2003non, murooka2020optimization}, 
\ie we iteratively linearize ${\mathcal{S}}(\cdot) $. This allows our method to work with arbitrary number of contact points and meshes (surfaces). As shown in~\textit{\cref{alg:bilevel_sqp}}, each linear step is realized by solving a \textit{Single Contact Searching QP} subproblem (line 8) and it's followed by a projection (line 9) back to the surface of the object. The series of linear steps are performed until convergence for all the contact points (see lines 4-11).

\paragraph*{Contact Location Iteration} 
Given a contact location $\bm p_{i-1}^k \in \mathcal{S}$ at iteration $i-1$, the $i_{th}$ QP subproblem aims to find the optimal displacement $\Delta \bm  p^k $ of the contact point $\bm p^k$ within a tangential plane  (see~\cref{fig:contact_searching}). The $i_{th}$ tangential plane is defined based on the contact location $\bm p_{i-1}^k$ and the corresponding normal vector $\bm n_{i-1}^k$. The $i_{th}$ contact point within the tangential plane is $ \hat{\bm p}_{i}^k = \bm p_{i-1}^k + \Delta \bm p_{i-1}^k$
and the optimal contact point $\bm p^i_k$ on the object's surface is obtained by projecting $\hat{\bm p}_{i}^k$ on the surface of the object. 
% The outline of the QP subproblem is given in~\cref{alg:contact_searching_qp}

The actual QP subproblem formulation is:
\begin{equation}\label{eqn:contact_searching_QP}
\begin{aligned}
\min_{\Delta \bm p^k} \quad  &\frac{1}{2} (\bm J^k\Delta \bm  p^k)^\top (\bm J^k \Delta \bm p^k) + \bm e^k \bm J^k \Delta \bm p^k  \\
\textrm{s.t.} \quad & \Delta \bm p_{min} \leqslant \Delta \bm p^k \leqslant  \Delta \bm p_{max},
\end{aligned}
\end{equation}
where %$\bm J^k = \frac{\partial \bm e^k}{\partial \Delta \bm p_k}$ is 
the Jacobian of contact location optimization~\eqref{eqn:contact_searching_QP} is:
\begin{equation}
    \bm J^k = \frac{\partial \bm e^k}{\partial \Delta \bm p^k} = \left( \begin{array}{c} 

    \omega_1 \bm v^{o,\top} \frac{\partial \bm n^k}{\partial \Delta \bm p^k} \\
    \omega_2 \frac{\partial \bm p^k}{\partial \Delta \bm p^k}  \\ 
    \omega_3 \bm n^j \frac{\partial \bm n^k}{\partial \Delta \bm p^k} \end{array} 
    \right),
\end{equation}

\begin{equation}
    \frac{\partial \bm p^k}{\partial \Delta \bm p^k} = \left( \bm t_x^{k}, \quad \bm t_y^{k} \right),
\end{equation}

\begin{equation}
    \frac{\partial \bm n^k}{\partial \Delta \bm p^k} = \left(\frac{ \bm n^k |_{+x}-\bm n^k |_{-x}}{2 \delta},\  \frac{ \bm n^k |_{+y}-\bm n^k |_{-y}}{2 \delta} \right),
\end{equation}
where $\bm n^k |_{\pm {x}}\left(\bm n^k |_{\pm{y}} \right)$
is the normal vector of the contact surface at contact location $\bm p^k$ when $\Delta p_{x}(\Delta p_{y})$ is $\pm \delta$. 

\textcolor{black}{The outline of the single contact searching method is provided in~\textit{\cref{alg:contact_searching_sqp}}.  % \paraDraft{About normal smoothing} 
% normal smoothing~\cite{}  
Also, note that as shown in Fig.~\ref{fig:contact_searching} (1), the normal vectors on the surfaces of the object are smoothed to ensure that the contact location $\bm p^k$ can be optimized across all the noncontinuous object surfaces.}
% \textcolor{blue}{\eqref{eqn:contact_opt_cost} can be considered a trust-region method with fixed size, which allows  to limit the displacement in the tangent plane.
% Also, the normal vectors on the object surface are smoothed (see~\cref{fig:contact_searching}(1)) to enable more suitable approximation with tangent planes. The two properties above ensure the convergence to local minima.}
% ensure that the projection on the object surface is possible. 

\setlength{\textfloatsep}{10pt}
\begin{algorithm}[t]
\caption{QP for Single Contact Searching ($SCS-QP$)}
\label{alg:contact_searching_sqp}
\begin{algorithmic}[1]
	\STATE {Initialize: $\bm p^k_{i-1}$, $\bm n^k_{i-1}$, ${\bm x}^o$, $\bm p^{\lnot k}_{i-1}$, $\bm n^{\lnot k}_{i-1}$}
	\STATE {$\bm J_{pos}, \bm J_{ori},\bm J_{coo} \leftarrow {\partial \bm e^k_{i}} / {\partial \Delta \bm p^k}$}
	\STATE {Jacobian matrix: $\bm J^k \leftarrow \left[\bm J_{pos}, \bm J_{ori},\bm J_{coo} \right]^\top$}
	\STATE {$\bm Q \leftarrow \bm J^{k,\top} \bm J^k; \bm q \leftarrow \bm J^{k,\top} \bm e^{k}_{i}; \bm {lb}, \bm {ub} \leftarrow  \Delta \bm p_{min}, \Delta \bm p_{max}$}
	\STATE  {$\Delta \bm p^k \leftarrow QP\_solver(\bm Q, \bm q, \bm {lb}, \bm {ub})$}
	\STATE  {$\hat {\bm p}^k_i \leftarrow \bm p^k_{i-1} +\Delta \bm p^k $}
\RETURN {$\hat{\bm p}_i^k$}
\end{algorithmic}
\end{algorithm}

% \paraDraft{QP} 

% As shown in Fig.~\ref{fig:contact_searching}, in each step of SQP, the value of $\Delta p_x^k$ and $\Delta p_y^k$ are optimized toward to the optimal contact point as a QP problem:  

%~\cite{jones2003non, murooka2020optimization}.

% and can be calculated as follows:

% \begin{equation}
%     \bm J_k = \frac{\partial \bm e^k}{\partial \Delta \bm p^k} = \left( \begin{array}{c} 
% \omega_1 \frac{\partial \bm p_t^k}{\partial \Delta \bm p^k}  \\ \omega_2 \bm v_o^\top \frac{\partial \bm n_c^k}{\partial \Delta \bm p^k} \\ \omega_2 \bm n_c^{k,T} \frac{\partial \bm n_c^{\lnot k}}{\partial \Delta \bm p^k} \end{array} 
%     \right)
% \end{equation}

% during the optimization of the optimal contact location , 

% \theo{write the closing paragraph for the section so we can transit to the next one}
By solving \textit{contact selection problem} using the CD-SQP, the proposed method is able to compute the optimal contact locations $\bm p^k$ and normals $\bm n^k$ %on the surface of 
on the moving object, which are used in the \textit{impact-aware planning} process (see~\cref{sec:multi_mode_optimization}).

\section{Multi-Mode Trajectory Optimization}\label{sec:multi_mode_optimization}
\label{sec:MMTO}
In this section, we describe the proposed motion planning method which is used to optimize the capturing motion, \ie \textit{the impact-aware planning} process. A dual-arm catching motion comprises of a variety of modes, such as free motion, motion in contact, free single-arm motion, constrained dual-arm motion, and compliant or stiff behaviour in interaction. In general, systems that transit through different modes are typically referred as hybrid. \textcolor{black}{These are typically not causal with respect to the state and control, hence the notion of mode is introduced. Specifically, for transitions from \textit{free-motion} to motion \textit{in-contact} and vice-versa 
we denote the dynamics mode~\cite{goebel2009hybrid} with $\mathcal{M}^d \in \mathbb{Z}$. For transitions from one controller to the other, such as from stiff to soft behaviour and vice-versa 
we denote the control mode with $\mathcal{M}^c \in \mathbb{Z}$. 
For transitions from single-arm free motion to dual-arm coordinated motion and vice-versa
we denote the planning mode with $\mathcal{M}^p \in \mathbb{Z}$. }

Our goal is to optimize a single trajectory that includes transitions from one mode to the other.  To this end, we define a single combination of contact mode, planning mode and corresponding controller as a mode tuple of the system~\cite{stouraitis2020multi} $\mathcal{M} = (\mathcal{M}^d, \mathcal{M}^c, \mathcal{M}^p)$. 
\textcolor{black}{Given a sequence of these modes $\bm{\mathcal{M}}: \{ \mathcal{M}_0, \mathcal{M}_1, ... \mathcal{M}_N \}$ of the trajectory,} we express the Multi-Mode Trajectory Optimization (MMTO) as follows:
% a function of the initial state~$\bm{x}(0)$ and the plant's control actions:
% \begin{IEEEeqnarray}{CCCCC}
%     \IEEEyesnumber\label{eq:csDSTO} 
%     \IEEEyessubnumber \label{eq:csDSTO_cost}
%     \min_{\bm{x}(t), \bm{u}(t), \bm{v}(t)}  &  \bm{c}\left(\bm{x}(t), \bm{a}(t), \bm{u}(t), {\mathcal{M}} \right)~ \\
%     ~\text{s.t.}~  
%     \IEEEyessubnumber \label{eq:csDSTO_system_model} &
%     \dot{\bm{x}}(t) = \bm{f} \left(\bm{x}(t), \bm{a}(t), \bm{u}(t), {\mathcal{M}} \right),~\hfill\\
%     \IEEEyessubnumber \label{eq:csDSTO_control_model} & 
%     \dot{\bm{v}}(t) = \tilde{\bm{h}} \left(\bm{a}(t), {\mathcal{M}} \right), \hfill\\
%     \IEEEyessubnumber \label{eq:csDSTO_constraints} & 
%     \bm{g}(\bm{x}(t), \bm{a}(t), \bm{u}(t), {\mathcal{M}}) \leq 0, \hfill
% \end{IEEEeqnarray}
\begin{IEEEeqnarray}{CCCCC}
    \IEEEyesnumber\label{eq:csDSTO} 
    \IEEEyessubnumber \label{eq:csDSTO_cost}
    \min_{\bm{x}, \bm{a}, \bm{u}}  &  \bm{c} \left(\bm{x}, \bm{a}, \bm{u}, \bm{\mathcal{M}} \right)~ \\
    ~\text{s.t.}~  
    \IEEEyessubnumber \label{eq:csDSTO_system_model} &
    \bm{x}_{n+1} = \bm{f} \left(\bm{x}_n,  \bm{a}_n, \bm{u}_n, \mathcal{M}_n \right),~\hfill\\
    \IEEEyessubnumber \label{eq:csDSTO_control_model} & 
    \bm{a}_{n+1} = \bm{h} \left(\bm{a}_n, \mathcal{M}_n \right), \hfill\\
    \IEEEyessubnumber \label{eq:csDSTO_constraints} & 
    \bm{g}(\bm{x}_{n}, \bm{a}_n, \bm{u}_n, \mathcal{M}_n) \leq 0, \hfill
\end{IEEEeqnarray}
where $\forall n \in \{0,...,N\}$; $\bm{x}_n$ %$\bm{x}_n\in \mathbb{R}^o$ 
is the state of the system, $\bm{a}_n$ 
%$\bm{a}_n \in \mathbb{R}^m$ 
is the control attributes of the robot (\eg robot stiffness) and $\bm{u}_n$ %$\bm{u}_n \in \mathbb{R}^o$ 
is the control input applied on the environment (\eg force applied on the object). 
\eqref{eq:csDSTO_cost} - \eqref{eq:csDSTO_constraints} are piece-wise functions from which the corresponding pieces are selected based on $\mathcal{M}_n$. \eqref{eq:csDSTO_cost} defines the objective function, \eqref{eq:csDSTO_system_model} and \eqref{eq:csDSTO_control_model} are difference functions (discretize differential functions) and~$\bm{g}(\cdot)$ in \eqref{eq:csDSTO_constraints} describes both the equality and the inequality constraints. 
Note that the formulation above defines an Optimal Control problem where both dynamics \eqref{eq:csDSTO_system_model} and control \eqref{eq:csDSTO_control_model} are hybrid. Unless otherwise specified, all the variables are represented in the world coordinate system.

% To solve the continuous optimization problem in \eqref{eq:csDSTO}, we discretize and represent the trajectory by $N+1$ collocation knots according to direct transcription~\cite{rawlings2017model}.

% , we discretize and represent the trajectory by $N+1$ 
% , betts2010practical}.
% kelly2017introduction,  

The discretized optimization problem in~\eqref{eq:csDSTO} involves $N-1$ collocation knots\footnote{Knots are the discretization points of the transcribed continuous problem.} using direct transcription~\cite{kelly2017introduction}. The transcription of MMTO is based on the phase-based parameterization used in our previous work~\cite{stouraitis2020multi, stouraitis2018dyadic} and is similar in spirit to~\cite{toussaint2018differentiable}. For each $n_{th}$ knot, the decision variables are (i) the pose of the object $\bm{x}_n^o$, (ii) the velocity of the object $\bm{\dot{x}}_n^o $,  (iii) action timings $\Delta \bm{T}_n$, (iv) the end-effector's position $\bm{x}_n^e$ and velocity $\dot{\bm{x}}_n^e$, (v) the contact force $\bm{F}_n$ and a parameter $ \bm{\alpha}_n$ (described in detail in the following section). We group these quantities into three vectors 
% vi) the rate of change of the contact force $\mathbf{\dot{f}}_i$ , vii) the second derivative of the contact force $\mathbf{\ddot{f}}_i$ 
% and \eqref{eq:optPolSpec},
% which we group 
% \begin{equation}\label{eq:decVec}
%     \bm{s}_i = \begin{bmatrix}\mathbf{y}_i & \mathbf{\dot{y}}_i & \mathbf{c}^k_i & \mathbf{\dot{c}}^k_i & \mathbf{f}^k_i & \mathbf{\Delta T}_i \end{bmatrix}^T,
% \end{equation}
\begin{IEEEeqnarray}{C}
    \IEEEyesnumber\label{eq:OptVariables}
    \bm{x}_n = \begin{bmatrix}\bm{x}_n^o & \bm{\dot{x}}_n^o & \bm{x}_n^e & \bm{\dot{x}}_n^e  \end{bmatrix}^\top_, \\
    \bm{a}_n = \begin{bmatrix} \bm{\alpha}_n & \Delta \bm{T}_n \end{bmatrix}^\top_, \\
    \bm{u}_n = \begin{bmatrix} \bm{F}_n & \bm{\dot{F}}_n  \end{bmatrix}^\top_,
\end{IEEEeqnarray}
% where $\forall i \in  \mathbb{N}$, 
% with the trajectories of $\bm{x}_n$, $\bm{a}_n$ and $\bm{u}_n$ 
that all together describe a multi-mode motion, which is tailored via the mode sequence ${\bm{\mathcal{M}}}$.
%according to $\bm{z}$.
This results in a TO problem with different constraints for each one of the modes.
Next, we describe the constraints that form~\eqref{eq:csDSTO_system_model} - \eqref{eq:csDSTO_constraints} according to the different modes.  

% \begin{figure}[t]
% 	\begin{center}
% 		\includegraphics[width=0.99\columnwidth]{figures/manipulation_modes.pdf}
% 		\vspace{-10pt}
% 	\end{center}
% 	\caption{Manipulation modes for hybrid dynamics, planning and control.}
% 	\label{fig:manipulation_modes}
% 	\vspace{-10pt}
% \end{figure}

\vspace{-2mm}
\subsection{Hybrid Motion Equations for the Object and Robots}
Both the object's and robots' motions depend on the current and the subsequent dynamics modes. Next, we present the models used to describe the behaviour of the object and the robots with respect to the dynamics modes, \ie free-motion mode ($\mathcal{M}^d=0$) and contact mode ($\mathcal{M}^d=1$).

\subsubsection{\underline{Object's dynamics}}
% As discussed at the beginning of this section, when 
The object dynamics in free motion are denoted with $\mathcal{M}^d=0$, while when $\mathcal{M}^d=1$, the object is in contact. According to this parametrization the dynamics of the object are described by
% \begin{align}
% \label{eq:objDyn}
% \begin{bmatrix} \bm{y}_{n+1},  \bm{\dot{y}}_{n+1}\end{bmatrix} =     \left\{ \begin{array}{ll} 
%         {f}^o(\bm{y}_n, \Delta \bm{T}_n), & \text{if $\mathcal{M}^d=0$},\\
%         {f}^o(\bm{y}_n, \Delta \bm{T}_n, \bm{F}_n), & \text{if $\mathcal{M}^d=1$},
%         \end{array} \right.
% %   \bm{f}_o(\mathbf{x}_o^i, \mathbf{\dot{x}}_o^i, \mathbf{r}^i, \Delta \mathbf{T}^i, \mathbf{f}^i, \mathbf{z}^i),
% \end{align}
\begin{align}
\label{eq:objDyn}
\begin{bmatrix} \bm{x}_{n+1}^o,  \bm{\dot{x}}_{n+1}^o\end{bmatrix} =     \left\{ \begin{array}{ll} 
        \bar{f}^o(\bm{x}_n^o, \Delta \bm{T}_n), & \text{if $\mathcal{M}^d=0$},\\
        {f}^o(\bm{x}_n^o, \Delta \bm{T}_n, \bm{F}_n), & \text{if $\mathcal{M}^d=1$},
        \end{array} \right.
%   \bm{f}_o(\mathbf{x}_o^i, \mathbf{\dot{x}}_o^i, \mathbf{r}^i, \Delta \mathbf{T}^i, \mathbf{f}^i, \mathbf{z}^i),
\end{align}
\noindent where \textcolor{black}{$\bar{f}^o(\cdot) : \mathbb{R}^{o+1} \rightarrow \mathbb{R}^{2 \cdot o}$, $f^o(\cdot) : \mathbb{R}^{2 \cdot o + 1} \rightarrow \mathbb{R}^{2 \cdot o}$ with $f^o(\cdot)$ (\eqref{eq:objDyn} for $\mathcal{M}^d=1$)} in detail being
\begin{equation}
    \label{eq:FullDyn}
    \begin{bmatrix} m \bm E^3 & 0 \\ 0 & \bm I \end{bmatrix} \bm{\ddot{x}}_n^o + \begin{bmatrix} m \bm g \\ \bm{\omega}_n \times (\bm I \bm{\omega_{n}}) \end{bmatrix} = \begin{bmatrix} \bm{F}_n \\ {\bm{r}}_n \times \bm{F}_n \end{bmatrix}  .
\end{equation} 
% where $ m_o \in \mathbb{R}$  and $\bm I_o \in \mathbb{R}_{\ge 0}^{\nu \times \nu}$ are the mass and inertia of the object, \(\bm E_n\) is the $n \times n$ identity matrix, $\bm g$ is the acceleration due to gravity, $\mathbf{\dot{x}}^i_{o,\omega}$ is the object's angular velocity, and we refer to the cross product matrix formed by the input vector with \(\hat{(\cdot)}\). 
\eqref{eq:FullDyn} makes the hybrid nature of the system's dynamics evident. With $\mathcal{M}^d=1$ the right hand side (RHS) of~\eqref{eq:FullDyn} remains while with $\mathcal{M}^d=0$, RHS of \eqref{eq:FullDyn} disappears.  The orientation of the object is represented with $ZYX$ Euler angles, hence, the angular velocity is first converted to the derivative of Euler angles and then it is integrated to the $ZYX$ Euler angles.

% \theo{Here, we need to say a few words, one sentence or so about the parametrisation that we used to represent the angular part of the objects pose, velocity .}

% \lei{Done for the object and robot together at the beginning of this subsection.}

\subsubsection{\underline{Robot's motion}}
% When planning robot motions with impacts, particular care needs to be taken while enforcing the integration constraints of the motion \cite{goebel2009hybrid}. The motion of the end-effector is described with the following function 
A key characteristic of motions with velocity jumps---such as impacts---is that the integration from accelerations to velocities needs to be skipped at specific mode transitions~\cite{goebel2009hybrid, rijnen2017control}.
% Hence, the motion of the robots' end-effectors is described with the following:
\textcolor{black}{Hence, with the following:
\begin{equation}
\label{eq:eeMotion} 
\hspace{-3mm} \begin{bmatrix} \bm{x}_{n+1}^e \\ \bm{\dot{x}}_{n+1}^e\end{bmatrix} \hspace{-1mm} = \hspace{-1mm} \left\{ \begin{array}{l} 
                \hspace{-2mm}f^e(\bm{x}_n^e, \bm{\dot{x}}_n^e, \bm{\ddot{x}}_n^e, \Delta \bm{T}_n), \text{ if $\mathcal{M}^d_{n} = \mathcal{M}^d_{n+1} = 0$ or $1 $}~\\
                \hspace{-2mm}{\bar{f}^e(\bm{x}_n^e, \bm{\dot{x}}_n^e, \Delta \bm{T}_n)}, \text{ if $\mathcal{M}^d_n=0,$ $\mathcal{M}^d_{n+1}=1$}~\\
                \hspace{-2mm}{\bar{f}^e(\bm{x}_n^e, \bm{\dot{x}}_n^e, \Delta \bm{T}_n)}, \text{ if $\mathcal{M}^d_n=1,$ $\mathcal{M}^d_{n+1}=0$}~ 
                \end{array} \right . \hspace{-12mm} ,
\end{equation}
we describe the motion of the robots' end-effectors, where $f^e(\cdot) : \mathbb{R}^{3 \cdot o + 1} \rightarrow \mathbb{R}^{2 \cdot o}$, $\bar{f}^e(\cdot) : \mathbb{R}^{2 \cdot o + 1} \rightarrow \mathbb{R}^{2 \cdot o}$ and both are integration functions.
The transitions in~\eqref{eq:eeMotion}} occur during the making and the breaking of contact, and they are susceptible to impact. We denote these as $\mathcal{M}_n^d~=~0 \rightarrow \mathcal{M}_{n+1}^d~=~1$ and $\mathcal{M}_n^d = 1 \rightarrow  \mathcal{M}_{n+1}^d = 0$, respectively. \textcolor{black}{By omitting $\ddot{\bm{x}}_n^e$ in $\bar{f}^e(\cdot)$, accelerations are free, the velocity jumps are possible. Hence, the solution space of~\eqref{eq:csDSTO} includes discrete events, typically described as jumps in hybrid systems literature~\cite{goebel2009hybrid}.}

% Hence, the solution space of the mathematical program (see~\eqref{eq:csDSTO}) includes discrete events, typically described as jumps in hybrid systems literature~\cite{goebel2009hybrid}. 
% Note that we need to omit the time integration of the robot, but not the one of the object.   
% The accelerations of the robot have to be bounded---which constrains velocity changes (curb velocity jumps)---according to the capabilities of the robot, we need to omit the time integration of the robot. ??????
% On the other hand, the forces in~\eqref{eq:objDyn} are unbounded (can trigger velocity jumps), hence there is no need to omit the time integration of the object.   

% \lei{about the simplified impact model added here}

% \subsection{Contact force during contact transitions} 
\vspace{-2mm}
\subsection{Control Modes during Contact Transitions}
\label{subsec:controlModes}
Similar to the dynamics modes ($\mathcal{M}^d$), here, we describe control modes ($\mathcal{M}^c$), which are relevant to mitigate impact in highly dynamic manipulation that involve contact transitions. 

% \subsubsection{\underline{Control modes for contact transitions}}
% \paraDraft{How do the different modes of control arise for this task}
Based on the relation between an impact event and the behaviour of a spring-damper system (see~\cref{subsec:3DcomplianceModel} and~\cref{fig:impact_model}), the characteristics of the physical system during impact, such as the duration of impact and the restitution coefficient\footnote{The restitution coefficient value $\epsilon_r=1$ represents a perfectly elastic collision,~$0\leq\epsilon_r\leq 1$ represents a real-world inelastic collision. ~$\epsilon_r=0$ represents a perfectly inelastic collision.}, can be associated to the mass, damping and stiffness parameters of a mechanical system~\cite{nagurka2004mass}. %, zhu1999theoretical}.
As it can be seen in~\cref{fig:impact_model}, an impact event can be divided into two stages: (i) the compression stage where the relative velocity is negative and  the contact force establishes a stable contact, and (ii) the restitution stage where the relative velocity is positive and the contact force yields the desired manipulation task. We encode these two stages with two different control modes %during contact as 
% We utilize this observation to accurately emulate the physical interaction through the impedance controller of the manipulator. 
% where stiffness and damping parameters can 
% Therefore, the physical interaction can be simulated by admittance control of the manipulator with different stiffness and damping parameters. 
% As shown in Fig.~\ref{fig:impact_model}, the negative contact is defined as the deformation stage during which the contact force for making a stable contact is generated. The positive contact is defined as the restitution stage which  generates the contact force for the manipulation tasks, such as pushing an object away. We encode these different modes of contact as
\begin{equation}\label{eqn:contact_phases}
    \mathcal{M}^c = \left\{ \begin{array}{cc}
        -1, & \bm{{v}}^- \rightarrow 0\\
        1, & 0 \rightarrow \bm{{v}}^+  
        % 0, & \text{no contact}.
    \end{array}\right. 
    .
\end{equation}
For dynamic motions with contact, during the compression stage (soft mode) the stiffness should be minimized to alleviate impulsive forces, while during the restitution stage (stiff mode) the stiffness should be optimized to achieve the desired motion.

% In terms of impact-aware manipulation, the stiffness should be minimized during negative contact to alleviate impact while it should be maximized during positive contact, to achieve accurate manipulation. 
% % These contact modes are encoded in \eqref{eq:csDSTO} in Section~\ref{sec:multi_mode_optimization} in the form of different controller parameters. 
% In this way, the controller parameters (stiffness) are optimized to conform with the different stages of the impact. 

\vspace{-2mm}
\subsection{Contact Force Transmission during Impact}
% \subsubsection{\underline{Contact Force Transmission Model}}
\label{subsec:cntforceTrans}

Based on the control modes introduced above, here, we devise a force transmission model to conform with the different stages during an impact event. 
The proposed force transition model is used in~\eqref{eq:csDSTO} to model the force evolution during an impact event and is a simplified version of the energy-based 3D impact model~\cite{jia2019batting} (see~\cref{subsec:3DcomplianceModel}). 
% that allows us to optimize the controller parameters (stiffness) 
% The proposed force transition model that we use in~\eqref{eq:csDSTO} to model the force evolution during an impact event, is a simplified version of the energy-based 3D impact model~\cite{jia2019batting} (see~\cref{subsec:3DcomplianceModel}). 

We adopt the simplified impact model mainly because of the following reasons. First,  our model is practical and can be embedded in a TO formulation like~\eqref{eq:csDSTO}, with which its parameters can be optimized (versus only forward integration). Second, our model considers the contact force only along the normal direction of the contact surface (versus all three), as it is the only one that can be actively controlled. Third, it allows us to describe the force evolution as a function of stiffness and time (versus the impulse increment $\Delta \Lambda_z$).

Yet, to devise a force model for impacts that is time dependent and can be used to plan the evolution of the contact force during impact, we make the following assumption. 

% first, we assert that if the distance between two surfaces along the normal direction is constant (sliding is permitted), then the contact is stable  and second, we make the following assumption. \lei{I don't see the relationship between the force transmission model and the assumptions.}

\begin{assumption}
The duration of an impact event can be prolonged by reducing the contact force magnitude during the compression stage.
% and restitution stages.

% The duration of an impact event can be prolonged by artificially control the magnitude of the contact force in compression and restitution stages.

% The duration of an impact event can be prolonged by reducing the relative velocity
% between two surfaces.

% if the relative velocity between two surfaces is negative and small. 
% The duration of an impact event can be prolonged if the relative velocity between two surfaces remains negative and can be downsized. 

% For making contact between two surfaces in non-prehensile manipulation, a stable contact can be only established along the normal direction of contact surfaces. 
% We make the assumption here if we can prolong the duration of an impact event, then we can control the evolution of the force along the normal direction. 

% For making contact between two surfaces in non-prehensile manipulation, a stable contact can be only established along the normal direction of contact surfaces. \theo{Does this simply mean that the only assumption that we make here is that the object is not going to bounce off? --- this allows us to consider only the effects of the impulse along the normal direction}
% We make the assumption here if we can prolong the duration of an impact event, then we can control the evolution of the force along the normal direction. 
\end{assumption}

\subsubsection{\underline{Contact force transmission model}}

% To guarantee the planned contact force is smooth without oscillations, the contact force transmission model in MMTO is defined as a second-order critically damped dynamical system (cd-DS). 
% in the hybrid TO. 
% A cd-DS~\cite{braun2013robots} was first used %for motor servo control 
% to generate the tractable motor position, and further used to guarantee constrained consistent output for any admissible input.

% In this paper, 
% In addition, 
\textcolor{black}{The ideal contact force profile should be smooth without oscillations. From~\eqref{eqn:admittance_transfer_function}, we can formulate the normal force transmission model as a second-order critically damped dynamical system (cd-DS)~\cite{braun2013robots} to enforce the smoothness of contact force. By setting ${\alpha} = \sqrt{{K}/{M}}$ %and $F(t)=K\Delta x$ 
%(given that $K {\Delta x} = {F}(t), {\Delta \dot{x}} \approx 0$ and ${\Delta \ddot{x}} \approx 0$ at the equilibrium point of \eqref{eqn:impact_force_model} when $F= {F}(t)$), 
\textcolor{black}{ and assuming the magnitude of the force is dominated by the stiffness and displacement components considering ${\Delta \dot{x}} \approx 0$ and ${\Delta \ddot{x}} \approx 0$, i.e. $F(t) \approx K \Delta x (t)$,~\eqref{eqn:admittance_transfer_function} becomes}:
\begin{equation}\label{eqn:cdDS_transfer_function}
{{F}}(t) = \frac{ {\alpha}^2}{s^2+2 {\alpha} s + {\alpha}^2}   {{F}},
\end{equation}
and the second-order cd-DS in time-domain:
\begin{equation}\label{eqn:critically_damped_differential_equation}
\ddot {{F}}(t) + 2 \alpha \dot {F}(t) + \alpha^2 {{F}(t)} = \alpha^2 {{F}},
\end{equation}
\textcolor{black}{where ${F(t)}$ is continuous, satisfying ${F(t)} \in [0, {F}]$ and it is the magnitude of the normal component (with respect to contact surface) of $\bm F_n$ in \eqref{eq:FullDyn}}, 
%satisfying ${F(t)} \in [0, {F}]$, 
while ${\dot{{F}}}$ and ${\ddot{{F}}}$ are its first and second derivatives. 
\textcolor{black}{$F$ represents the optimised force at the end of each control mode and it is not an explicitly variable in the optimisation~\eqref{eq:csDSTO}.}
% (cd-DS parameter) 
% this 
For any $ {\alpha} >0$, the contact force transmission model is critically damped, therefore it will converge to its equilibrium point asymptotically in the shortest possible time without oscillation.}

Given any two out of three from; the stiffness parameter $K$, the mass parameter $M$, and the cd-DS parameter $\alpha$, we can determine the other one through ${\alpha} = \sqrt{{ K}/{M}}$ and the damping parameter via $B = 2\sqrt{M K}$. This leads to: %to form the cd-DS model.  
\begin{remark}
\label{rem:alpha2stiff}
Stiffness parameter $K$ %of~\eqref{eqn:impact_force_model} 
is related to parameter $\alpha$.
\end{remark}

% Considering the quasi-static problem at each node where $\Delta \ddot{\bm c}_i \approx 0$ and $\Delta \dot{\bm c}_i \approx 0$, we can have $k \Delta \bm c_i \approx \bm f_i$. Therefore, given the desired contact force ${\bm f}_d$, the contact force generation model can be written as a second-order differential equation:
% \begin{equation}\label{eqn:impact_force_model}
%     \frac{m}{k} \ddot{\bm f}_i + \frac{b}{k} \dot{\bm f}_i + {\bm f}_i = {\bm f}_d
% \end{equation}

% In order to make 
In addition to the smoothness enforced by the cd-DS~\eqref{eqn:critically_damped_differential_equation} on the contact force transmission model, from the differential equation~\eqref{eqn:critically_damped_differential_equation}
we can derive a relationship between the settling time $t_s$ and $\alpha$ (within 5\%, $t_s \approx \frac{3.0}{\omega \mathrm{\zeta}} = \frac{3.0}{\alpha}$). This is the same as settling time of the spring-damper system~\eqref{eqn:impact_force_model}, which reveals:
\begin{remark}
\label{rem:impdur2alpha}
Impact duration is related to parameter $\alpha$.
\end{remark}

% Further, given that $K {\Delta p} = {F}(t), {\Delta \dot{p}} \approx 0$ and ${\Delta \ddot{p}} \approx 0$ at the maximum deformation point of \eqref{eqn:impact_force_model}, the
% % Through 
% Laplace transform of \eqref{eqn:impact_force_model} can be written as
% % we can obtain the following transfer functions
% \begin{equation}\label{eqn:admittance_transfer_function}
% {\Delta p} = \frac{1}{ M s^2+  B s+  K} {F}_d  = \frac{\frac{K}{M}}{ s^2+  \frac{B}{M} s+ \frac{K}{M}} \frac{{F}_d}{K},
% \end{equation}
% and the Laplace transform of~\eqref{eqn:critically_damped_differential_equation} is
% \begin{equation}\label{eqn:cdDS_transfer_function}
% {{F}}(t) = \frac{ {\alpha}^2}{s^2+2 {\alpha} s + {\alpha}^2}   {{F}}_d.
% \end{equation}
% Hence, the following relations between parameter $\alpha$ and the 
% % mass, damping and stiffness 
% parameters of the contact model~\eqref{eqn:impact_force_model} can be obtained:
% \vspace{1mm}

% % \noindent
% \begin{minipage}{.45\linewidth}
% \begin{equation}\label{eqn:mk2alpha}
%     {\alpha} = \sqrt{\frac{ K}{M}},
%     \end{equation}
% \end{minipage}%
% \begin{minipage}{.5\linewidth}
% \begin{equation}\label{eqn:mk2b}
%     B = 2\sqrt{M K}.
%     \end{equation}
% \end{minipage}
% \vspace{1mm}

% \subsubsection{Critically damped system for contact force}
\subsubsection{\underline{Hybrid contact force constraint}}
The forces of the MMTO (see~\eqref{eq:csDSTO}) are generated based on the contact force transmission model described above. Yet, as it has been discussed in~\cref{subsec:controlModes} there can be two different control modes (corresponding to two stages of contact, see~\eqref{eqn:contact_phases}) during an impact event. Hence, to encode the two different control modes we consider two different contact force transmission models with different $\alpha$ parameters, denoted as $\alpha_{\mathcal{M}^c}$, which are respectively (i) for the compression stage with ${\mathcal{M}^c}=-1$ and (ii) for the restitution stage with ${\mathcal{M}_c}=1$. This leads to: 
% the following differential equation
% the impact model and
%(see \cref{subsec:force_transmission_model}), %is enforced in 
\begin{equation}
    \label{eq:ControlDyn}
    \hspace{-3mm}
    \begin{bmatrix} {\dot{F}}  \\ \ddot{F} \end{bmatrix} = 
    \begin{bmatrix} 0 & 1 \\ - \alpha_{\mathcal{M}^c}^2 & -2\alpha_{\mathcal{M}^c} \end{bmatrix} \begin{bmatrix} F \\ 
    \dot{F} \end{bmatrix} + \begin{bmatrix} 0 \\  \alpha_{\mathcal{M}^c}^2 \end{bmatrix}  F_{\mathcal{M}^c},
\end{equation}
that describes the contact force profile through two different control modes during an impact event. \textcolor{black}{Note that each control mode (\ie compression and restitution) has different $F_{\mathcal{M}^c}$. For each control mode, $\alpha_{\mathcal{M}^c}$ is optimized} to modulate the contact duration $\Delta T$ (see~\cref{rem:impdur2alpha}) and contact force profile along the surface normal. These are optimised within the MMTO according to the feasible contact duration, which is related to the velocity of object, and the workspace of the robot arms.

\subsubsection{Friction cone constraint} 
To model the force in the two tangent dimensions of the surface, we use a friction cone constraint. This is activated when the robot switches from free motion ($\mathcal{M}^d = 0$) to contact ($\mathcal{M}^d = 1$) mode, and it is enforced in the MMTO with: 
% the following equation:
\begin{equation}
    \angle \left( \bm F_n, \bm n \right) \leq \arctan \left(\mu\right) ,
\end{equation}
where $\angle(\cdot)$ is the operator used to denote the included angle between two vectors and $\mu$ is the friction coefficient. 

% contact time which is optimized for making the desired contact force. 
% Also, note that the feasible contact duration is related to the velocity of object and the workspace of the robot arms; thus, $\alpha(\mathcal{M}^c)$
% should be limited according to the attributes of the physical system.
% In the proposed MMTO, both $\alpha(\mathcal{M}^c)$---\ie stiffness---and contact duration $\Delta T$ are optimized to satisfy the workspace limits of the robot.
% , i.e. optimizing the stiffness of the contact force transmission model. 

% According to \eqref{eq:ControlDyn}, the contact force transmission model is parameterized by only one parameter $\alpha_{\mathcal{M}_c}$.
% $\alpha_l$ is related to different contact phases as in 
% Hybrid control is realized with two compliance controllers (i) ${\mathcal{M}_c}=-1$ and (ii) ${\mathcal{M}_c}=1$, and one position controller (iii) ${\mathcal{M}_c}=0$, one for each contact stage defined in \eqref{eqn:contact_phases}. 

% is realized by optimizing 
% for each controller (i) $l=-1$ and (ii) $l=1$ which corresponds
% Based on 
% to the contact stages defined with \eqref{eqn:contact_phases}.
\vspace{-2mm}
\subsection{\textcolor{black}{Stiffness Modulation during Contact}}
The main benefit of the proposed force transmission model 
is that based on the optimized parameter $\alpha_{\mathcal{M}^c}$, the corresponding damping and stiffness parameters of the impedance controller on a robot can be obtained. We can achieve this as the same mass-spring-damper system model (see~\cref{sec:impact_modelling}) which establishes the relationship between force and position is also adopted in the impedance controller to regulate the operational force of a robot manipulator.
% Furthermore, based on the relationship between $\alpha$ and $K$ (see~\cref{rem:alpha2stiff} and~\eqref{eqn:mk2alpha}), 
In this way, in MMTO, the stiffness characteristics of the robots are also optimized in a coherent way through $\alpha_{\mathcal{M}^c}$---considering both stages of impact (see~\cref{subsec:controlModes})---without separating the contact scheduling from stiffness modulation into two separate optimisation levels as in~\cite{nakanishi2013spatio}.

\vspace{-2mm}
\subsection{Manipulation Planning Modes for Catching} 
In addition to the dynamics ($\mathcal{M}^d$) and the control ($\mathcal{M}^c$) modes, we use manipulation planning modes ($\mathcal{M}^p$), commonly used in multi-modal planning to enforce geometric constraints, such as the object being coupled to the robots' end-effectors \textcolor{black}{(without modelling any force interaction), dual-arm motion coordination in free-motion, \etc }
% end-effector and the object are design according to different contact modes respectively, such that
Here, we have two different modes; (i) the decoupled mode, where the robots' end-effectors are allowed to move independently, denoted as $\mathcal{M}^p = 0$, and (ii) the coupled mode, where the robots' end-effectors maintain a fixed relative pose to the object \textcolor{black}{and between them}, denoted as $\mathcal{M}^p = 1$. According to these two modes we define the following constraint: 
\begin{equation}
       \text{d}(\bm x^e, \bm p) \left\{ \begin{array}{clr} > 0  &\mbox{for} & \mathcal{M}^p=0 \\ = 0 &\mbox{for} & \mathcal{M}^p=1 \end{array}\right. ,
\end{equation}
where $\text{d}(\cdot)$ is the signed distance operator 
% which calculates the  
(w.r.t the surface) between two elements, one of which can be the desired contact point $\bm p$ and the other is the end-effector's position $\bm x^e$. 

\textcolor{black}{Note that for the catching scenario specifically, we assume that for the short time duration of the motion, each end-effector of the robot remains on the same side of contact surface, and both end-effectors make contact with the object at the same time.} The mode ($\mathcal{M}^p$) transition from decoupled to coupled happens when the mode ($\mathcal{M}^d$) transitions from free-motion to contact, yet for other tasks this might not be the case. Most importantly, once the coupled mode is active the motion of the (two) robot arms needs to be coordinated, as they are constrained on the object. Hence, the MMTO needs to generate a motion that transits from independent motion (decoupled) to coordinated motion (coupled) of the robot arms.

% \begin{assumption}
% In the short time duration of catching, the end-effectors of the robot remain in the same side of contact surfaces.
% \end{assumption}

\vspace{-2mm}
\subsection{Mode-Independent Constraints}
Here, we introduce all the mode-independent constraints in the proposed MMTO (see~\eqref{eq:csDSTO}), i.e. the constraints that are not related to the mode sequence $\mathcal{M}$. 

\subsubsection{Initial and final conditions} The initial states of the object and the robot are respectively denoted as $\bm{x}_0^o=\bm{x}_*^{o}$, $\dot{\bm{x}}_0^o=\dot{\bm{x}}_*^{o}$ and $\bm{x}_0^e=\bm{x}_*^{e}$, $\dot{\bm{x}}_0^e=\dot{\bm{x}}_*^{e}$, while the final state of the object is defined according to the task requirement. For the catching task without a target position we set the final state of the object as: $\dot {\bm{x}}^o_N = \bm 0$ and for catching with a target position we set the final state as  $\dot {\bm{x}}^o_N = \bm 0$, ${\bm{x}}^o_N = {\bm{x}}_d^{o}$.

\subsubsection{Workspace limit of the bimanual robot} %For the 7 DOF KUKA-iiwa
For each robot manipulator, we use a sphere located at the center of the second joint, denoted as $\bm c_{j2}$, to approximately define the workspace: $\| \bm c - \bm c_{j2}\| \leq R_w$, with $R_w$ being the radius of the workspace sphere.

\subsubsection{Time bound between adjacent knots} Further, we also set upper and lower bound on the time duration between any two adjacent knots as: $\Delta T_l \leq \Delta T_n \leq \Delta T_u$.

% %=============================================================================== 
% \section{Hybrid Control Based on Reference Spreading/ Indirect Force Control}\label{sec:hybrid_control}
% \input{sections/impact_aware_control}

% %===============================================================================
\section{Simulation Results}\label{sec:sim_results}

% In this section, we will first present the analysis results of normal impact and tangential impact when making contact along different directions and locations. Then we will present the autonomous impact-aware contact selection algorithm for arbitrary shape of object. At last, the numerical results of impact-agnostic and impact-aware methods for catching will be compared. All simulations are conducted on a 64-bit Intel Quad-Core i9 3.60GHz computer with 64GB RAM and are realized with the Bullet physics simulation library.

% \theo{Remember statistical analysis of convergence under noise and error analysis of prediction w.r.t ground truth} 

In this section, we 
%present a number of numerical and physical simulation studies that we performed in order to 
analyse and validate the hypotheses (see~\cref{sec:impact_modelling}), models (see~\cref{subsec:cntforceTrans}) and algorithms (see~\cref{sec:contact_searching} and~\cref{sec:MMTO}) described in the previous sections, using various numerical simulation results. 
First, we analyse the impulse distribution along normal and tangential directions, for the case of a moving object making contact with the robot at different contact locations and directions.
\textcolor{black}{Second, we present the results of the impact-aware contact selection algorithm, %showing its convergence 
on several discontinuous and non-convex shaped objects.}
Third, we perform an ablation study by numerically comparing our impact-aware method with its impact-agnostic version, on a bimanual catching task.
We used a 64-bit Intel Quad-Core i9 3.60GHz computer with 64GB RAM for running all the numerical computations and simulations.

\begin{figure}[t]
    \centering
    \def\svgwidth{0.95\linewidth}
%% Accompanies image file 'contact_scenarions.pdf' (pdf, eps, ps)
\begingroup%
  \makeatletter%
  \providecommand\color[2][]{%
    \errmessage{(Inkscape) Color is used for the text in Inkscape, but the package 'color.sty' is not loaded}%
    \renewcommand\color[2][]{}%
  }%
  \providecommand\transparent[1]{%
    \errmessage{(Inkscape) Transparency is used (non-zero) for the text in Inkscape, but the package 'transparent.sty' is not loaded}%
    \renewcommand\transparent[1]{}%
  }%
  \providecommand\rotatebox[2]{#2}%
  \newcommand*\fsize{\dimexpr\f@size pt\relax}%
  \newcommand*\lineheight[1]{\fontsize{\fsize}{#1\fsize}\selectfont}%
  \ifx\svgwidth\undefined%
    \setlength{\unitlength}{595.27559055bp}%
    \ifx\svgscale\undefined%
      \relax%
    \else%
      \setlength{\unitlength}{\unitlength * \real{\svgscale}}%
    \fi%
  \else%
    \setlength{\unitlength}{\svgwidth}%
  \fi%
  \global\let\svgwidth\undefined%
  \global\let\svgscale\undefined%
  \makeatother%
  \begin{picture}(1,0.23809524)%
    \lineheight{1}%
    \setlength\tabcolsep{0pt}%
    \put(0,0){\includegraphics[width=\unitlength,page=1]{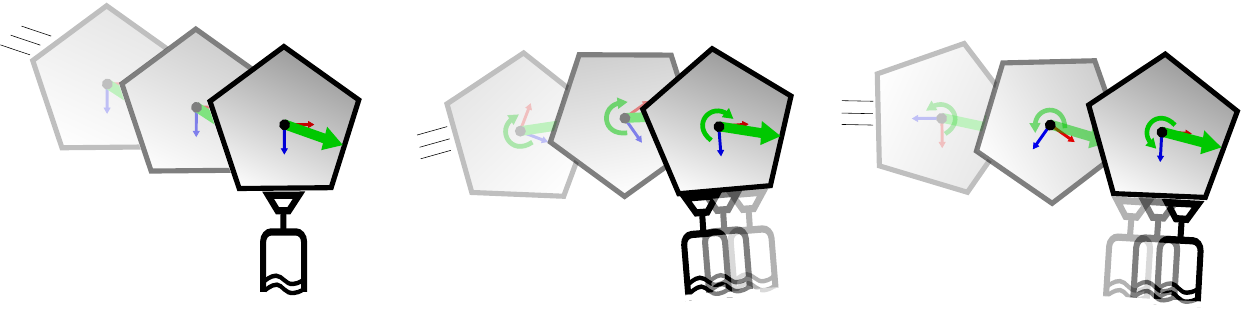}}%

    \put(0.03,0.04){\color[rgb]{0,0,0}\makebox(0,0)[lt]{\lineheight{1.25}\smash{\begin{tabular}[t]{l}\scriptsize{end-effector}\end{tabular}}}}%
    
    \put(0.0,-0.02){\color[rgb]{0,0,0}\makebox(0,0)[lt]{\lineheight{1.25}\smash{\begin{tabular}[t]{l}\footnotesize{(a) Different directions of motion}\end{tabular}}}}%

    \put(0.525,-0.02){\color[rgb]{0,0,0}\makebox(0,0)[lt]{\lineheight{1.25}\smash{\begin{tabular}[t]{l}\footnotesize{(b) Different contact locations}\end{tabular}}}}%

    \put(0.4, 0.06){\color[rgb]{0,0,0}\makebox(0,0)[lt]{\lineheight{1.25}\smash{\begin{tabular}[t]{l}\footnotesize{ $\bm{\omega} > 0$ }\end{tabular}}}}%
    
    \put(0.725, 0.06){\color[rgb]{0,0,0}\makebox(0,0)[lt]{\lineheight{1.25}\smash{\begin{tabular}[t]{l}\footnotesize{ $\bm{\omega} < 0$  }\end{tabular}}}}%

    \put(0.575, 0.23){\color[rgb]{0,0,0}\makebox(0,0)[lt]{\lineheight{1.25}\smash{\begin{tabular}[t]{l}\scriptsize{ Spinning object}\end{tabular}}}}%
    
  \end{picture}%
\endgroup%

    \caption{Making contact with different contact directions and locations. (a) Same velocity magnitude and different inclination angles. (b) Different locations along the normal direction of the motion with positive and negative angular velocity, respectively.}
	\label{fig:impulse_directions_locations}
    \vspace{-2mm}
\end{figure}
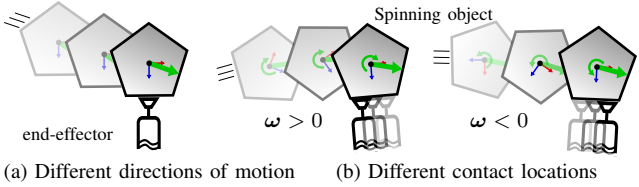
\vspace{-2mm}
\subsection{Impulse Distribution along Normal-Tangential Directions}
Our aim here is to determine the optimal contact direction and location for the robots to make contact with a moving object.
We use the 3D compliance impact model (see~\cref{subsec:3DcomplianceModel}) to
analyse the impulse distribution along normal and tangential directions in different scenarios.
\cref{fig:impulse_directions_locations} illustrates various scenarios where a static end-effector makes contact with a surface of a moving object. 
The scenarios considered involve different linear and angular velocities of the object, and different contact locations and directions.

The first scenario, exemplified in~\cref{fig:impulse_directions_locations}(a), considers an object making contact with the robot at a speed of~\SI{2}{\meter\per\second} and along different directions.
We consider the contact inclinations of \SI{180}{\degree}, \SI{175}{\degree}, \SI{160}{\degree}, \SI{150}{\degree}, \SI{140}{\degree}, \SI{130}{\degree}, \SI{120}{\degree}, \SI{100}{\degree}, \SI{95}{\degree}, and \SI{91}{\degree}, where~\SI{180}{\degree} corresponds to a contact velocity aligned with the normal vector of the contact surface and~\SI{90}{\degree} corresponds to a contact velocity aligned with the tangential vector. Note that an inclined angle of~\SI{90}{\degree} result in a normal contact velocity of zero, hence the contact will not happen.
%which would result in a normal contact velocity of zero.
%and makes contact with the robot end-effector.
% The inclined angles that we consider are $[180^{\circ}, 175^{\circ}, 160^{\circ}, 150^{\circ}, 140^{\circ}, 130^{\circ}, 120^{\circ}, 110^{\circ}, 100^{\circ}, 95^{\circ}, 91^{\circ}]$, where $180^{\circ}$ would be aligned with the normal vector of the surface. 
% Note that the inclined angle cannot be $90^{\circ}$ (i.e. the normal contact velocity is equal to zero) otherwise the contact will not happen.

\begin{figure*}[htb]
	\begin{center}
		\includegraphics[width=1.8\columnwidth]{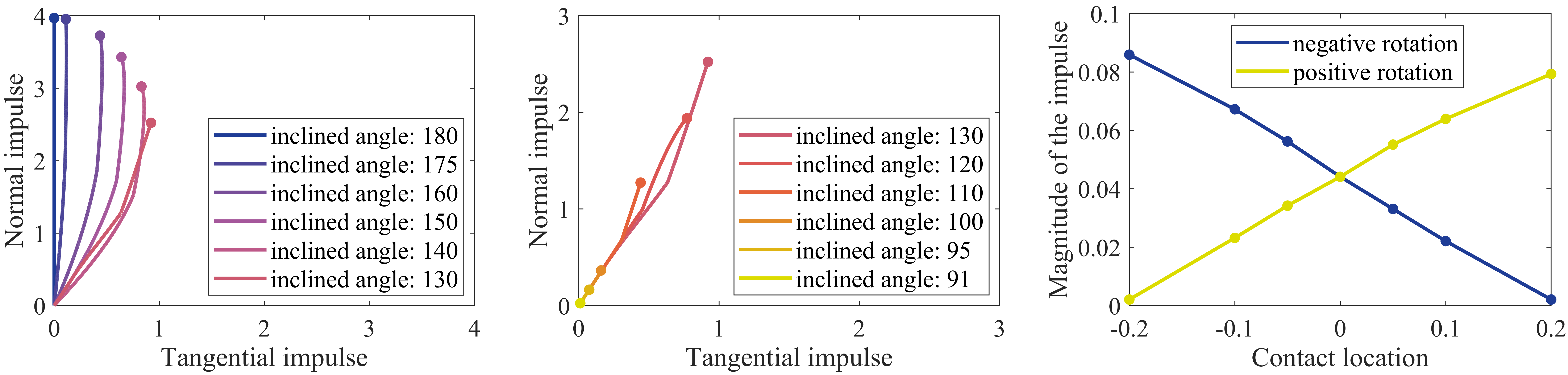}
		\vspace{-12pt}
	\end{center}
	\caption{Impulse distributions of different contact directions and locations (unit: \SI{}{\newton \second}). 1) Impulse distributions of the same velocity magnitude and different inclined angles; 2) Impulse distributions of different contact locations when making contact along the normal direction of the motion (positive angular velocity); 3) Impulse distributions of different contact locations when making contact along the normal direction of the motion (negative angular velocity).}
	\label{fig:impulse_analysis}
	\vspace{-5pt}
\end{figure*}

The first and second plots of~\cref{fig:impulse_analysis} show the 
%The quantities of interest, 
impulse distribution along the normal and tangential directions.
%, are shown in the first and second plots of~\cref{fig:impulse_analysis}.
As the inclination angle decreases, from a normal direction (\SI{180}{\degree}) to almost tangential ($\SI{91}{\degree}$), the total impulse magnitude decreases from~\SI{3.969}{\newton \second} to~\SI{0.028}{\newton \second} (the normal impulse decreases from~\SI{3.969}{\newton \second} to~\SI{0.025}{\newton \second} while the tangential impulse increases from ~\SI{0.0}{\newton \second} to~\SI{0.922}{\newton \second}, then decreasing to~\SI{0.012}{\newton \second}).
\textcolor{black}{Note that when the contact direction changes from the normal to tangential, there will be sliding contact during the impact evolution between the object and the robot. As a result, the friction cone will limit the impulse magnitude, 
which is consistent with \textit{~\cref{hypo:orientation}} in~\cref{subsec:impactProps}.}
%As a result, the impulse magnitude will decrease and be limited by the friction cone constraint,
% Therefore, it is better to make contact along the normal direction of the object motion as much as possible, which means that the projection of momentum along this direction will close to zero.

In the second scenario, we investigate the optimal contact location on a spinning object for mitigating the impulse resulting from the motion uncertainty.
In this scenario, the contact between the object and the end-effector occurs when the object travels with linear velocity of~\SI{2}{\meter\per\second}, and angular velocity of $\pm\SI{0.2}{\radian\per\second}$ (see~\cref{fig:impulse_directions_locations}(b)).
Further, we consider seven contact locations (one at the center, three left and three right) and at the moment of contact we ensure that the surface normal (of the end-effector) is perpendicular to the linear velocity of the contact point.
% As shown in the second and third figure of Fig.~\ref{fig:impulse_directions_locations} where we consider the negative relative rotation ($- 0.5 rad/s$) and positive relative rotation ($+ 0.5 rad/s$), 
\cref{fig:impulse_analysis}(3) shows 
the corresponding impulse distribution for different contact locations.
For different relative angular velocities, the impulses at the same location are different.
The contact location which is closest to the center of mass of the object has moderate impulse value irrespective of the different direction of relative angular velocity, i.e. minimizing the effect of motion uncertainty on the impact, which is consistent with \textit{~\cref{hypo:position}} in~\cref{subsec:impactProps}.
%Therefore, the second principle for contact selection that the impact-aware contact location should be close to the centre of mass of the object is verified.

% \begin{figure}[t]
% 	\centering
%     \def\svgwidth{0.95\linewidth}
%     \input{./figures/contact_direction_location.tex}
% 	\caption{Making contact with different contact directions and locations. 1) Same velocity magnitude and different inclined angles; 2) Different locations along the normal direction of the motion (positive angular velocity); 3) Different locations along the normal direction of the motion (negative angular velocity).}
% 	\label{fig:impulse_directions_locations}
% % 	\vspace{-10pt}
% \end{figure}

\begin{figure*}[htb]
	\begin{center}
		\includegraphics[width=1.9\columnwidth]{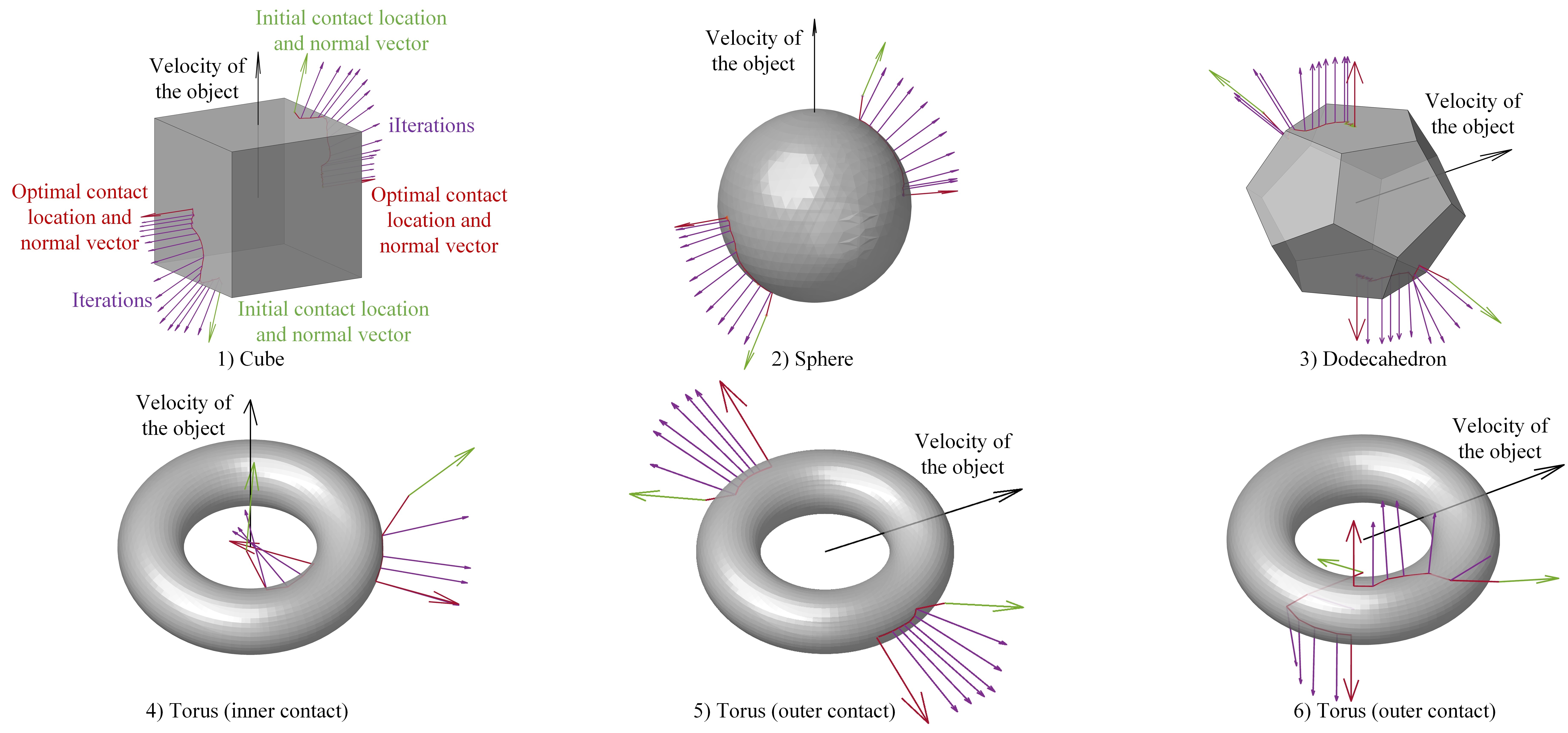}
		\vspace{-12pt}
	\end{center}
	\caption{Contact searching for objects with different kind of shapes (noncontinuous and nonconvex): 1) Cube; 2) Sphere; 3) Dodecahedron; 4/5/6) Torus. Black arrow: the velocity of the object; Green arrow: initial contact location and its corresponding normal vector on the contact surface; Red arrow: optimal contact location and its corresponding normal vector on the contact surface; Purple arrow: the iteration of contact location and normal vector.}
	\label{fig:sim_contact_searching}
	\vspace{-10pt}
\end{figure*}

\vspace{-2mm}
\subsection{Impact-Aware Contact Searching for Moving Objects}
In this section, we will demonstrate the effectiveness of the proposed impact-aware contact selection algorithm (see~\cref{sec:contact_searching}). Given any kind of mesh that describes the shape of an object and the object velocity (linear and angular), the proposed algorithm can find the optimal contact locations, which are resistant to impact. Without loss of generality, here we consider four objects with different shapes and velocities, i.e. cube (side length $l=1.0m$), sphere (radius $r=0.5m$), dodecahedron (side length $l=0.3m$) and torus (major radius $R=0.225m$, minor radius $r=0.075m$). The cost function weights of~\eqref{eqn:contact_opt_cost}  are $w_1 = w_2 = 2.0$ and $w_3 = 1.0$. The results of impact-aware selection algorithm are shown in~\cref{fig:sim_contact_searching}. Given the velocity of the object (black arrow) and any arbitrary initial contact location (green arrows), the optimal contact locations and their corresponding normal vectors on the surface (red arrow) can be found in $10 ms$ with the convergence threshold of contact locations being $10 mm$. The contact points along with their normal vectors during the iterations of the algorithm are shown with the purple arrows. Also, note that the proposed algorithm is able to select the optimal contact locations across discontinuous and non-convex contact surfaces, due to
its sequential linearization nature and the smoothing of the normal vectors between different contact surfaces. 
Especially for the case shown in ~\cref{fig:sim_contact_searching} (4-6), the proposed algorithm is able to find both the inner and outer contacts for the torus with the non-convex shape.
% \theo{Let's give exact number here, maybe mean and variance. Let;s say that we evaluated 10 different velocities with 4 different objects and the mean is ... and variance ... for finding two contact points}.

% \begin{figure*}[t]
%     \centering
%     \def\svgwidth{0.95\linewidth}
%     \input{./figures/contact_searching.tex}
%     \caption{Contact searching for objects with different kind of shapes: 1) Cube; 2) Regular dodecahedron; 3) Sphere.}
%     \label{fig:impact_model}
%     \vspace{-3mm}
% \end{figure*}
\vspace{-2mm}
\subsection{Impact-Agnostic vs Impact-Aware Methods}
\label{subsec:impact-agn-aware-sim}
\begin{figure*}[htb]
	\begin{center}
		\includegraphics[width=2.02\columnwidth]{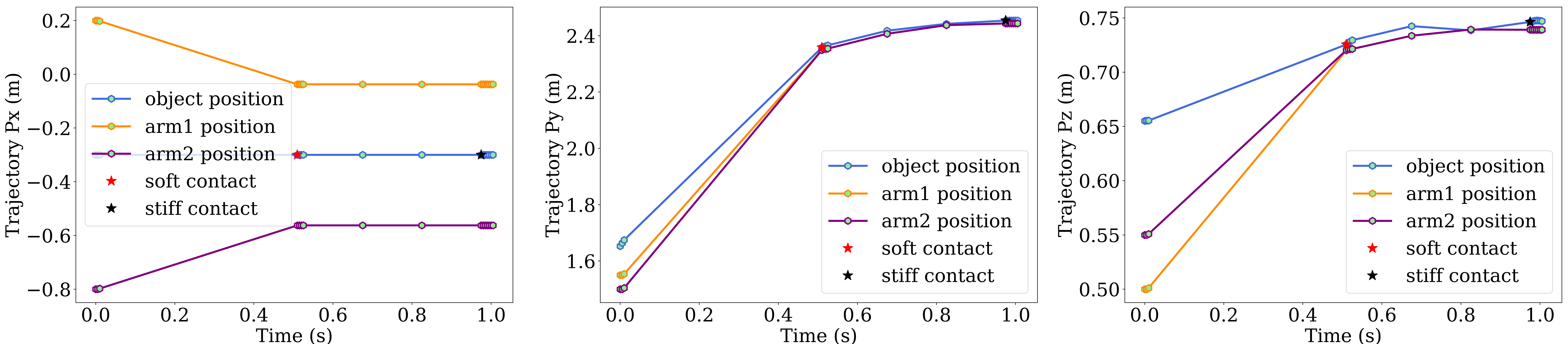}
		\vspace{-12pt}
	\end{center}
	\caption{Trajectory optimization results of impact-agnostic method. The red star and black star belong to the single contact phase of the impact-agnostic method, there is no difference between them.}
	\label{fig:sim_trajectory_baseline}
	\vspace{-10pt}
\end{figure*}

\begin{figure*}[htb]
	\begin{center}
		\includegraphics[width=2.02\columnwidth]{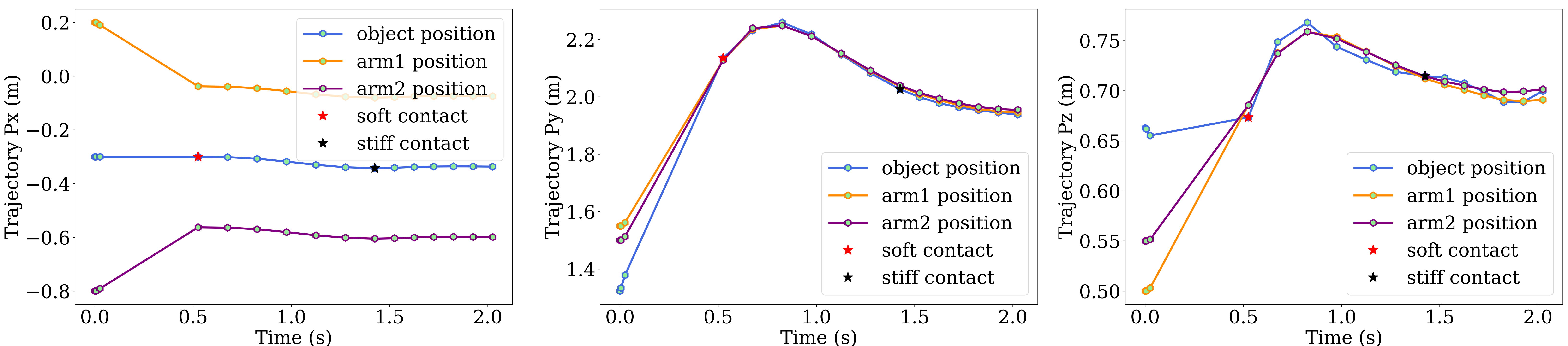}
		\vspace{-12pt}
	\end{center}
	\caption{Trajectory optimization results of impact-aware method. The red star represents the beginning of the first contact phase, while the black star represents the beginning of the second contact phase.}
	\label{fig:sim_trajectory_ours}
	\vspace{-10pt}
\end{figure*}

\begin{figure*}[htb]
	\begin{center}
		\includegraphics[width=2.02\columnwidth]{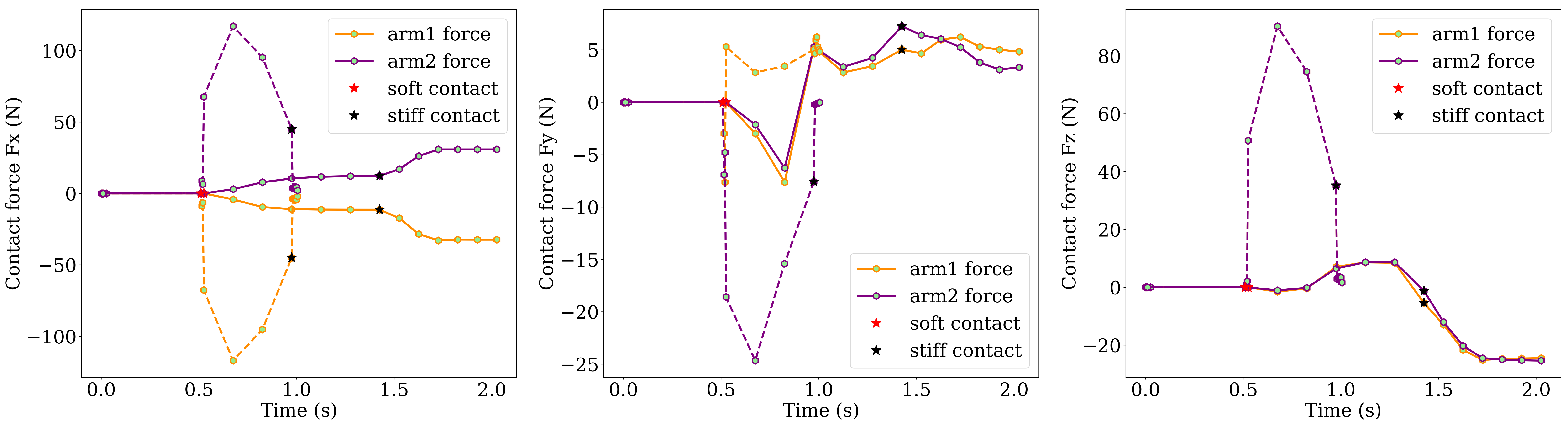}
		\vspace{-12pt}
	\end{center}
	\caption{Desired contact forces resulted from impact-aware (solid line) and impact-agnostic (dashed line) MMTO method. %We use blue line to denote the trajectory of the swing object while the yellow and purple lines are used to represent the trajectory and contact force of two arms. 
    The red and black stars are respectively used to denote the beginning of compression contact phase (soft contact) and restitution contact phase (stiff contact) for impact-aware method in Fig.~\ref{fig:sim_trajectory_ours}, while for impact-agnostic method in Fig.~\ref{fig:sim_trajectory_baseline} both of them belong to the single contact phase. }
	\label{fig:sim_force_comp}
	\vspace{-2mm}
\end{figure*}

\textcolor{black}{Here, we compare the planned trajectories generated by the proposed impact-aware MMTO method that uses our contact force transmission model (see~\cref{subsec:cntforceTrans}) against an impact-agnostic method (MMTO without contact force transmission model).} The task is specified as catching and stopping a swinging object using two contacts (end-effectors), in which the cost function is defined to minimize the contact force, stiffness and velocity for each robot end-effector. For this scenario, at the moment of contact ($t \approx 0.55s $) the object has a linear velocity around $1.0 m/s$ and an angular velocity around $\SI{0.5}{\radian\per\second}$.~\cref{fig:sim_trajectory_baseline} and~\cref{fig:sim_trajectory_ours} show the three dimension trajectories of the object and the end-effectors generated from impact-agnostic method and impact-aware method, respectively, while~\cref{fig:sim_force_comp} shows the corresponding force profiles. 
%of both methods.  

At the moment of contact, the impact-agnostic trajectory optimization plans a large contact force (more than 100 N) with short contact duration (less than 0.5 s) for both contacts, which stops the object suddenly (see~\cref{fig:sim_trajectory_baseline}). However, as shown in~\cref{fig:sim_force_comp}, our impact-aware MMTO method minimizes %the stiffness and 
planned contact force at the initial phase of the contact making process, to establish a stable contact, and after that 
% the stiffness and 
the contact force is increased such that the desired manipulation task (stopping the moving object) is achieved. This corresponds to smooth robot motion and contact force, consequently smooth stopping of the object motion, as shown in~\cref{fig:sim_trajectory_ours}.  

% We use blue line to denote the trajectory of the swing object while the yellow and purple lines are used to represent the trajectory and contact force of two arms. The red and black stars are respectively used to denote the beginning of compression contact phase (soft contact) and restitution contact phase (stiff contact) for impact-aware method in Fig.~\ref{fig:sim_trajectory_ours}, while for impact-agnostic method in Fig.~\ref{fig:sim_trajectory_baseline} both of them belong to the single contact phase. 

% As shown 
% For catching a swinging object with two KUKA-iiwa robots as shown in Fig.~\ref{fig:exp_swinging_object}, 

% \subsection{Catching and Throwing Flying Objects}

% %===============================================================================
\section{Experiment Results}\label{sec:exp_results}

\begin{figure*}[htb]
	\begin{center}
		\includegraphics[width=1.93\columnwidth]{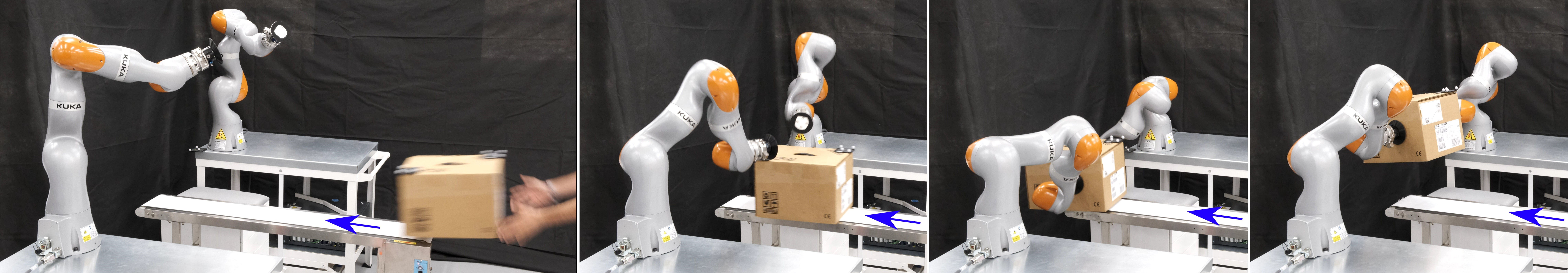}
		\vspace{-12pt}
	\end{center}
	\caption{Two KUKA iiwa arms capture a moving object which is thrown onto the conveyor belt by an operator.}
	\label{fig:exp_conveyor_belt}
	\vspace{-10pt}
\end{figure*}

In this section, we describe the experimental setup (both hardware and software) and the scenarios adopted to validate the proposed impact-aware methods in the real world. \textcolor{black}{The goal is to minimize the impulsive force when halting or manipulating a moving object} given only measurements of the object's pose. As shown in the attached video\footnote{\url{https://youtu.be/1__FuHY3-qo}}, this task is realized under different conditions where the object follows: (i) a definite motion on a conveyor belt (see~\cref{fig:exp_conveyor_belt}), (ii) a constrained (swinging) motion by being tethered to the ceiling and (iii) a free-flying motion after being thrown.

\textcolor{black}{Next, we will first introduce the components of hardware setup and implementation details of each module in Fig.~\ref{fig:overview}. Then we compare the experiment results of impact-agnostic method and the proposed impact-aware method. We further investigate the benefit of indirect force controller which generates penetration depth from desired contact force and stiffness, compared with the prescribed penetration method which uses a fixed penetration depth regardless of contact force and stiffness, both of them are used to track the desired motion and contact force from MMTO. At last, we validate and demonstrate the capabilities of the proposed system on the scenario of catching a free-flying object. }

\begin{figure*}[t]
\centering
  \newcommand\fh{2.7cm}
  \subfloat{%
      \begin{tikzpicture}%
        \node[above right,inner sep=0, outer sep=0] (image) at (0,0) {%
         \includegraphics[height=\fh]{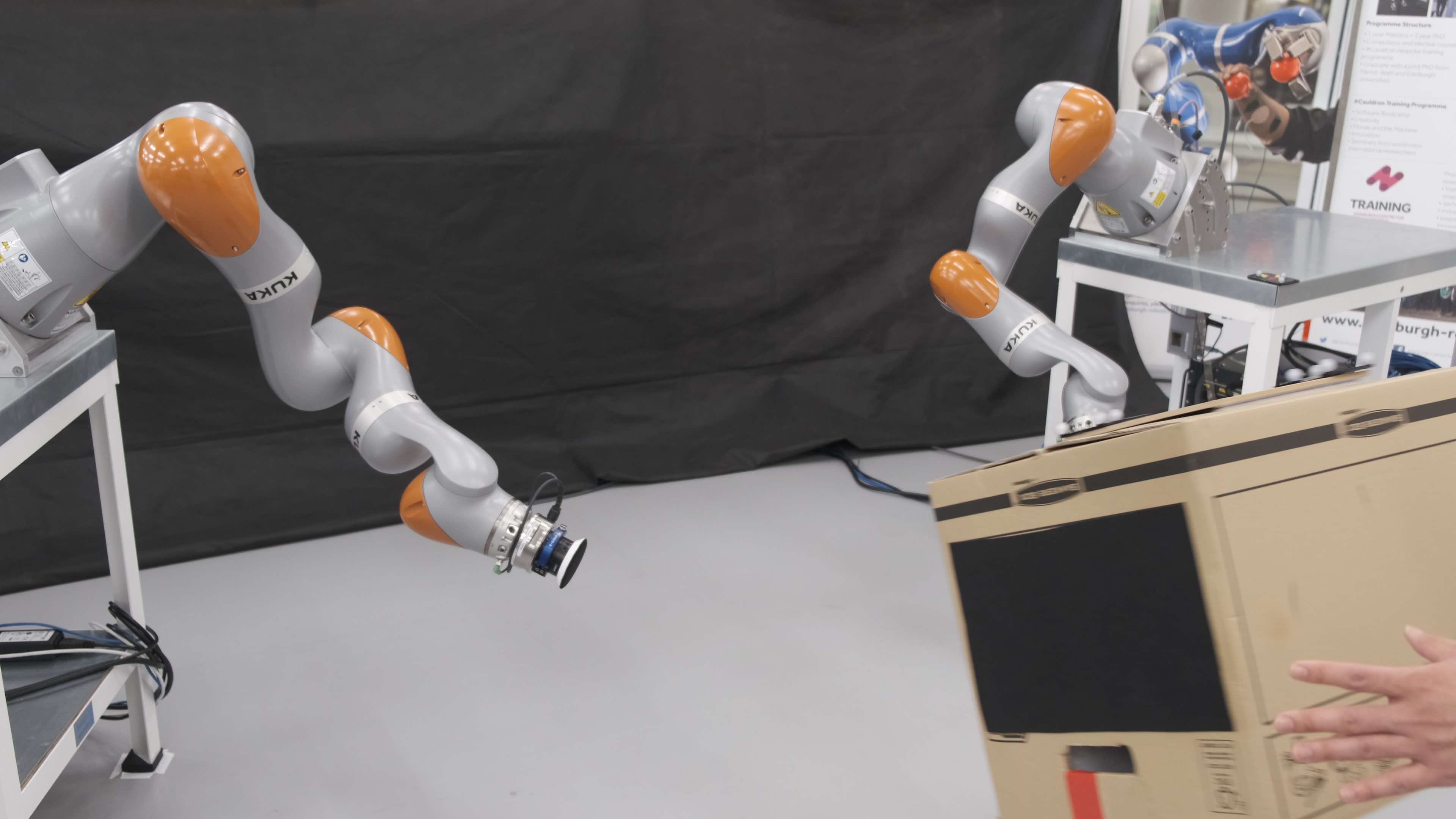}\label{fig:catching_swinging_object:moment1}%
        };%
        % Create scope with normalized axes
        \begin{scope}[%
        x={($0.1*(image.south east)$)},y={($0.1*(image.north west)$)}]%
            \draw[latex-,thick,blue] (6.2,2.6) -- ++(-2.0,-1.0) node[below,black,inner sep=0pt]{\small\textbf{Swinging object}};%
        \end{scope}%
        \end{tikzpicture}%
  }\,%
  \subfloat{\includegraphics[height=\fh,trim={0 0 22cm 0},clip]{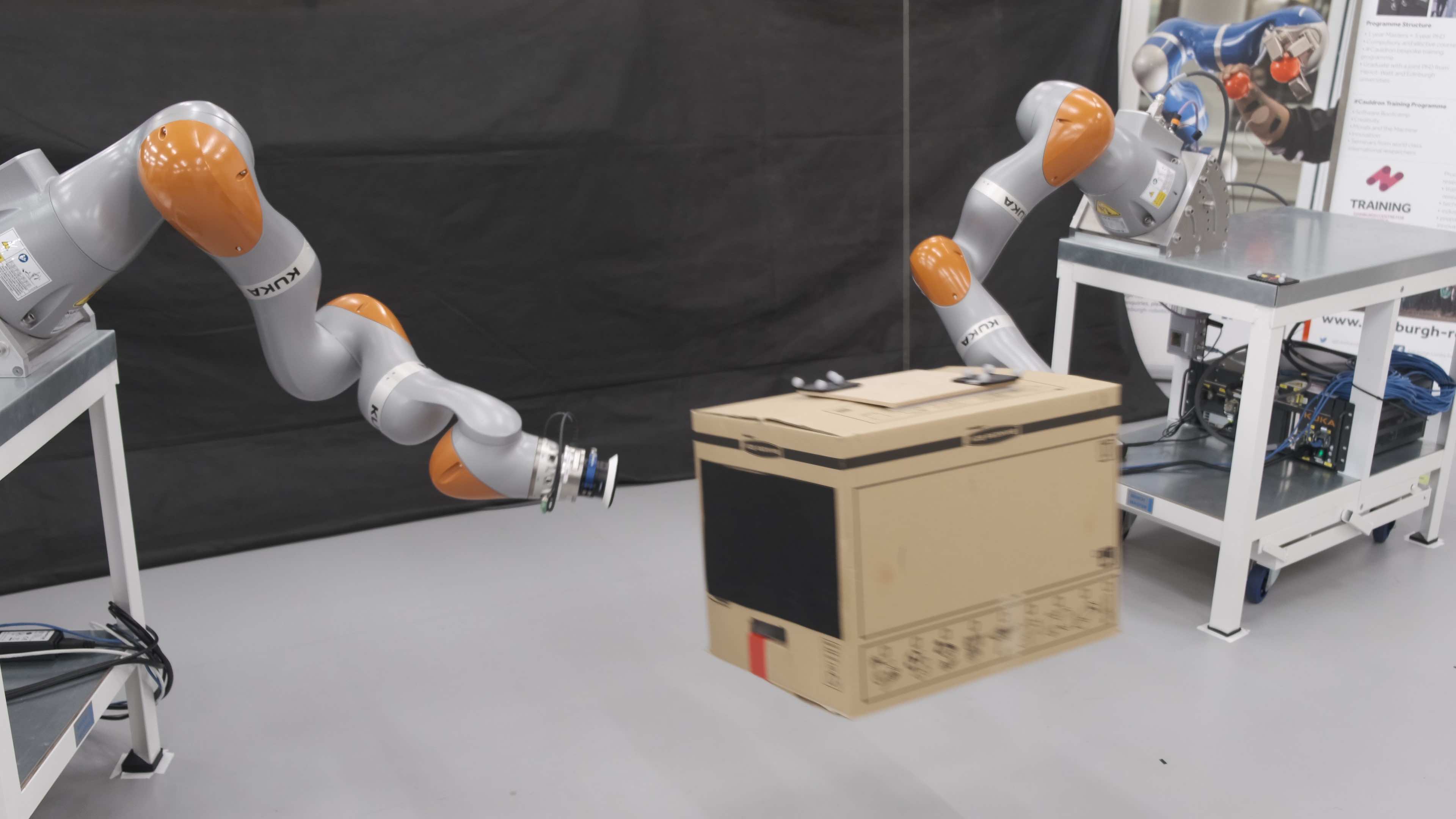}\label{fig:exp_swinging_objec:moment2}}\,%
  \subfloat{\includegraphics[height=\fh,trim={0 0 22cm 0},clip]{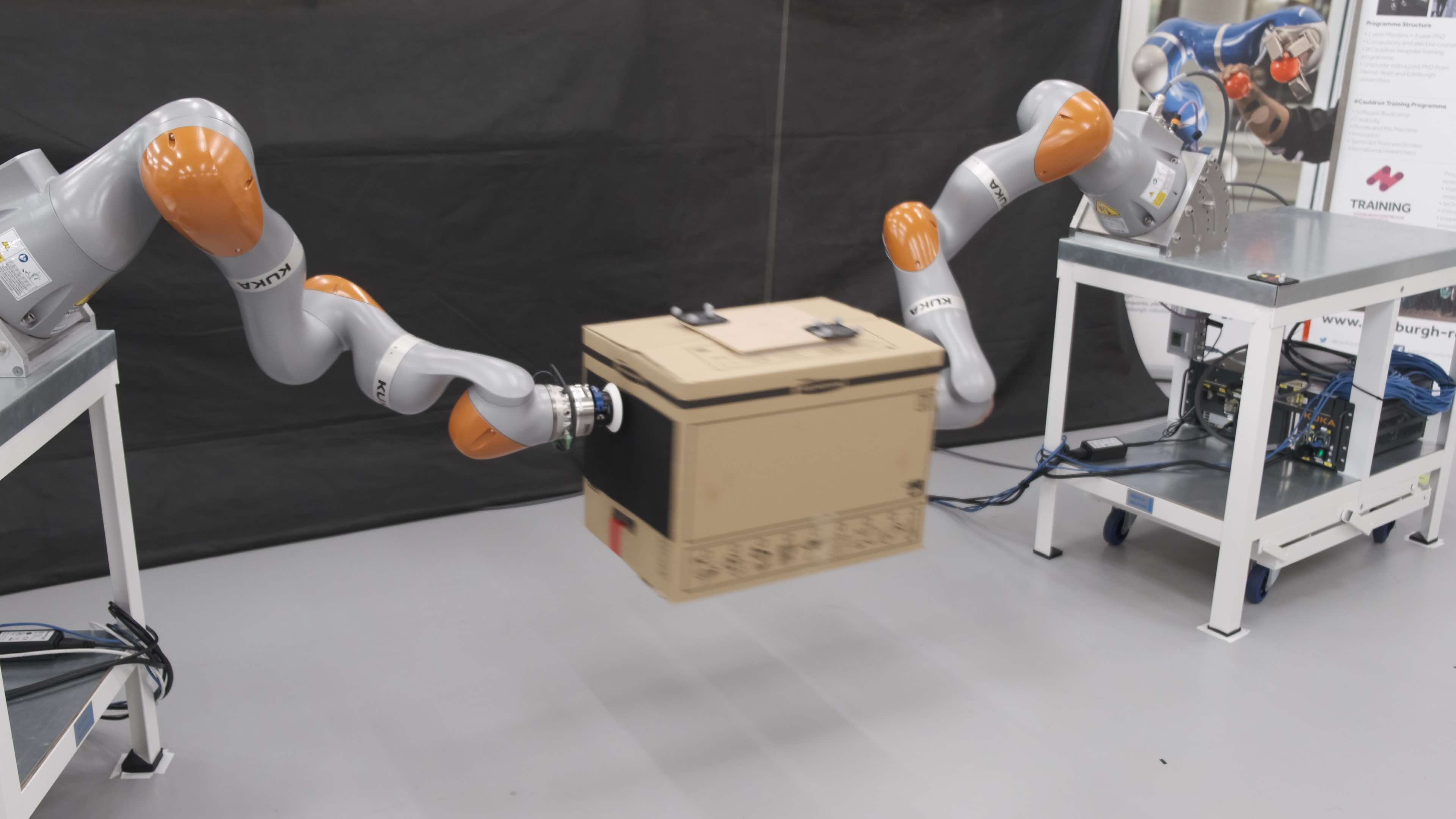}\label{fig:exp_swinging_object:moment3}}\,%
  \subfloat{\includegraphics[height=\fh,trim={0 0 22cm 0},clip]{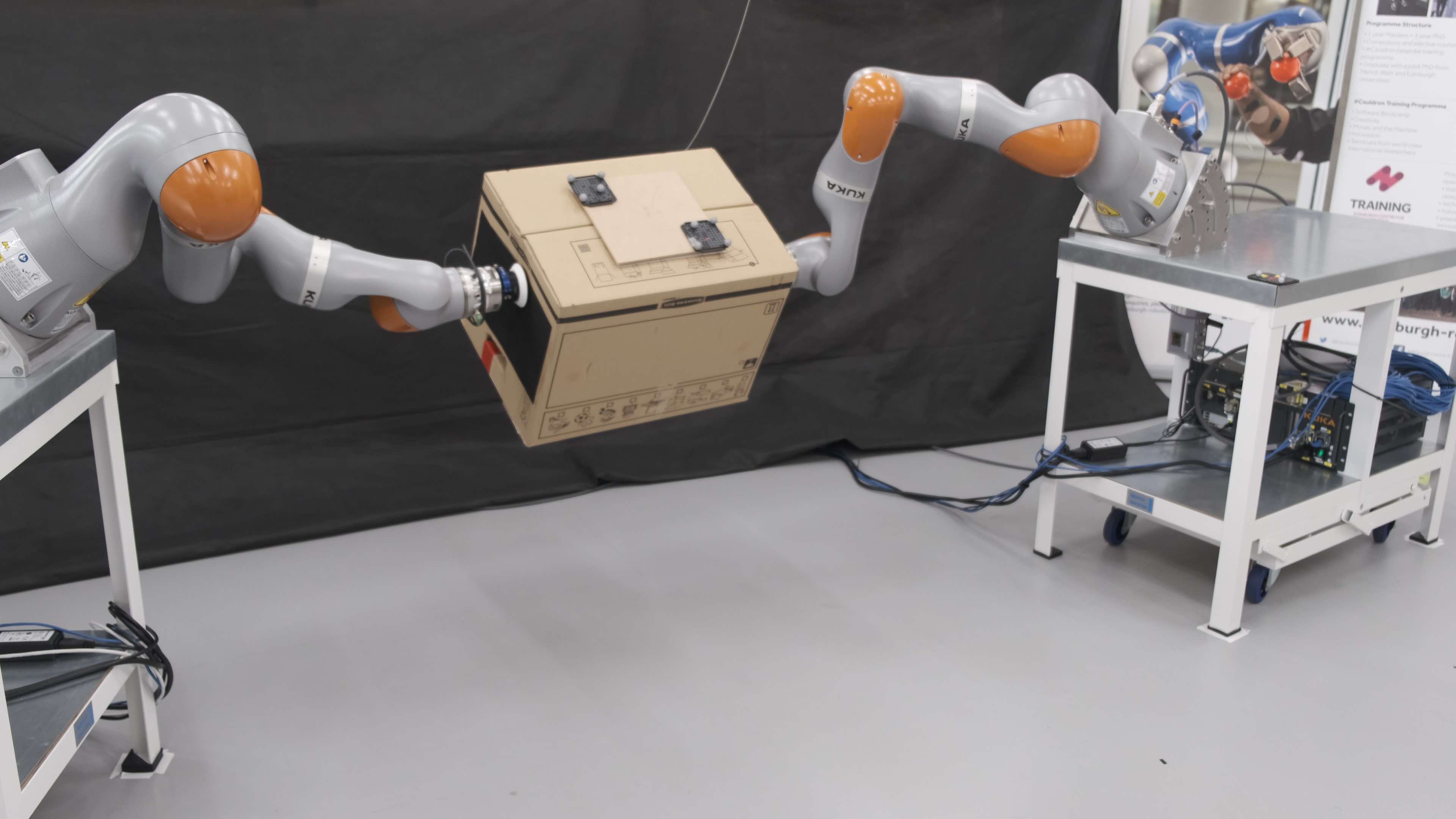}\label{fig:exp_swinging_object:moment4}}\\\vspace{-0.25cm}%
  % \subfloat{\includegraphics[height=\fh]{figures/catching_swinging_object/red_swing_object_perfect_catch_top_Moment1.jpeg}\label{fig:exp_swinging_object:top1}}\,%
  % \subfloat{\includegraphics[height=\fh,trim={8cm 0 8cm 0},clip]{figures/catching_swinging_object/red_swing_object_perfect_catch_top_Moment2.jpeg}\label{fig:exp_swinging_object:top2}}\,%
  % \subfloat{\includegraphics[height=\fh,trim={8cm 0 8cm 0},clip]{figures/catching_swinging_object/red_swing_object_perfect_catch_top_Moment3.jpeg}\label{fig:exp_swinging_object:top3}}\,%
  % \subfloat{\includegraphics[height=\fh,trim={8cm 0 8cm 0},clip]{figures/catching_swinging_object/red_swing_object_perfect_catch_top_Moment4.jpeg}\label{fig:exp_swinging_object:top4}}%
\caption{Snapshots of the two KUKA-iiwa robots catching a swinging object.}
\vspace{-0.25cm}
\label{fig:exp_swinging_object}
\end{figure*}

\vspace{-2mm}
\subsection{Experiment Setup}\label{sec:exp_setup}
The hardware setup includes a~\SI{4.2}{\kilogram} object with dimensions $0.55m\times0.40m\times0.42m$, a motion capture system (Vicon) which  measures the 3D pose of the object with 250 Hz and two \textit{KUKA-iiwa} compliant robots (LBR iiwa 14 R820) equipped with an ATI F/T sensor that measures force with 100 Hz at each end-effector. All force reported here are raw without any filtering. All the experiments are conducted on a 64-bit Intel Quad-Core i9 3.60GHz computer with 64GB RAM.
% Next, we provide details on each module mentioned in~\cref{sec:prob_desc}.
\textcolor{black}{The proposed system communicates with the KUKA iiwa robots through the Fast Research Interface (FRI) which allows us to change the stiffness parameter of the robots on-the-fly.} Next, we provide a sequential description of the modules of the system (also see~\cref{sec:prob_desc}). 
% Next, following the algorithmic flow we provide a sequential description of the modules of the system (see~\cref{sec:prob_desc}).

\textit{Estimation}: For all the experiments we estimate on-the-fly the velocity (linear and angular) of the object using an EKF~\cite{MooreStouchKeneralizedEkf2014} based on the \textit{robot\_localization} package\textcolor{blue}{\footnote{https://github.com/cra-ros-pkg/robot\_localization}}. The tuning of the EKF parameters, \textcolor{black}{such as the frequency of 250Hz, and the queue size of 5} %and the validation of the estimation results 
were done using the \textit{pyBullet} physics simulation~\cite{mower2022rospybullet} in which we have access to ground-truth velocities of the object. \textcolor{black}{Also, starting the object with zero or small velocities is key to obtaining accurate velocity estimations for the spinning/flying motions.}
% The root mean squared velocity error along the linear axis of motion is \textcolor{blue}{$ ??? m/s$} and along the angular is \textcolor{blue}{$ ??? rad/s$} for 20 different velocity configuration.

\textit{Prediction}: Given the pose and the estimated velocity of the object, its future trajectory is predicted by solving an IVP. The IVP is a nonlinear program similar in form to~\eqref{eq:csDSTO}, but only simpler, as it only considers the nominal dynamics (without modes) and geometric constraints that describe the environment, \eg tether of the swinging object. For the IVP, \textcolor{black}{we selected $N = 60$ knots with $\Delta \bm{T}_n = 0.03s$ $\forall n=\{0,...,N\}$ to predict the object motion going through the dual-arm workspace while it can be solved in less than 25 ms (using pre-stored initial guess). } %\textcolor{blue}{20 predicted trajectories of the swinging object were compared against traces of the object with root mean squared error within the area of interest (inside the workspace of the bimanual robot system) being for the position $ [ 0.008, 0.052, 0.011] m $ and for the orientation $ [ 2.7, 6.4, 3.1 ]^\circ$.}

\textit{Contact selection}: Once the predicted trajectory of the object is computed, the object pose (time instance of the trajectory) that is closest to the center of the workspace of the bimanual robot system is selected. Given this object pose, the CD-SQP contact selection algorithm is used to determine optimal contact locations for impact-aware catching. This is particularly useful for scenarios where the object is tumbling during the swinging motion, as shown in Fig.~\ref{fig:exp_rotating_object}. For this setup, two robots were used. Hence, $K=2$; \textcolor{black}{the selected weights are $w_1 = w_2 = 2$ and $w_3 = 1$ to prioritise impact mitigation over distribution of contact locations; and $\Delta \bm p_{min}=10mm $, $\Delta \bm p_{max} =50mm$ to ensure convergence of locally optimal contact locations in less than 10 ms.}

\textit{Multi-mode planning}:
Given the above as initial conditions (predicted object pose and velocity) and contact information (optimal contact location), the proposed \textcolor{black}{MMTO} algorithm is adopted to compute the optimal trajectories in real-time to capture and halt or manipulate the moving object. The \textcolor{black}{MMTO} simultaneously generates the optimal end-effector motion, force and stiffness profiles for the two robot arms, before, during and after contact. For all the experiments, we compute hybrid plans with the MMTO in less than 260 ms (using stored initial guess). Note that, to achieve the real-time computation, \textcolor{black}{we use a rough discretization with $N = 12$ collocation knots ($4$ for free motion, $4$ for soft contact phase and $4$ for stiff contact phase) with %constrain the time duration to 
$0.05\leq \Delta T_n \leq 0.4$ for free motion phase and $0.01\leq \Delta T_n \leq 0.1$ for two contact phases. Parameter tuning was performed in simulation with various initial object poses and velocities.}

\textit{Indirect force-control and IK}:  As the output of the MMTO are end-effector trajectories in Cartesian space, these are mapped to the robot motion in configuration space using differential inverse kinematics
and are streamed to the two position-controlled \textit{KUKA-iiwa} robots along with the prescribed Cartesian impedance (stiffness profiles). Further, in order to track the optimal contact force planned with the MMTO, we use an indirect force control scheme to regulate the end-effector's penetration of the \textit{KUKA-iiwa} robots, which is detailed in Appendix\ref{appendix:indirect_control}. Commanding stiffness enables impact-aware contact transitions, while the indirect force control enables fast contact force tracking without oscillations. 

The communication between modules is realized using \textit{ROS}, while the prediction (IVP) and the multi-mode planning (MMTO) are realized using \textit{CasADi} with its automatic differentiation and solved by the \textit{Ipopt} solver.

\begin{figure*}[htb]
	\begin{center}	\includegraphics[width=1.95\columnwidth]{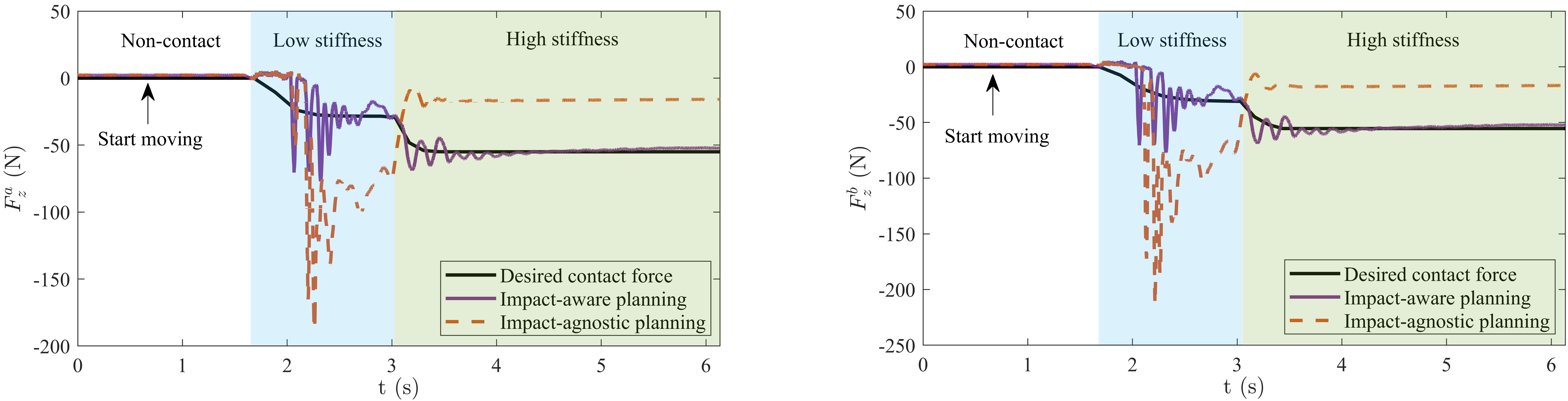}
		\vspace{-12pt}
	\end{center}
	\caption{Normal contact forces of two \textit{KUKA-iiwa} corresponding to impact-aware planning and impact-agnostic planning.}
	\label{fig:exp_impact_aware_contact_force}
 \vspace{-10pt}
\end{figure*}

\begin{figure*}[htb]
	\begin{center}	\includegraphics[width=1.95\columnwidth]{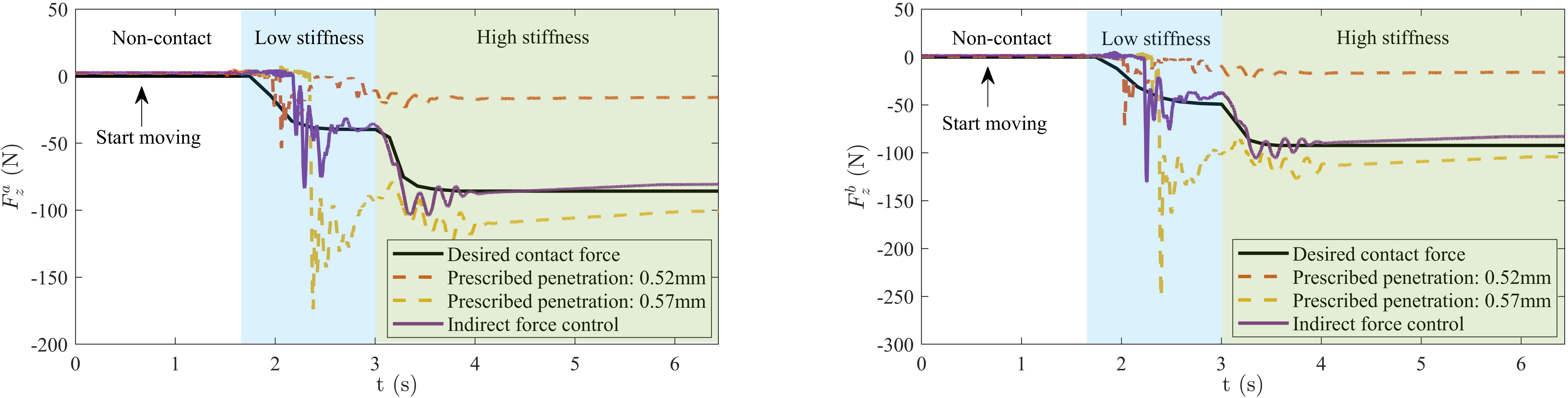}
	\vspace{-12pt}
	\end{center}
	\caption{Normal contact forces of two \textit{KUKA-iiwa} corresponding to indirect force control and prescribed penetration.}
	\label{fig:exp_indirect_force_control_results}
    \vspace{-10pt}
\end{figure*}
\vspace{-2mm}
\subsection{Impact-Agnostic vs Impact-Aware Method}\label{sec:impact_aware_agnostic_comparison}
\textcolor{black}{Here, similar to~\cref{subsec:impact-agn-aware-sim}, we compare the proposed impact-aware MMTO that uses our contact force transmission model (see~\cref{subsec:cntforceTrans}) against an impact-agnostic method (MMTO without contact force transmission model).}
In this scenario (see~\cref{fig:exp_swinging_object}), the object is tethered to the ceiling (tether length is $3 m$) and it is released from an initial position {(at an angle of $20^\circ$)} with zero linear and angular velocities. This results in a maximum speed at the contact moment being around ~\SI{1.8}{\meter\per\second}. We perform the catching task with both the impact-agnostic and the impact-aware methods, and we plot the normal contact forces in~\cref{fig:exp_impact_aware_contact_force}. 

It can be seen that the impact-agnostic method results in a large impulsive contact force at and right after the contact moment, i.e. $188 N$ and $213 N$ for two arms, respectively. In contrast, our impact-aware MMTO
results in smoothly-increasing contact forces, due to the low stiffness planned for making contact and the high stiffness for post-contact manipulation. The maximum contact force is around $77 N$ and $119 N$ for both arms, which are $59\%$ and $44\%$ smaller than the ones resulted from impact-agnostic method. \textcolor{black}{Further, it is noteworthy that compared with the impact-aware method (16 successful catches in 20 trials), the impact-agnostic method (3 successful catches within 10 trials) is more prone to lose the object (bouncing off) due to the large impulsive force at the moment of contact.}

% It can be seen that, without contact force constraint the impact-agnostic MMTO will generate a huge contact force right after the contact moment, which will results in a huge impulse force, i.e. respectively 188 $N$ and 213 $N$ for two arms.

% As a comparison, the impact-aware MMTO will generate the smoothly-increasing contact force with low stiffness at the beginning of the contact while generating the contact force with high stiffness for post-contact manipulation. 

% Because of the impact-aware planning and stiffness regulation, the impulse forces of two arms resulted from impact-aware method are both around 77 $N$, which are respectively 59 $\%$ and 64 $\%$ smaller than the ones resulted from impact-agnostic method. 

% The corresponding 
% in Fig.~\ref{fig:exp_swinging_object},
% For capturing the swinging object as shown  the impact-aware MMTO (with cdDS contact force constraint) and impact-agnostic MMTO (without cdDS contact force constraints) are respectively adopted to generated the dual-arm capturing motion. 

% The corresponding normal contact forces of two arms are shown in Fig., which is raw data without filtering.

\vspace{-2mm}
\subsection{Indirect Force Control vs Prescribed Penetration}
On the same setup as in~\cref{sec:impact_aware_agnostic_comparison}, the swinging object is released from a higher initial position which brings the object to a speed of around~\SI{2}{\meter\per\second} at the  moment of contact. The impact-aware MMTO is used to plan the dual-arm motion and contact force profiles for catching the swinging object. Yet, the 
joint position controller with Cartesian impedance control mode of the \textit{KUKA-iiwas} does not allow tracking the force profile directly. To be able to exert the desired force at the end-effector of the robot using Cartesian impedance control mode of the \textit{KUKA-iiwas}, we adopted an indirect force control scheme (see Appendix\ref{appendix:indirect_control}). The penetration depth of the robot end-effector is calculated based on the optimal contact force and stiffness, both computed from the impact-aware MMTO, and it is compared against a prescribed penetration approach. 
% Further, we compare this approach with a prescribed penetration approach. 

For the prescribed penetration approach, a constant penetration depth ($0.52 mm$ and $0.57 mm$ are respectively used) is defined for each robot to generate a contact force and make a stable contact with the swinging object. However, as shown in~\cref{fig:exp_indirect_force_control_results}, this will either result in large impulsive force at the moment of contact (penetration depth of $0.57 \ mm$ results in $174 N$ and $251 N$ for the two arms, respectively) or result in small contact forces which is not able to hold the object (penetration depth of $0.52 mm$). In contrast, as shown in~\cref{fig:exp_indirect_force_control_results}, the exerted force using the indirect force control scheme (purple solid line) are $84 N$ and $130 N$, which are respectively $48 \%$ and $52 \%$ smaller than the ones resulting from prescribed penetration method (brown dashed line). 

% In this experiment, the proposed method demonstrated its capability of catching a 4 $kg$ object swinging with a maximum speed around 2.3 $m/s$, while the impulse forces are still limited within its nominal payload (14 $kg$ for the KUKA-iiwa robots used in this paper).

% However, for the two position-controlled \textit{KUKA-iiwa} robots the contact force can not be directly controlled.

% In this section, two different methods,

% which are around 174 $N$ and 251 $N$ for two arms (penetration depth: $0.57 \ mm$), 
% As a comparison, an indirect force controller (more details can be found in Appendix.B) 
% 
 % the indirect force control approach is designed to generate the contact force for position-controlled robots, where

\begin{figure*}[t]
\centering
  \newcommand\fh{2.7cm}
  \subfloat{%
      \begin{tikzpicture}%
        \node[above right,inner sep=0, outer sep=0] (image) at (0,0) {%
            \includegraphics[height=\fh]{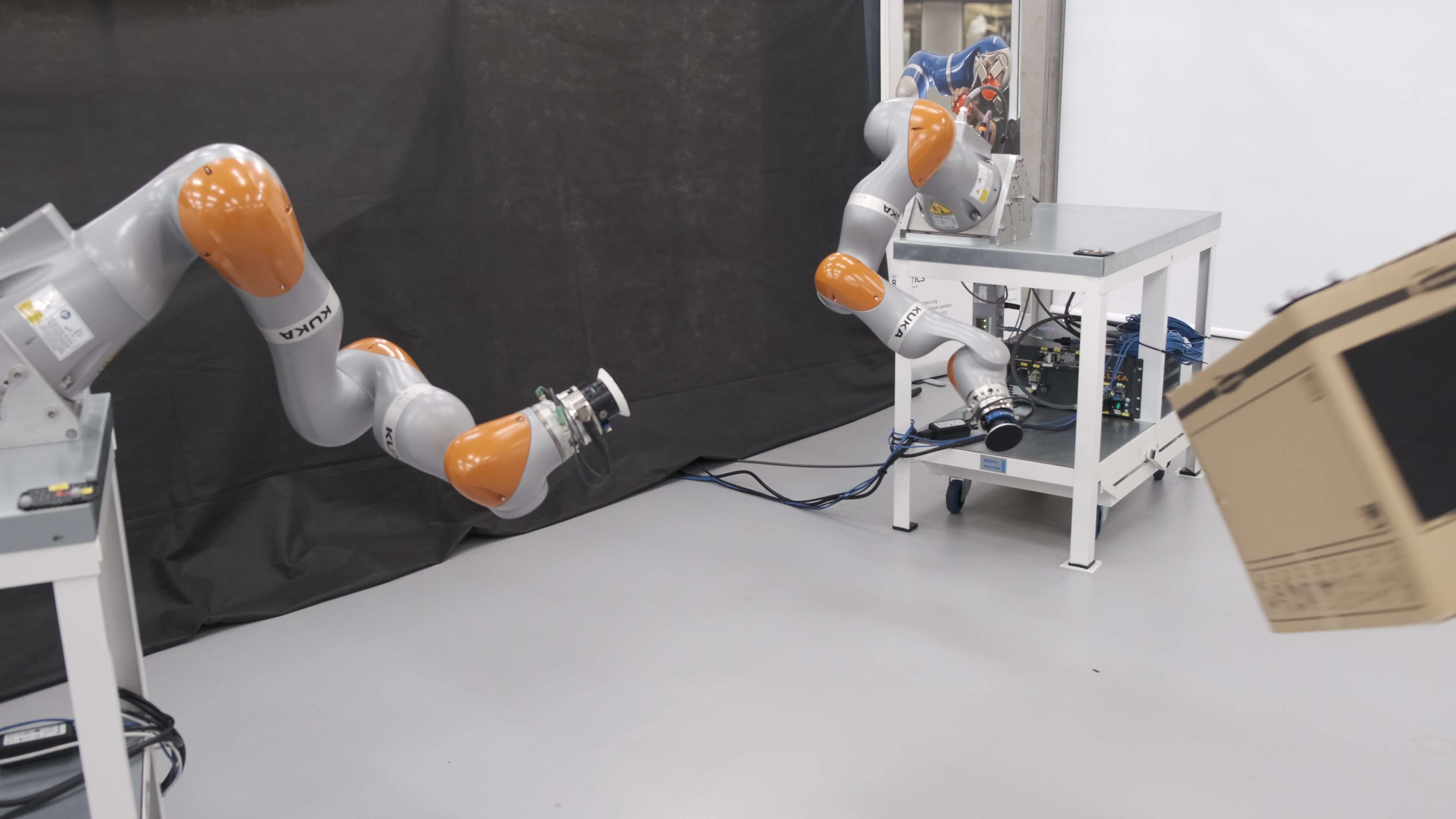}\label{fig:catching_rotating_object:moment1}%
        };%
        % Create scope with normalized axes
        \begin{scope}[%
        x={($0.1*(image.south east)$)},y={($0.1*(image.north west)$)}]%
            \draw[latex-,thick,blue] (8.3,3.2) -- ++(-1.0,-1.5) node[below,black,inner sep=0pt]{\small\textbf{Swinging object}};%
        \end{scope}%
        \end{tikzpicture}%
  }\,%
  \subfloat{\includegraphics[height=\fh,trim={0 0 25cm 0},clip]{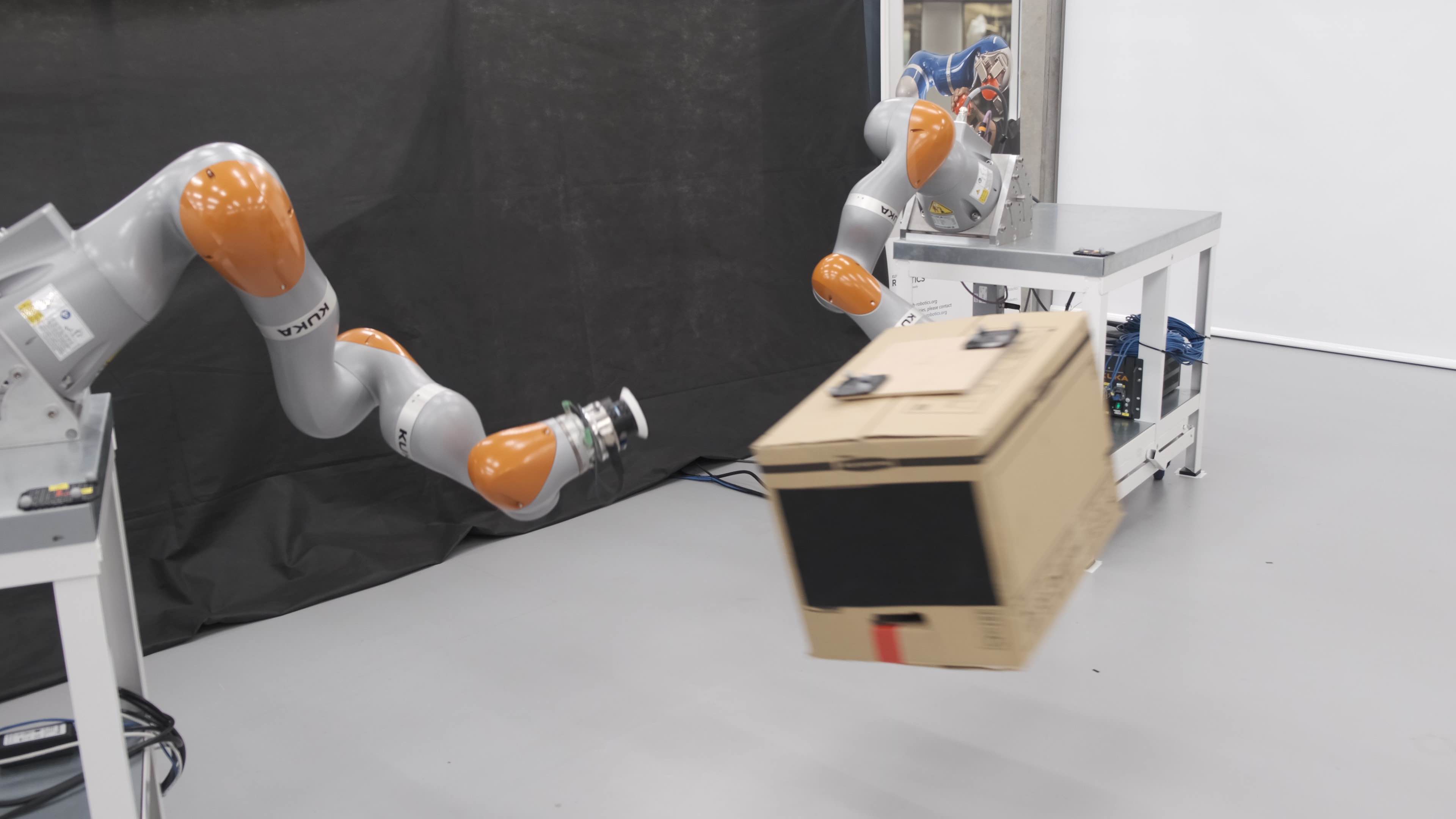}\label{fig:exp_rotating_objec:moment2}}\,%
  \subfloat{\includegraphics[height=\fh,trim={0 0 25cm 0},clip]{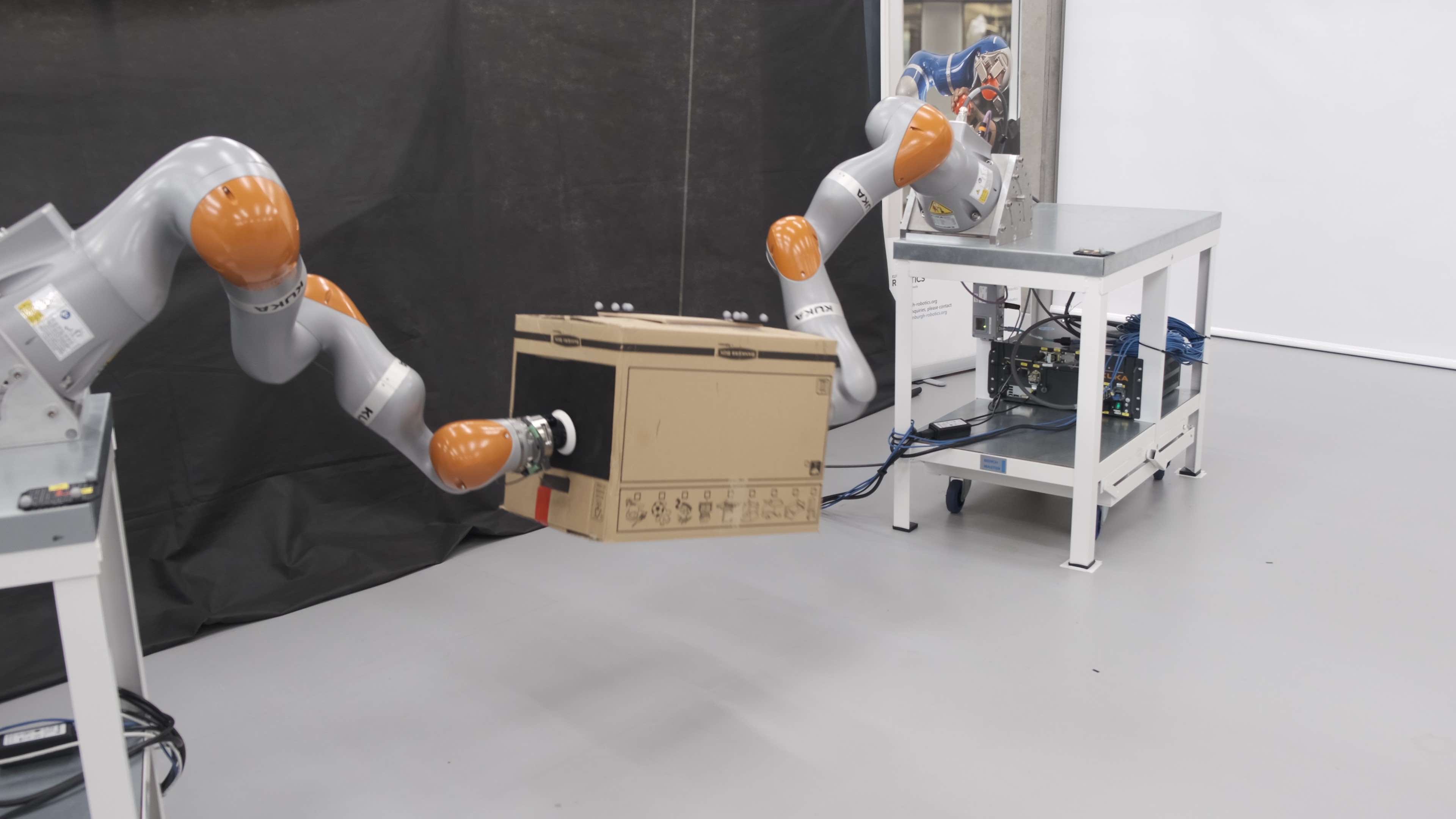}\label{fig:exp_rotating_object:moment3}}\,%
  \subfloat{\includegraphics[height=\fh,trim={0 0 25cm 0},clip]{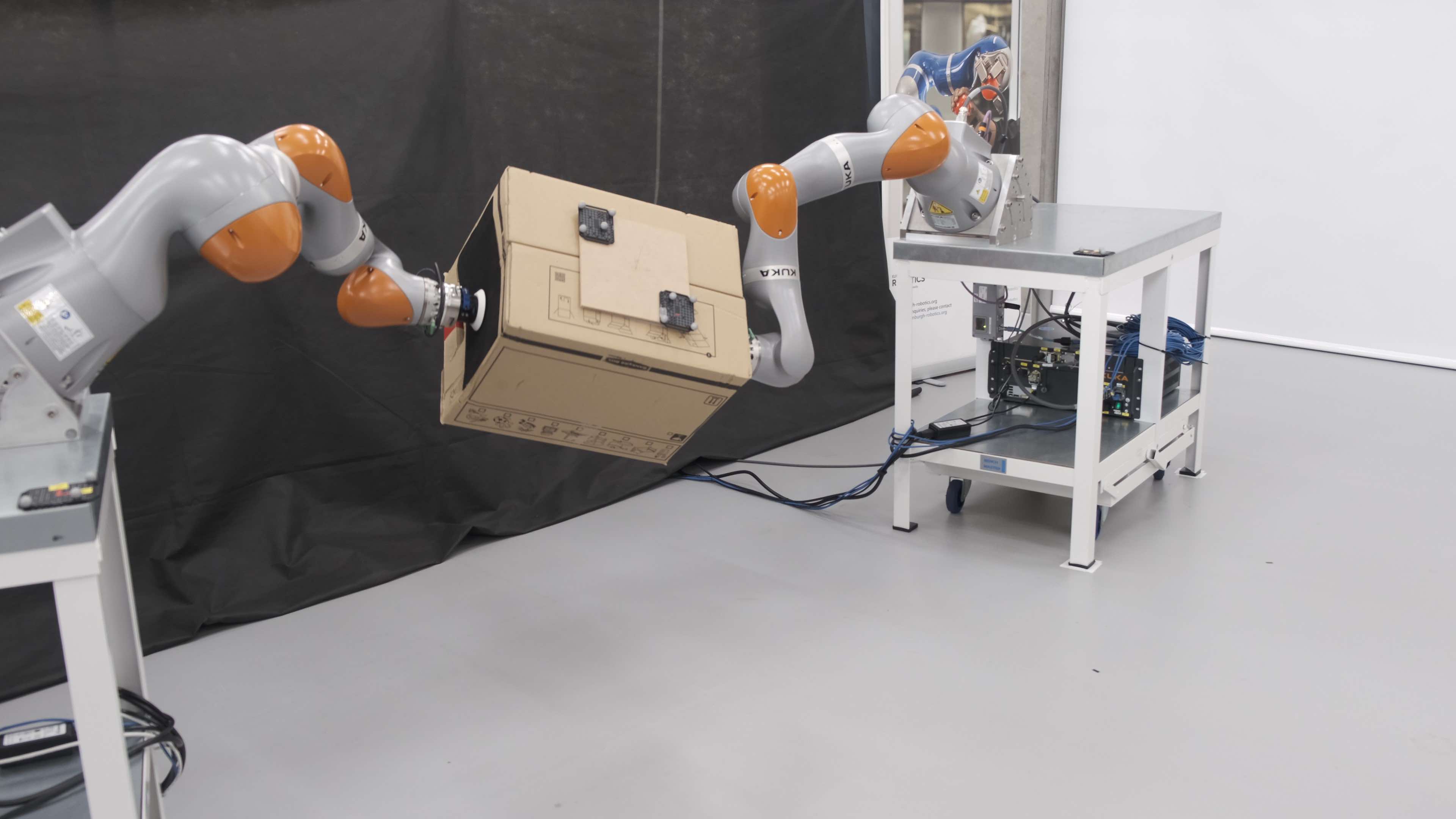}\label{fig:exp_rotating_object:moment4}}\\\vspace{-0.25cm}%
  \subfloat{\includegraphics[height=\fh]{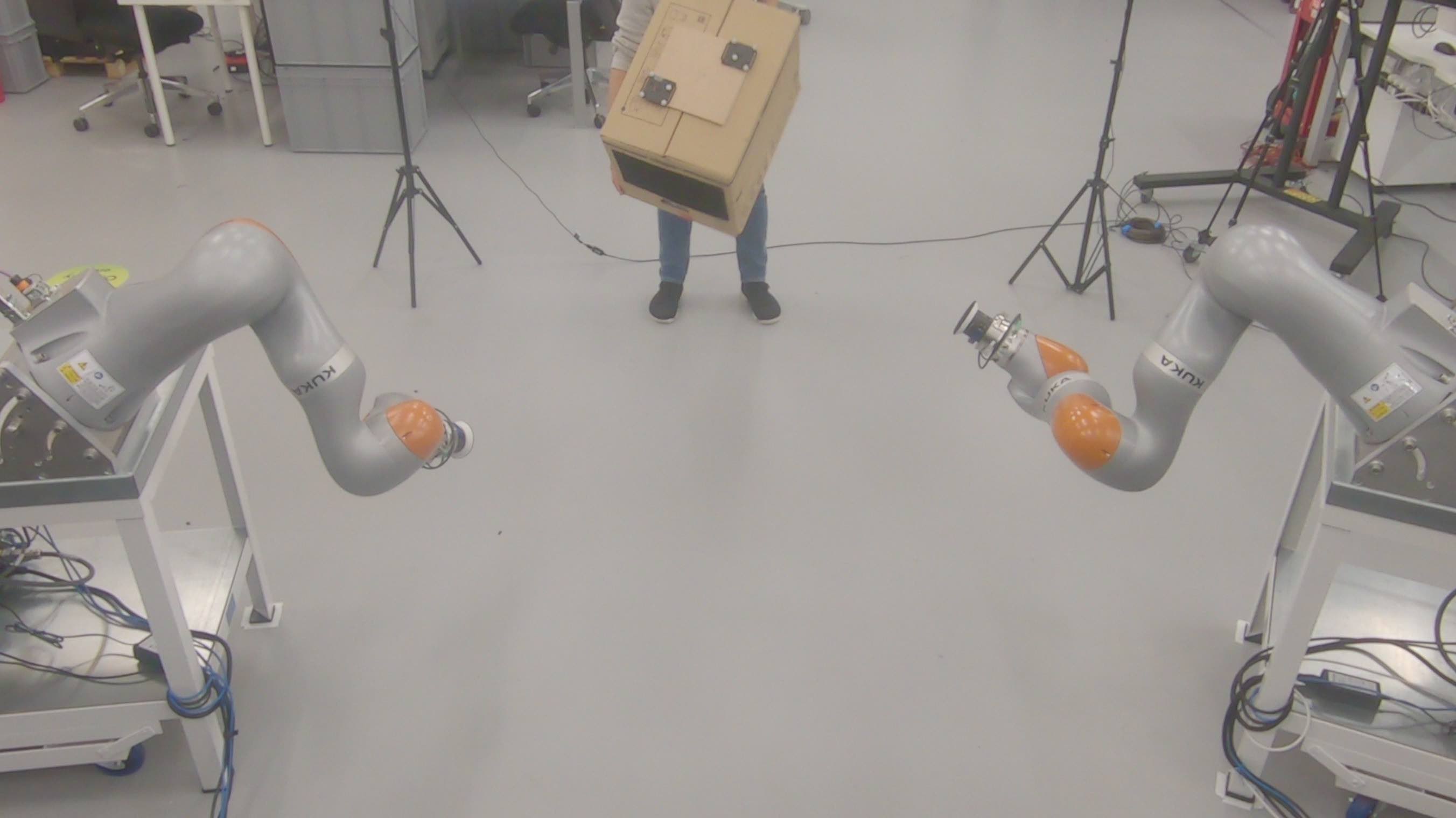}\label{fig:exp_rotating_object:top1}}\,%
  \subfloat{\includegraphics[height=\fh,trim={9cm 0 9cm 0},clip]{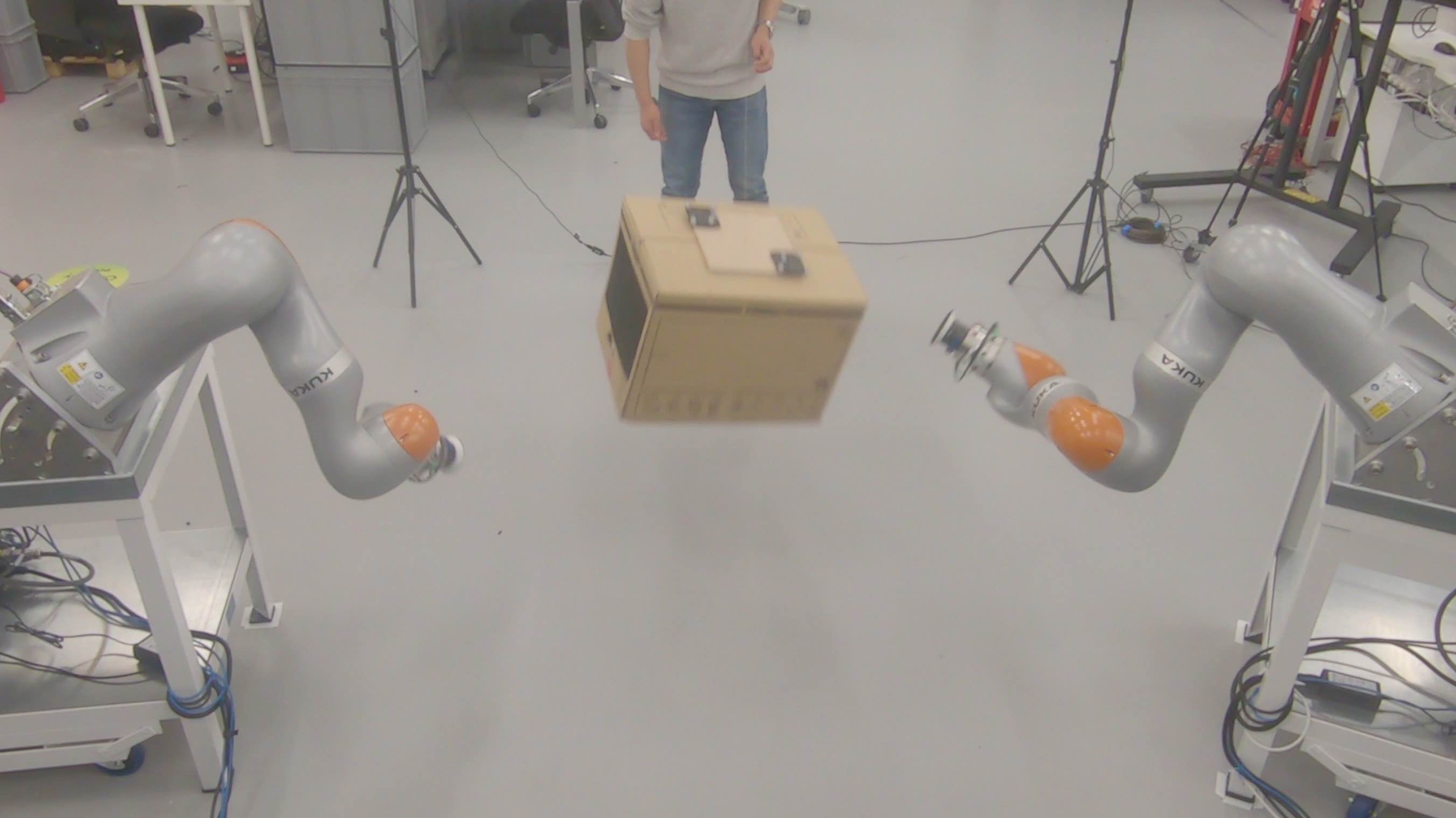}\label{fig:exp_rotating_object:top2}}\,%
  \subfloat{\includegraphics[height=\fh,trim={9cm 0 9cm 0},clip]{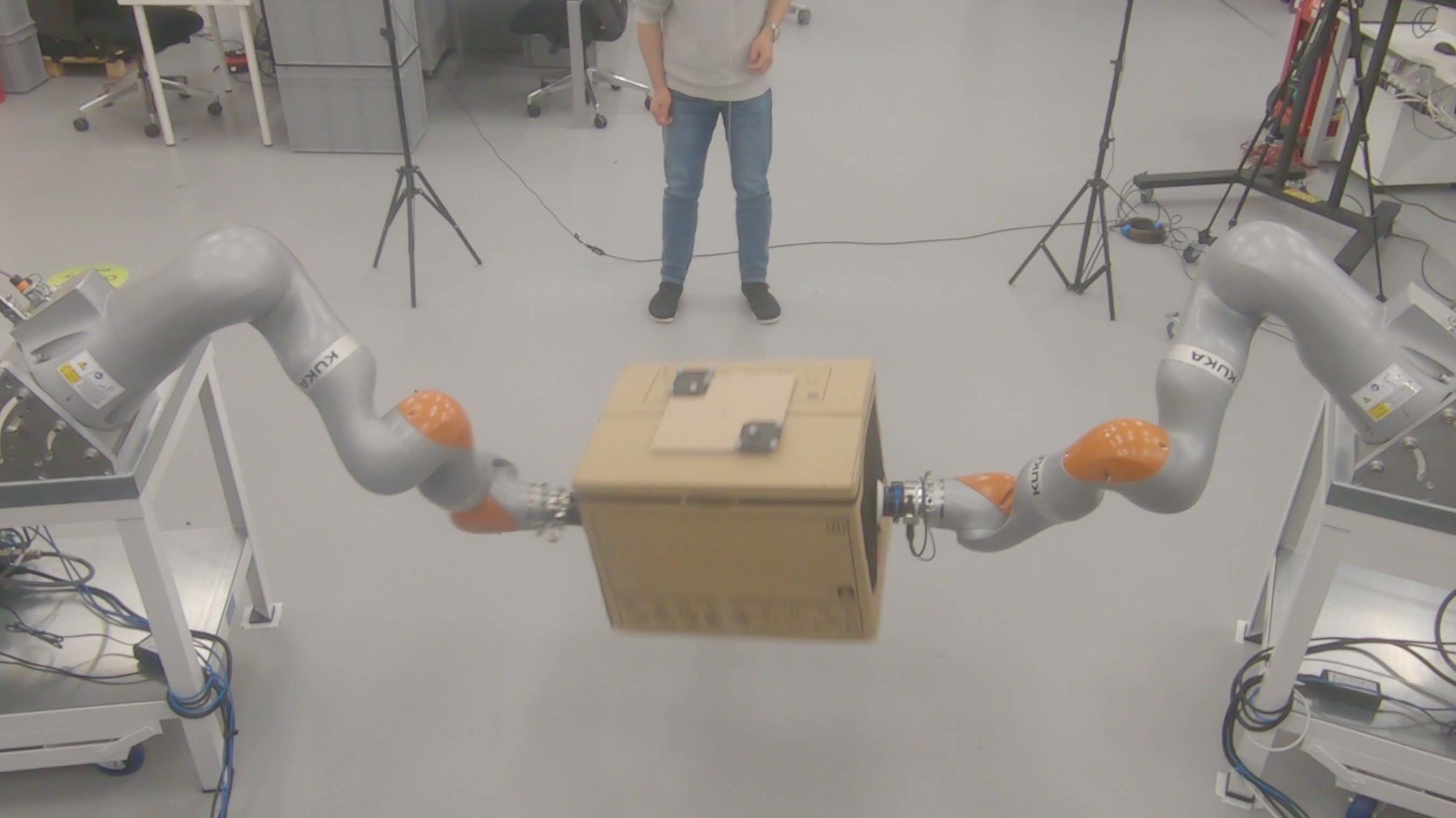}\label{fig:exp_rotating_object:top3}}\,%
  \subfloat{\includegraphics[height=\fh,trim={9cm 0 9cm 0},clip]{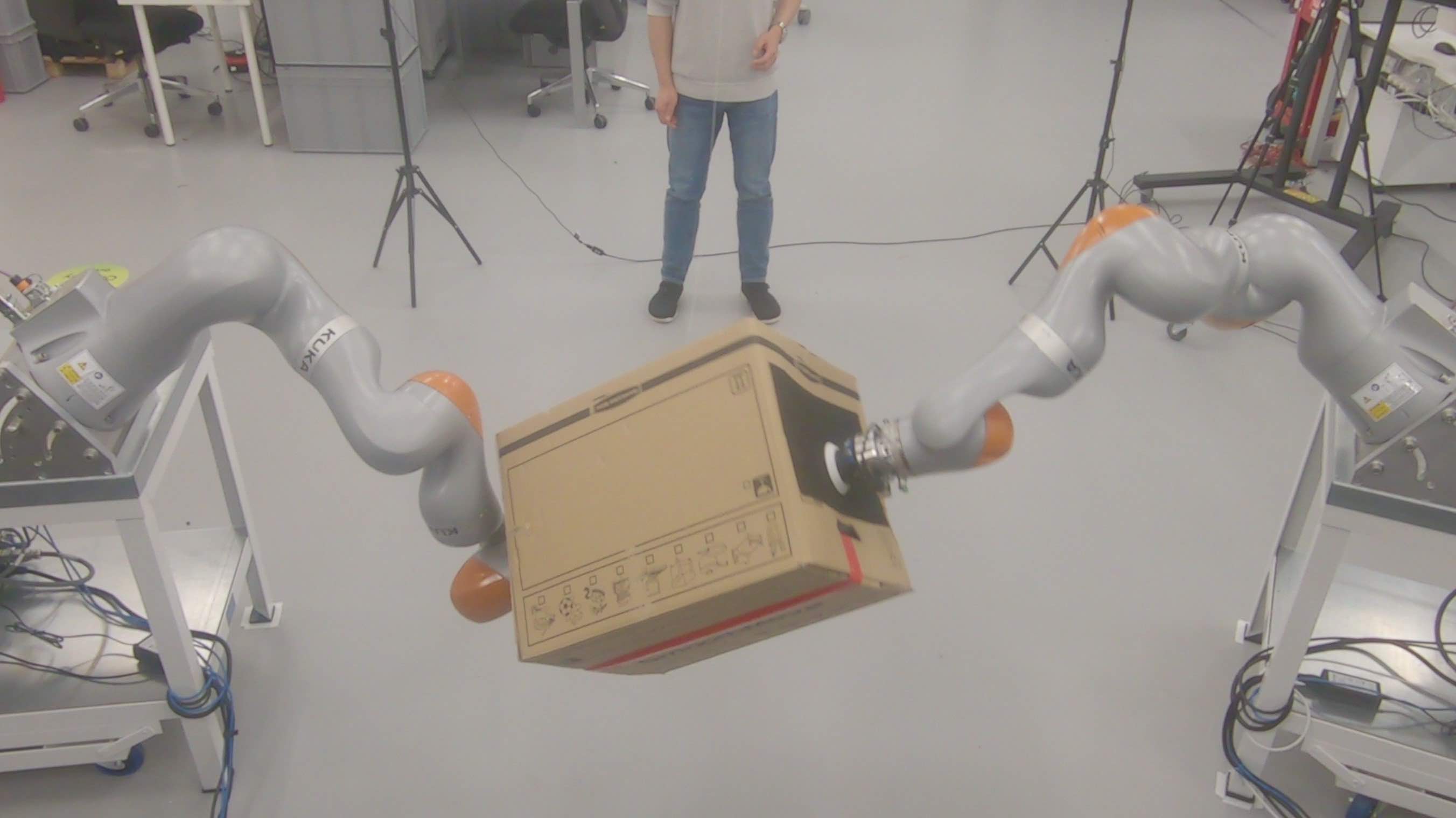}\label{fig:exp_rotating_object:top4}}%
\caption{Snapshots of the two KUKA-iiwa robots catching a swinging and tumbling object; (above) side-view camera, (below) top-view camera.}
\vspace{-0.1cm}
\label{fig:exp_rotating_object}
\end{figure*}

\begin{figure*}[t]
	\begin{center}
	\includegraphics[width=1.9\columnwidth]{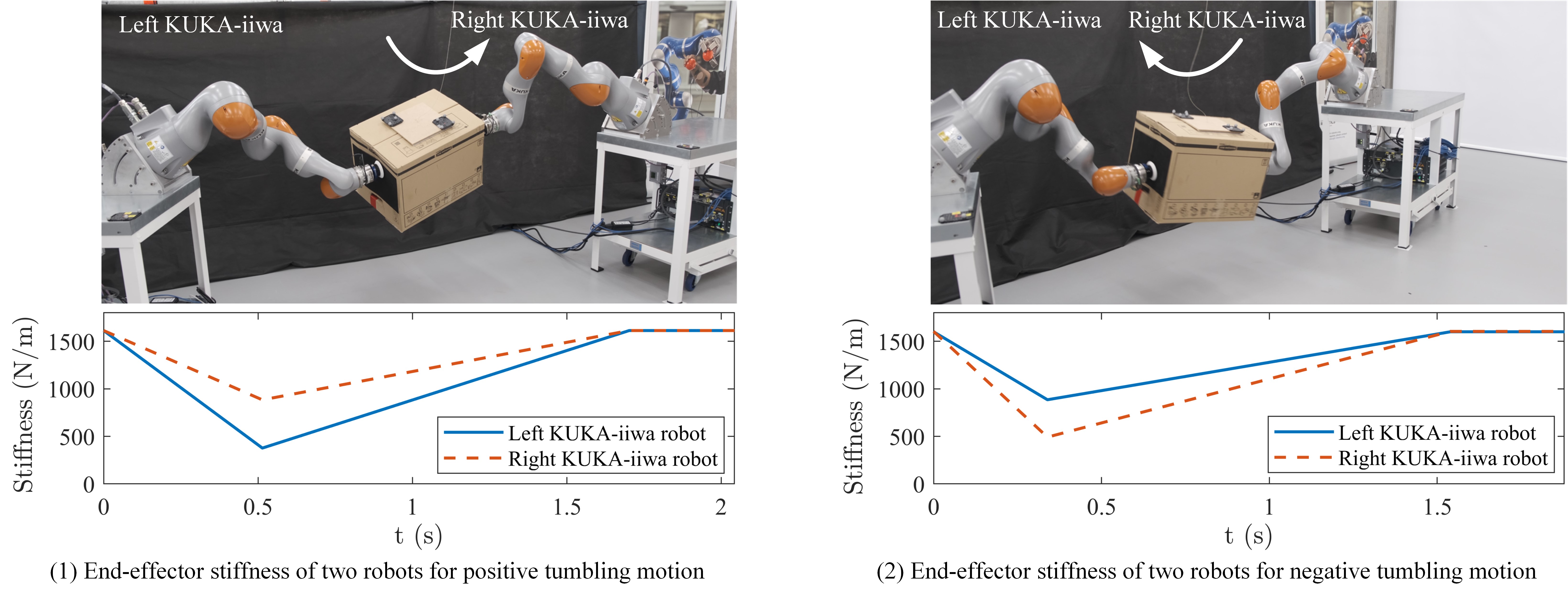}
	\vspace{-12pt}
	\end{center}
	\caption{End-effector stiffness of two robots for positive tumbling motion and negative tumbling motion.}
	\label{fig:stiffness4tumbling}
	\vspace{-10pt}
\end{figure*}
\vspace{-2mm}
\subsection{Stiffness Adjustment when Catching Tumbling Objects}
Here, we present a setup where the two \textit{KUKA-iiwa} robots catch a tumbling-swinging object that rotates with angular velocity $\geq 112 ^\circ/s$, as shown in~\cref{fig:exp_rotating_object}. This scenario encompasses the challenge of selecting contact locations on-the-fly, which is addressed using the CD-SQP (see~\cref{sec:contact_searching}). 

In addition, when respectively considering a positive and a negative tumbling motion as shown in~\cref{fig:stiffness4tumbling}, the dual-arm catching motion generated from the proposed algorithm is asymmetric. This implies that one of the two arms needs to contribute more to stop the motion of the object, which is also reflected in the planned optimal stiffness profiles. For the positive tumbling motion (see~\cref{fig:stiffness4tumbling}(a)), higher stiffness is planned for the right robot which actually decelerates the object, while for the negative tumbling motion the left robot has higher stiffness that is used to decelerate the motion of the object. 
It is worthy to note that the impact-aware MMTO algorithm is able to autonomously optimize the stiffness for each robot to achieve a soft catching behavior according to different initial conditions.

\vspace{-2mm}
\subsection{Catching Free-Flying Objects}
\begin{figure*}[t]
\centering
  \newcommand\fh{2.85cm}
  \subfloat{%
      \begin{tikzpicture}%
        \node[above right,inner sep=0, outer sep=0] (image) at (0,0) {%
            \includegraphics[height=\fh]{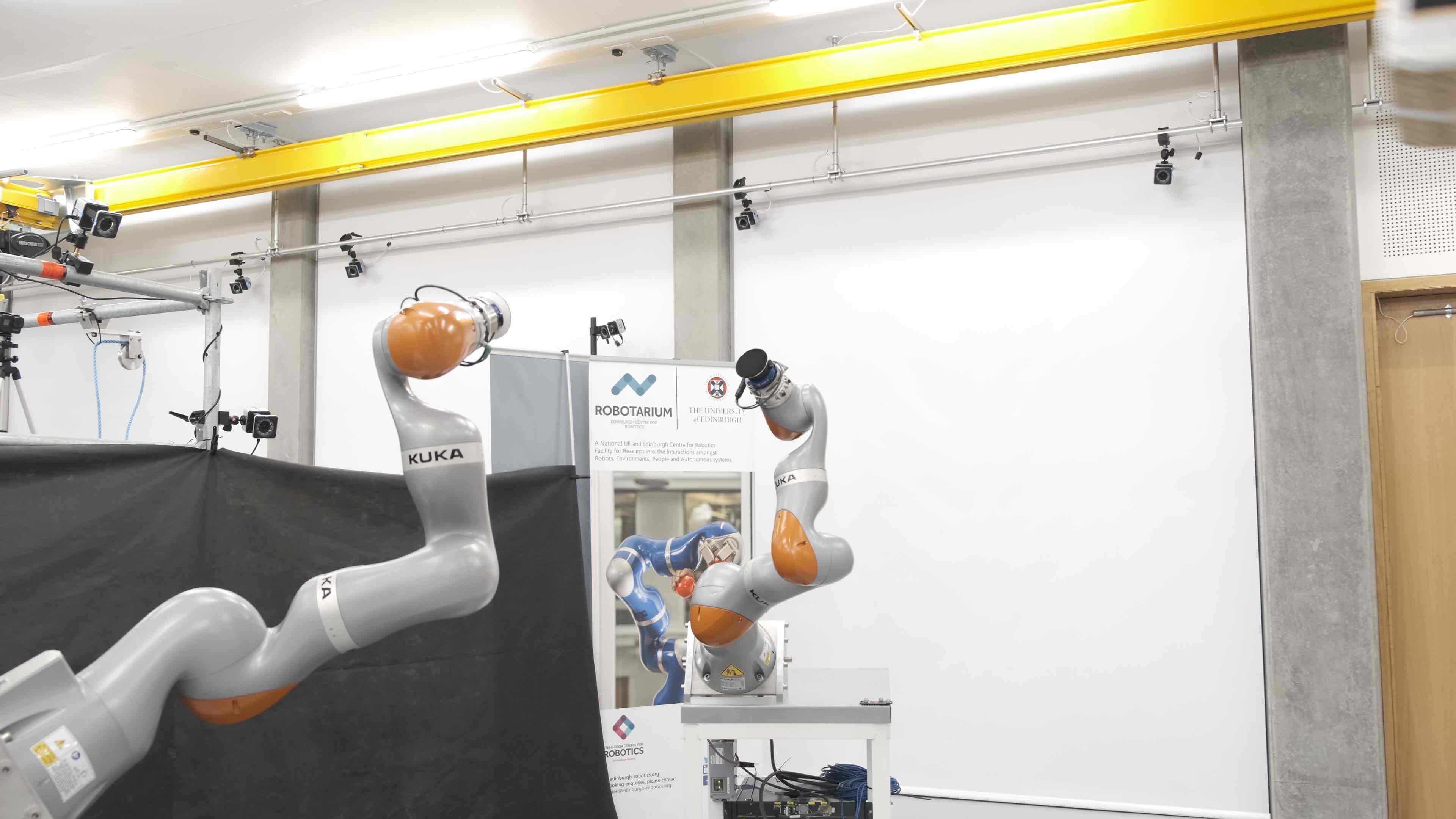}\label{fig:catching_flying_object:moment1}%
        };%
        % Create scope with normalized axes
        \begin{scope}[%
        x={($0.1*(image.south east)$)},y={($0.1*(image.north west)$)}]%
            \draw[latex-,thick,blue] (9.5,8.5) -- ++(-2.0,-2.5) node[below,black,text width=1cm]{\small\textbf{Flying object}};%
        \end{scope}%
        \end{tikzpicture}%
  }\,%
  \subfloat{\includegraphics[height=\fh,trim={0 0 25cm 0},clip]{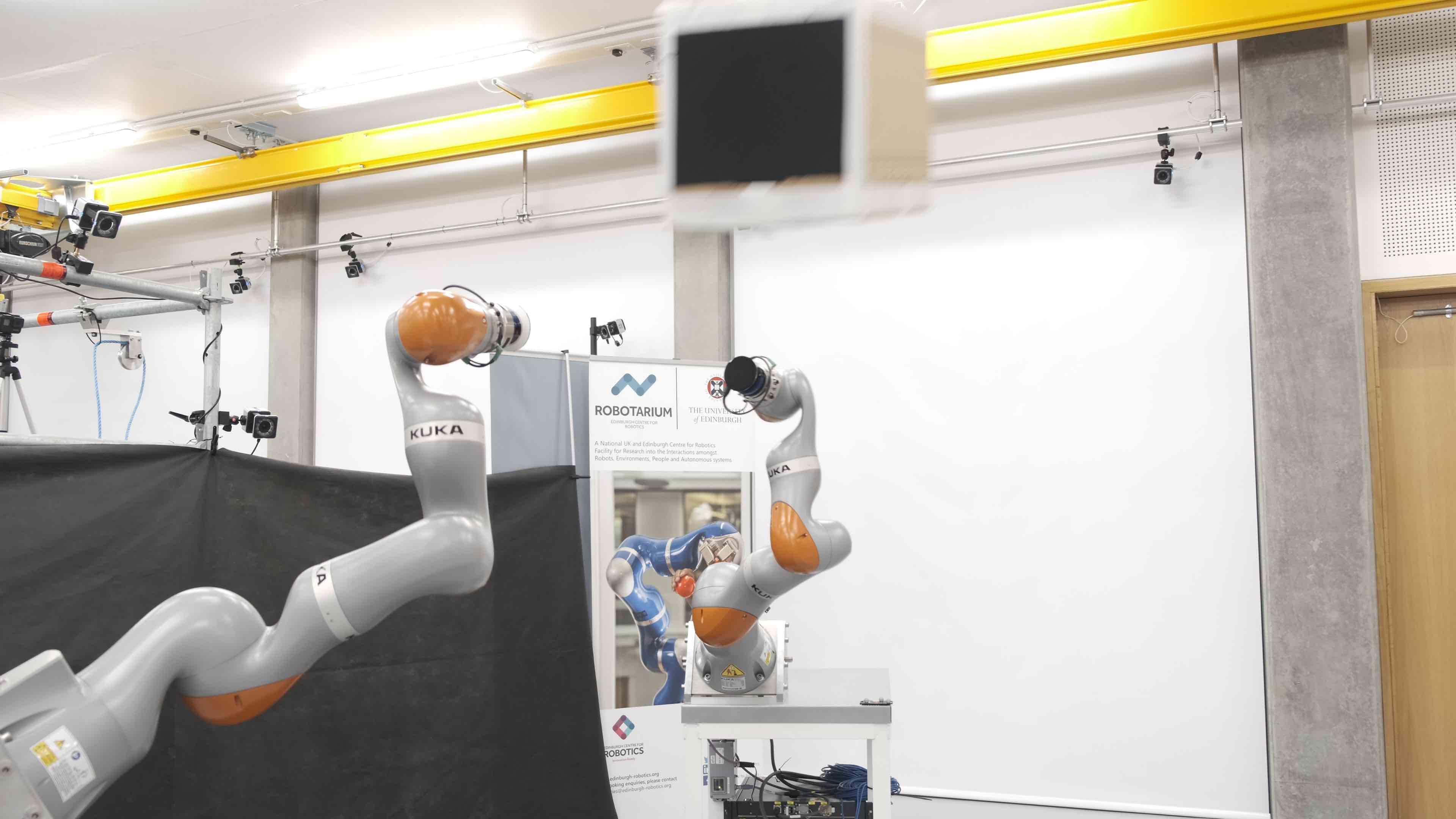}\label{fig:exp_flying_objec:moment2}}\,%
  \subfloat{\includegraphics[height=\fh,trim={0 0 25cm 0},clip]{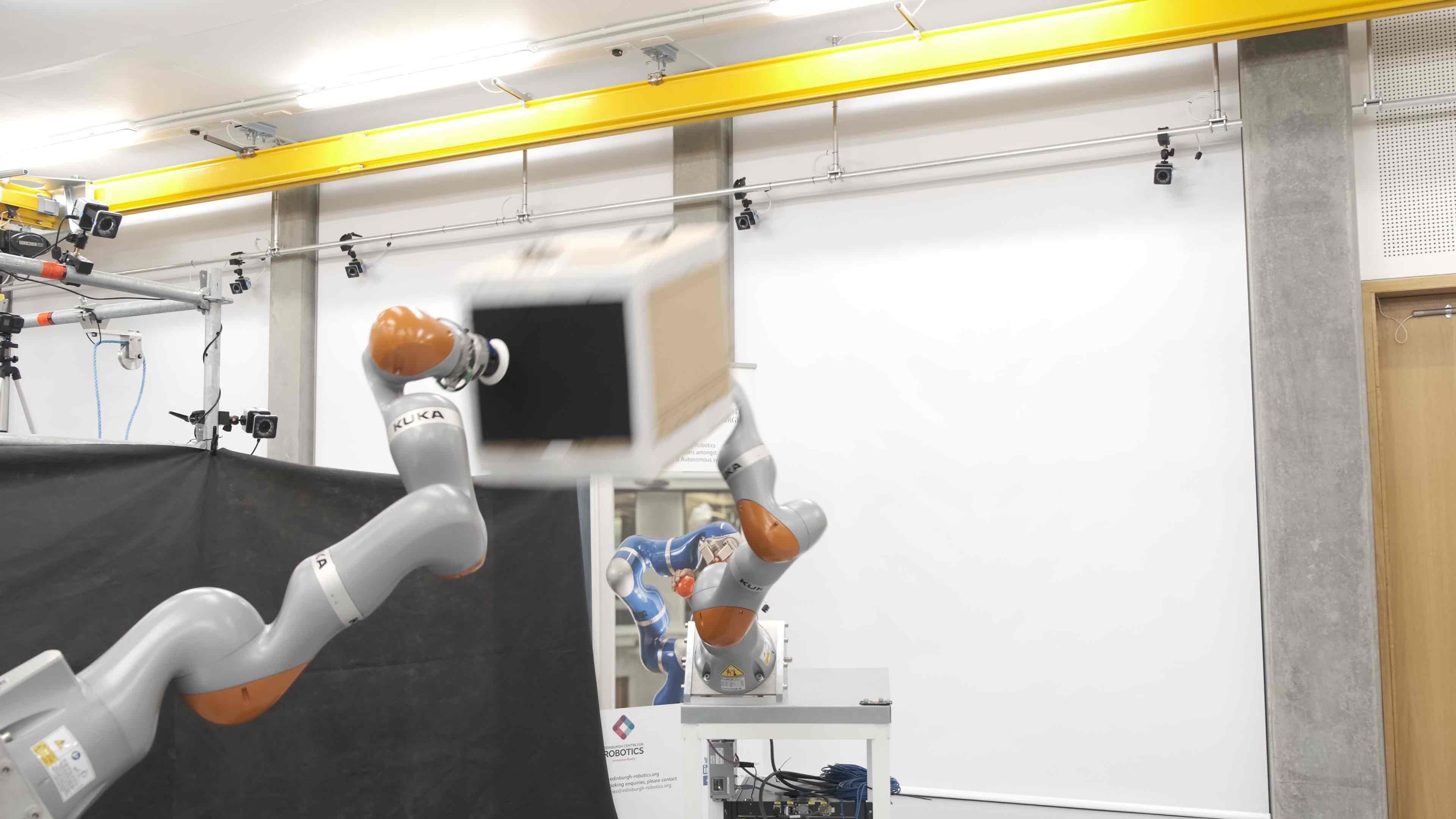}\label{fig:exp_flying_object:moment3}}\,%
  \subfloat{\includegraphics[height=\fh,trim={0 0 25cm 0},clip]{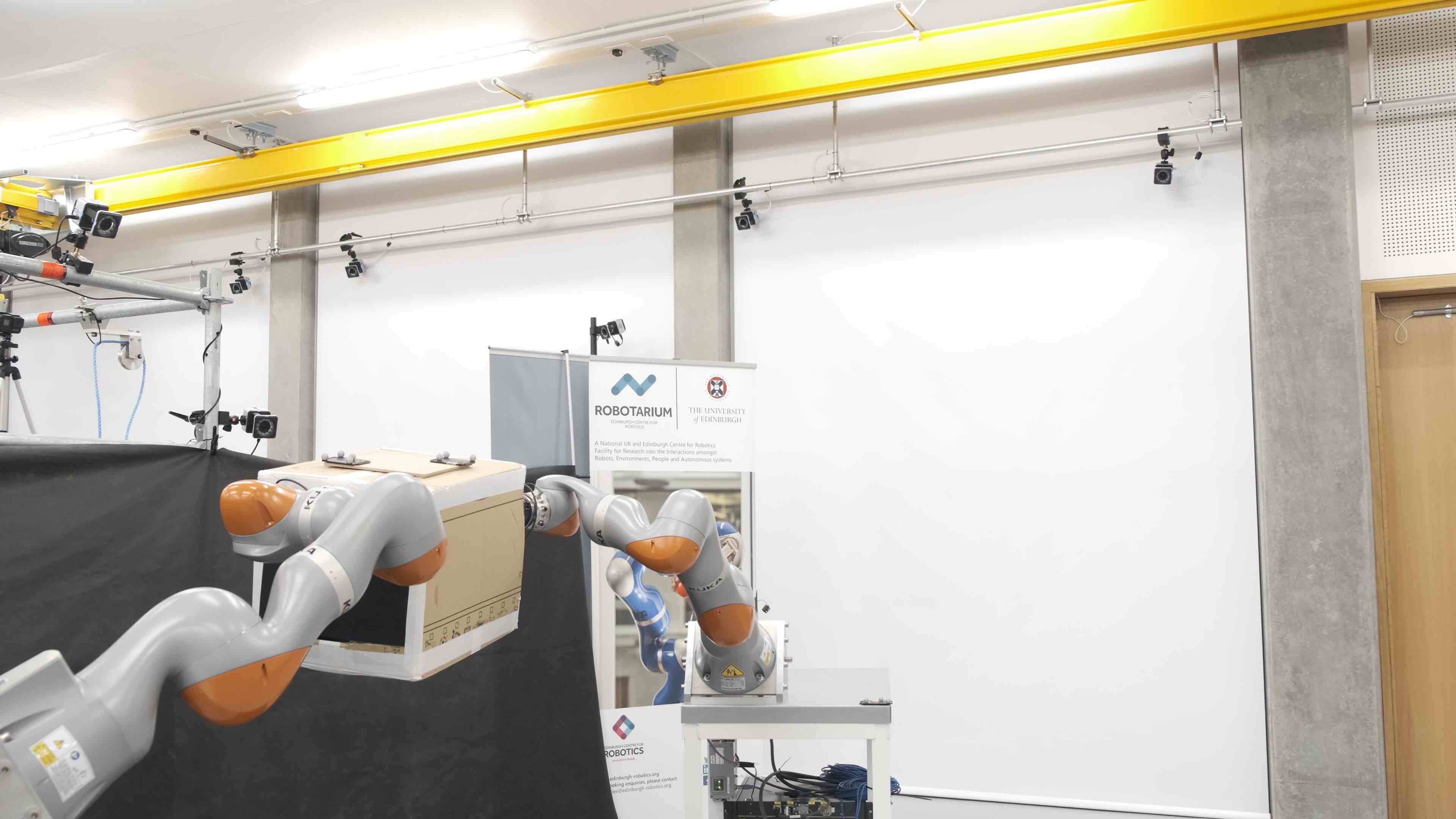}\label{fig:exp_flying_object:moment4}}\\\vspace{-0.25cm}%
  \subfloat{\includegraphics[height=\fh]{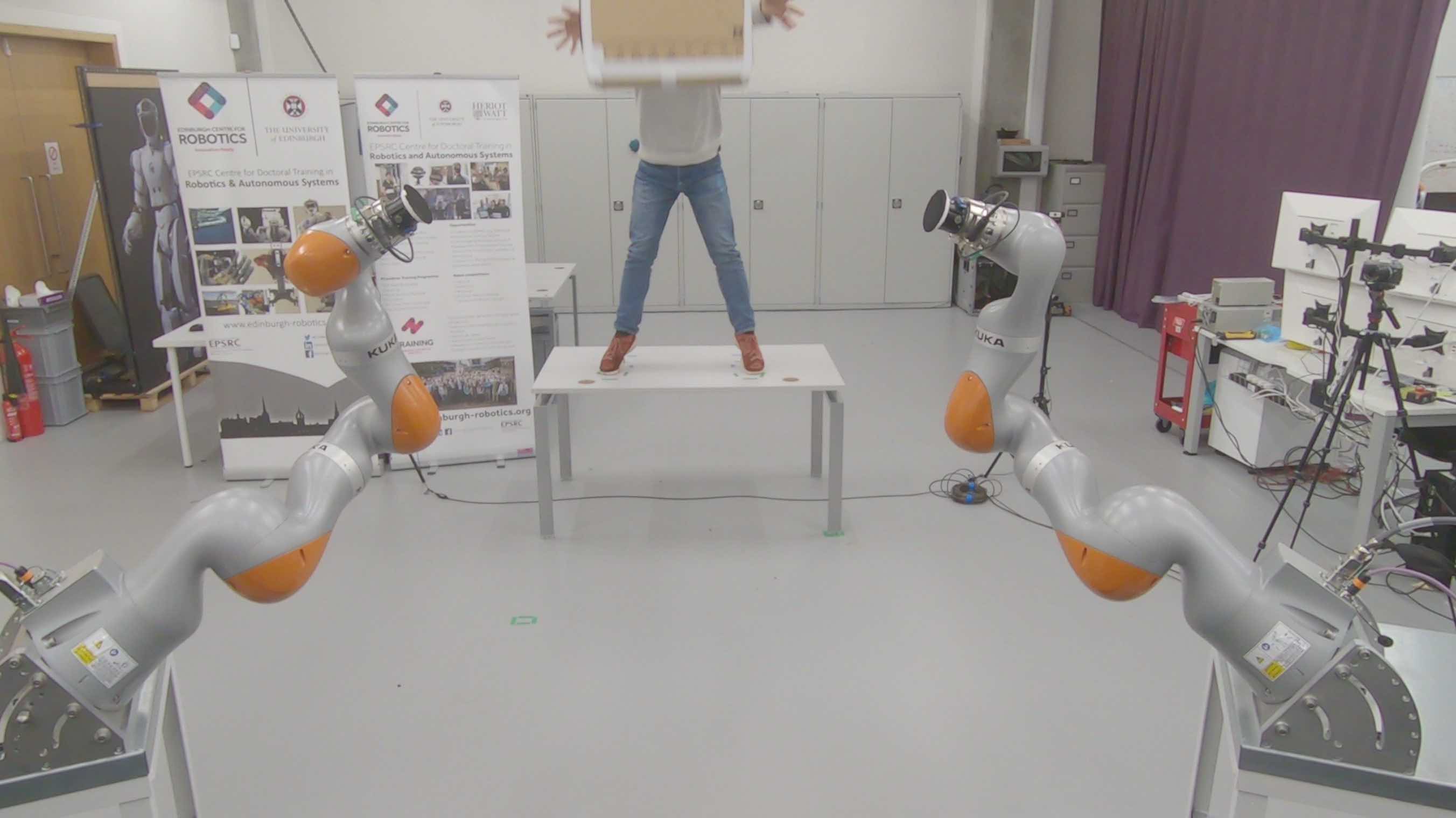}\label{fig:exp_flying_object:top1}}\,%
  \subfloat{\includegraphics[height=\fh,trim={7cm 0 11cm 0},clip]{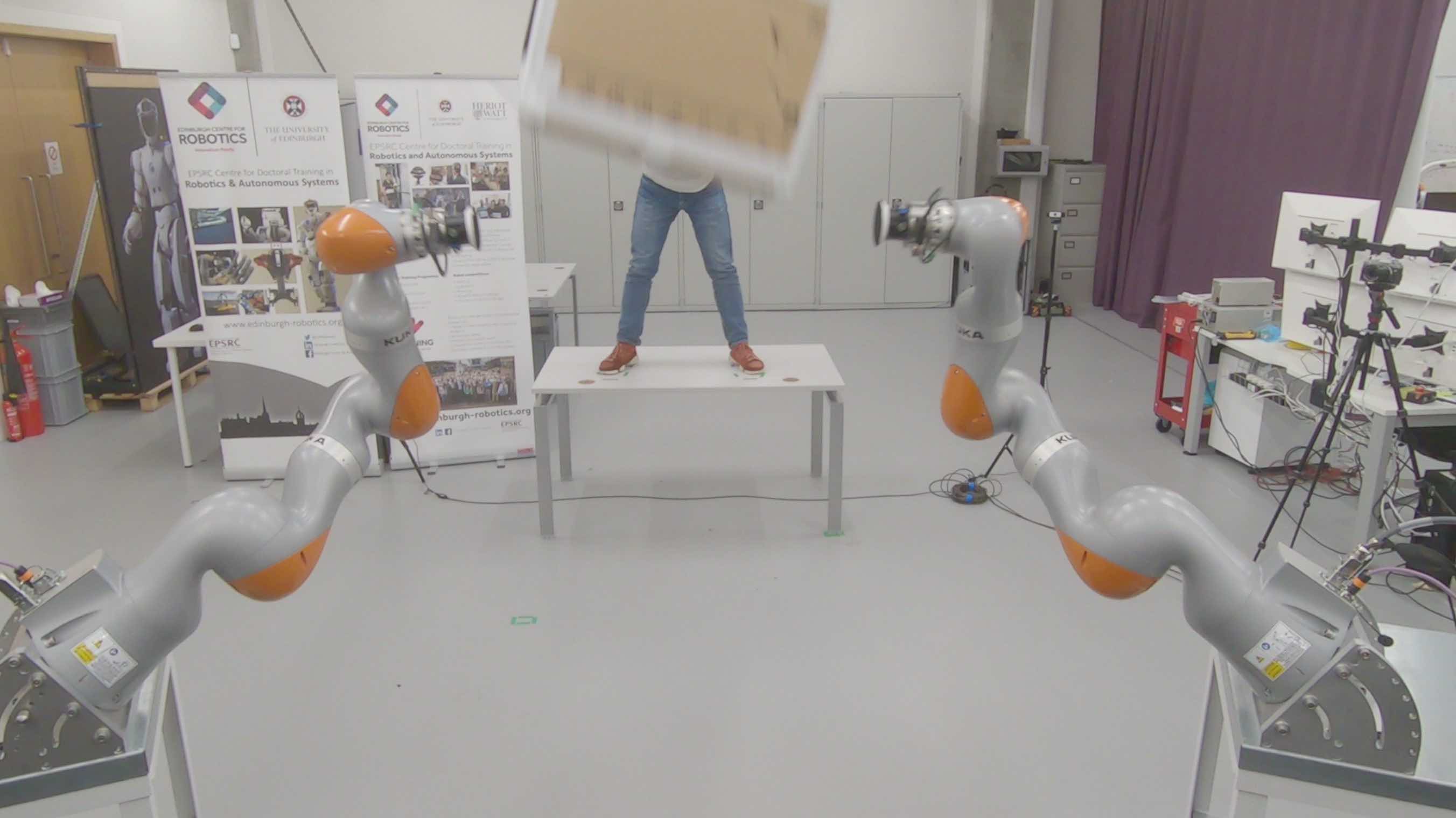}\label{fig:exp_flying_object:top2}}\,%
  \subfloat{\includegraphics[height=\fh,trim={7cm 0 11cm 0},clip]{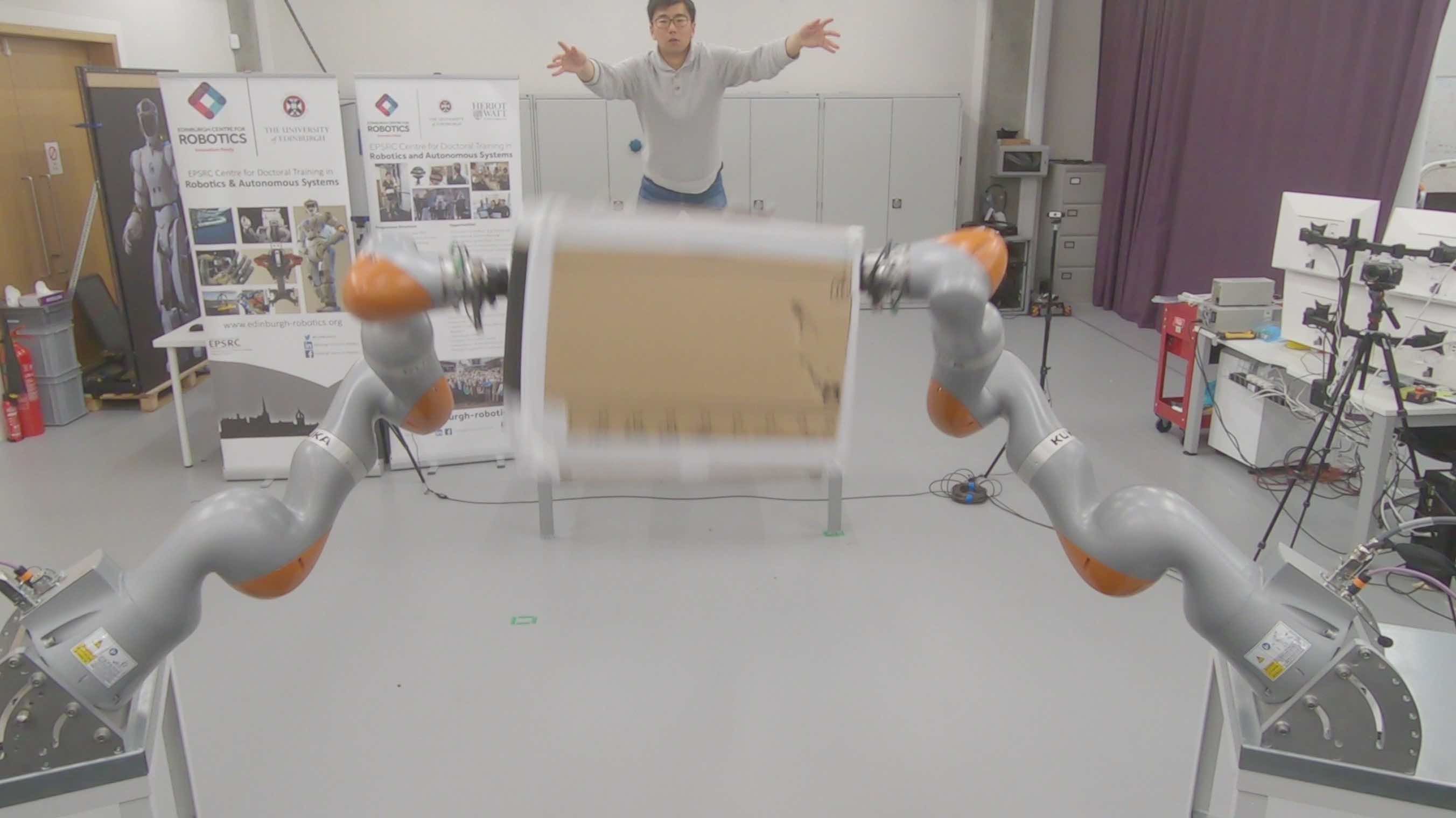}\label{fig:exp_flying_object:top3}}\,%
  \subfloat{\includegraphics[height=\fh,trim={7cm 0 11cm 0},clip]{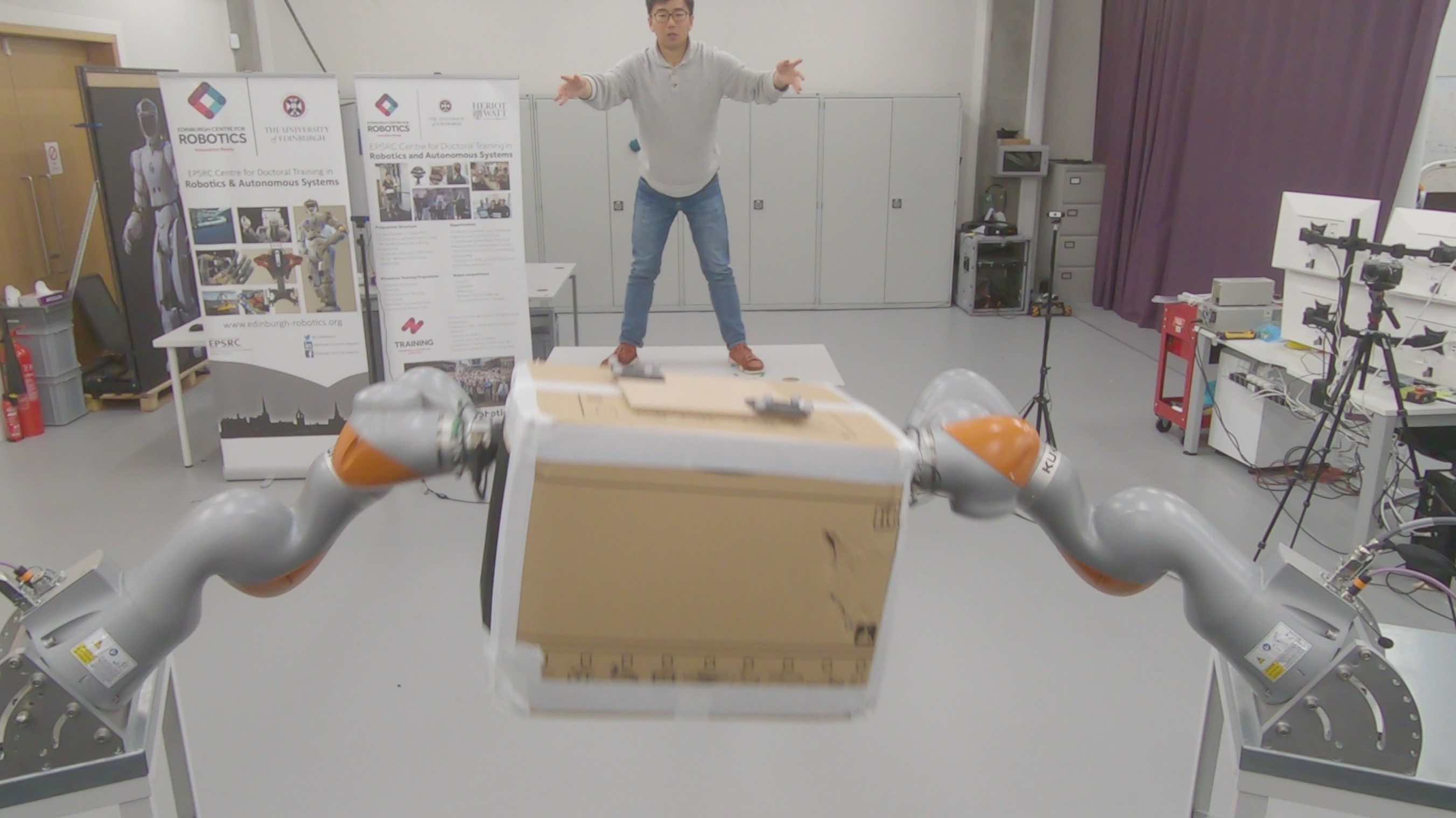}\label{fig:exp_flying_object:top4}}%
\caption{Snapshots of the two KUKA-iiwa robots catching a flying object thrown by an human; (above) side-view camera, (below) top-view camera.}
\vspace{-3mm}
\label{fig:exp_flying_object}
\end{figure*}

In order to further validate and demonstrate the capabilities of the proposed system, we consider the scenario of a free-flying object, as shown in~\cref{fig:exp_flying_object}. The object is thrown approximately $2.7 m$ away from the robots and close to the moment of contact, travels with linear velocity $\geq$\SI{3.5}{\meter\per\second} on y axis and $\geq$\SI{2}{\meter\per\second} on z axis of the motion. For this scenario, the system performs \textit{estimation}, \textit{prediction}, \textit{contact selection} and \textit{hybrid planning} in less than $350ms$. The catching motion is truly at the limits of the robot hardware with the end-effectors of the arms reaching speeds higher than $2m/s$. Yet, as it is demonstrated, the two \textit{KUKA-iiwa} robots are able to catch a flying object that weighs~\SI{4.2}{\kilogram}. \textcolor{black}{Although the two robots can catch the tumbling and flying objects, 
%the success rate \textcolor{blue}{($\sim 20 \%$)} is much lower than the scenarios in previous sections. \textcolor{blue}{Please note that large part of the failed attempts were due 
we found that the robots frequently go to E-stop due to the very high accelerations and speed of the motions.} For this kind of extreme manipulation tasks, we encounter limitations that emerge not only from the computation time of the proposed algorithm but also the maximum payload and speed of the robot hardware. We will discuss the limitations in detail in~\cref{sec:discussion}.  

% Until now we have not seen any limitation neither on the robot hardware nor on the proposed method. 

%% \begin{figure*}[t]
%% 	\begin{center}
%% % 	\includegraphics[width=0.99\columnwidth]{figures/catching_rotating_object.pdf}
%% 	\includegraphics[width=1.99\columnwidth]{figures/catching_rotating_object.png}
%% 	\vspace{-12pt}
%% 	\end{center}
%% 	\caption{Two KUKA-iiwa robots catching a swinging and tumbling object.}
%% 	\label{fig:exp_rotating_object}
%% 	\vspace{-10pt}
%% \end{figure*}

%% \begin{figure*}[t]
%% 	\begin{center}
%% % 	\input{figures/catching_flying_object}
%% 	\includegraphics[width=1.99\columnwidth]{figures/catching_flying_object.png}
%% 	\vspace{-12pt}
%% 	\end{center}
%% 	\caption{Two KUKA-iiwa robots catching a flying object thrown by a human.}
%% 	\label{fig:exp_flying_object}
%% 	\vspace{-10pt}
%% \end{figure*}

% %===============================================================================
\section{Discussion}\label{sec:discussion}
% Both simulation and experiment results in previous sections show that the robot can make a stable contact with a large-momentum object (heavy object moving at fast speed) and achieve impact-aware manipulation (a soft catching behavior). 
% However, there are still some limitations which hinder the performance and capability of our proposed method.
Here, we discuss practical considerations, some limitations that hinder the performance and capabilities of the proposed system and possible extensions of the proposed work.

\subsubsection{IK with joint limit constraints} Throughout the experiments, we found out that the feasible volume where catching can be realised is very small due to the limited coordinated dual-arm workspace, the joint limits of the robots, singularity configurations and the fast moving object. These made it difficult to demonstrate the more extensive capabilities of the proposed algorithm, as for each scenario we needed to carefully choose the initial configurations of the robots \textcolor{black}{to avoid the robots stopping in the middle of the motion due to joint limits.} An IK algorithm that can avoid joint limits and singularities could further enhance the robustness and success rate of the proposed system.

\subsubsection{Tracking the planned motions} 
Given the very high speed of the moving object, the two robots need to accelerate with the maximum acceleration 
to follow the moving object at their maximum speed and catch it. Following the object motion requires accurate tracking of the planned motions (output of MMTO), which deteriorates as: (i) each robot can not immediately reach its maximum velocity, due to its own mass and inertia, and (ii) accurate position tracking with a compliance controller, such as the Cartesian impedance control mode of the \textit{KUKA-iiwas}, cannot be achieved out-of-the-box. As a result, the mismatch between the planned motion and the actual motion of the robots \textcolor{black}{can result in real-world catching failures, due to the mismatch at the moment of contact. }
%degrades the real-world catching performance. 
One could improve the tracking performance with a Model Predictive Controller that has a prediction horizon and takes into account the hardware specifications, as well as any discrepancies resulting from the impedance controller.

% crucial to the bimannual catching task. In order to avoid particularly designing initial configuration for each task, an fast IK algorithm with joint limit consideration need to be further investigated.
 % to avoid singularity during the motion execution. 
% Not only the initial configurations of the robots

\subsubsection{Computation time} 
To catch a large-momentum object, the allowed computation time to estimate and predict the object's motion and plan the motion of the arms is limited. For the scenarios considered in this paper, it is less than $0.5s$.
Both the IVP and the MMTO are formulated as non-linear programming and are solved with \textit{Ipopt}.
Hence, in order to meet the allowed computation time bounds, we need to provide appropriate initial guesses.
However, for highly-dynamic manipulation task that involve many variations and uncertainties, such as catching an object that swings or that is thrown by a human, generating an appropriate initial guess might not be easy.
Therefore, investigation for more efficient solvers and problem formulations, as well as methods to store and quickly obtain robust initial guesses for warm-starting could be very beneficial towards achieving on-line computation reliably.

% the feasible catching time for the robot is limited, which further limited the available computation time for the impact-aware MMTO. As the NLP solver (\textit{Ipopt}) is adopted to solve the non-linear MMTO problem, which is time-consuming, 
% s normally far away from the solution of the actual problem, which makes the optimization problem difficult to converge. 
% we provide an initial guess for solving MMTO in order to satisfy online computation requirement % (approximately ? $ms$) 
% in all the experiments. 

\subsubsection{Hardware}
Last but not least, the catching task requires high load-to-weight ratio robots. A more lightweight robot would have better performance for such dynamic manipulation tasks, yet it should also have large payload capability. Another useful extension of the system would be to mount each \textit{KUKA-iiwa} onto a mobile base to enlarge their limited workspace. 

\textcolor{black}{In summary, the proposed framework is a first attempt to obtain a feasible online optimization approach that can plan and execute catching motions while considering impulsive forces and slippage limits. Further, more detailed impact dynamics modeling is encouraged and required towards achieving more accurate impact predictions.}

% Mobile base more dynamic robot arms.
% \textcolor{blue}{Generally speaking, the robot with heavier payload will have corresponding larger mass and inertia. As the moving object approaches two KUKA-iiwa robots with high speed, the two robots need to reach their maximum velocity with maximum acceleration to follow and catch the moving object. However, due to the large mass and inertia, the robot has limited acceleration and can not immediately reach its maximum velocity. The mismatch between the planned motion and the actual motion deteriorate the catching performance. }

% Hence, the robot with high load to weight ratio is further needed for highly dynamic motion, such as catching large-momentum flying object.

% %===============================================================================
\section{Conclusion}\label{sec:conclusion}

% \lei{what did we do/ what is the contribution/ limitation/ future work.}
In this paper, we present an optimization framework for catching
large-momentum objects whose total movement duration is less than \textcolor{black}{one second} and with speeds higher than~\SI{3.5}{\meter\per\second}. Further, we developed the first ever system able to realise \textcolor{black}{dynamic} bimanual catching of flying objects. This includes motion estimation and prediction, contact selection, multi-mode trajectory optimization, variable-stiffness modulation and motion execution on two \textit{KUKA-iiwa} robots. The executed motions are at the maximum speed limit of the hardware.  
In our optimization framework, based on the 3D compliance impact model, we propose impact-aware contact selection and impact-aware motion planning in order to mitigate impulsive forces when the robots make contact with the moving object. A variety of bimanual catching simulations and real-world experiments validate and demonstrate the effectiveness of the proposed methods and system. The scenarios extend from grasping a moving object on a conveyor belt to catching a tethered tumbling object, and even catching a flying object thrown by a human. 

\textcolor{black}{Our future work will focus on enabling our MMTO formulation to modify the contact sequence schedule with the object (continuing our work beyond simultaneous contacts for the two arms) as well as to achieve on-line computation with offline knowledge, towards demonstrating different type of dynamic bimanual manipulation behaviours. }

% , for the MMTO formulated in this paper we assume the two arms make contact with the object simultaneously, which is quite unrealistic and restricted for the real applications. For future work, we are interested in 

% integrating the contact sequence of the manipulators within the multi-mode TO framework, and further demonstrate different type of dynamic bimanual catching. 

% %===============================================================================

%%%%%%%%%%%%%%%%%%%%%%%%%%%%%%%%%%%%%%%%%%%%%%%%%%%%%%%%%%%%%%%%%%%%%%%%%%%%%%%%
\begin{appendices}
    \crefalias{subsection}{appendix}
    \section*{APPENDIX}
    \subsection{Energy-based 3D Impact Model}
\label{appendix:impact_modelling}
%\theo{Boss, why do we have details of the model here in the appendix and we don't just refer to the paper? Did we modify something? For completeness? or for clarity, knowing that the paper from Jia is not that clear...?}
%Inspired by the energy-based impact modelling with tangential compliance~\cite{jia2013three}, in this paper the same impact model is used to analyse the impact along normal and tangential directions. The impact occurs when a manipulator makes contact with a moving object. 
In this appendix, we will briefly introduce the energy-based impact dynamics model -- more details can be found in~\cite{jia2013three}.

Based on the 3D compliance impact model with three spatial springs, the energy stored in these three
springs during impact can be calculated. From the three components of the energy, we can determine whether a contact is in slipping mode or sticking mode. Consequently, the impulse and the state of the object are updated till the end of the impact.

\subsubsection{Impact modelling with three virtual springs}
By modelling the contact between two bodies with three virtual springs along three unit orthogonal vectors $\hat{\bm x}$, $\hat{\bm y}$ and $\hat{\bm z}$, we can decompose the impulse $\bm \Lambda$, contact force $\bm F$, and contact velocity $\bm v$ along the three directions:
\begin{equation}\label{eqn:impact_decompositon_impulse}
{\bm \Lambda} = {\Lambda_x} \hat{\bm x} + \Lambda_y \hat{\bm y} + \Lambda_z \hat{\bm z},
\end{equation}
\begin{equation}\label{eqn:impact_decompositon_force}
{\bm F} = {F_x} \hat{\bm x} + F_y \hat{\bm y} + F_z \hat{\bm z},
\end{equation}
\begin{equation}\label{eqn:impact_decompositon_velocity}
{\bm v} = {v_x} \hat{\bm x} + v_y \hat{\bm y} + v_z \hat{\bm z}.
\end{equation}
where $\hat{\bm z}$ is the normal vector of contact tangential plane; $\hat{\bm x}$ is the opposite direction of the tangential component of initial contact velocity $\bm v_{tan}$, where $\bm v_{tan}= \bm v - v_z \hat{\bm z}$; and $\hat{\bm y} = \hat{\bm z} \times \hat{\bm x}$.

At the same time, the changes in length of the three springs are respectively denoted as $u$, $w$, and $n$. Correspondingly the rates of change of the three springs are denoted as $\dot u$, $\dot w$, $\dot n$. Therefore, the contact forces and energies stored in the three springs can be obtained as: 
\begin{equation}\label{eqn:contact_force}
    F_x = -K_t u, \quad F_y = -K_t w, \quad F_z = -K_n n,
\end{equation}
\begin{equation}\label{eqn:contact_energy}
    E_x = \frac{1}{2} K_t u^2, \quad E_y = \frac{1}{2} K_t w^2, \quad E_z = \frac{1}{2} K_n n^2.
\end{equation}
where $K_t$ and $K_n$ are the stiffness of the springs along tangential and normal directions. 

In order to analyse the impact mechanism, the normal impulse, denoted with $\Lambda_z$, is taken as the independent variable instead of time. We can represent the change rate of the normal impulse $\Lambda_z$ in terms of energy $E_z$ as
\begin{equation}
\dot \Lambda_z = \frac{d\Lambda_z}{dt} = F_z = \sqrt{2 K_n E_z},
\end{equation}
the derivatives of the other two tangential impulses and the normal energy with respect to $I_z$ can be derived as:
\textcolor{black}{
\begin{equation}\label{eqn:impact_derivatives}
        \Lambda_x' = \frac{\dot \Lambda_x}{\dot \Lambda_z} = -\sigma_x \sqrt{\frac{2K_t E_x}{2 K_n E_z}} = -\frac{\sigma_x}{\eta}\sqrt{\frac{E_x}{E_z}},
\end{equation}
\begin{equation}\label{eqn:impact_derivatives}
        \Lambda_y' = \frac{\dot \Lambda_y}{\dot \Lambda_z} = -\sigma_y \sqrt{\frac{2K_t E_y}{2 K_n E_z}} = -\frac{\sigma_y}{\eta} \sqrt{\frac{E_y}{E_z}} ,
\end{equation}
\begin{equation}\label{eqn:impact_derivatives}
        E_z' = \frac{dE_z}{d\Lambda_z} = \frac{\dot E_z}{\dot \Lambda_z} = \frac{kn \dot n}{-kn} = -\dot n = -v_z,
\end{equation}
where $\sigma_x$ and $\sigma_y$ are set to 1 if the springs are extended and are set to -1 if they are compressed; $\eta  = \eta_0$ during the compression stage and $\eta = {\eta_0}/{e}$ during the restitution stage, where $\eta_0^2 = {K_n}/{K_t}$ is the constant stiffness ratio and $e$ is the energetic coefficient of restitution.}

\subsubsection{Slipping and sticking contact modes}
Here, we describe the conditions that specify slipping and sticking contact modes and their transitions from one to the other. According to Coulomb's law, the contact between two bodies sticks if
\begin{equation}
    \label{eqn:coulomb}
    \sqrt{F_{x}^2 + F_{y}^2} < \mu F_{z}.
\end{equation}
From~\eqref{eqn:contact_force},~\eqref{eqn:contact_energy} and~\eqref{eqn:coulomb} we obtain the following
% .. which is equivalent to the following equation by~(\ref{eqn:contact_force_energy}) \theo{what do we mean here? what is by ?}
\begin{equation}
    E_{x} + E_{y} < \mu^2 \eta^2 E_{z}.
\end{equation}
Therefore, if a contact is in sticking mode, and $E_{x} + E_{y} = \mu^2 \eta^2 E_{z}$, the contact mode will switch to slipping mode. Otherwise, if contact is in slipping mode and $\bm v_t - \dot u \hat{\bm x} - \dot w \hat{\bm y} = 0$, the contact mode will switch to sticking mode. 

\subsubsection{Impulse evolution along normal and tangential directions}
Given the current contact mode, we can determine the change rates of the tangential springs, based on\textcolor{black}{
\begin{align}
\label{eqn:u_dot}
\dot u = \left\{\begin{array}{ll} 
        v_x,\hfill \small \text{stick},\\
        \frac{\sigma_x\mu^2\eta^3E_z'\sqrt{E_zE_x}+ v_xE_y - \sigma_x \sigma_y v_y \sqrt{E_xE_y}}{\mu^2\eta^2E_z}, \quad \hfill \small \text{slip},
        \end{array} \right.
\end{align}
\begin{align}
\label{eqn:w_dot}
\dot w = \left\{\begin{array}{ll} 
        v_y,\hfill  \small \text{stick},\\
        \frac{\sigma_y\mu^2\eta^3E_z'\sqrt{E_zE_y}+ v_yE_x - \sigma_x \sigma_y v_x \sqrt{E_xE_y}}{\mu^2\eta^2E_z}, \  \small \text{slip}.
        \end{array} \right.
\end{align}}
Further, the energy stored in the three springs is obtained by
\begin{equation}
    E_x = \frac{G_x^2}{4\eta_0^2}; \ E_y = \frac{G_y^2}{4\eta_0^2}; \ \small \text{and}\ E_z = E_z + E_z'\Delta \Lambda_z,
\end{equation}
\textcolor{black}{where $G_x$ and $G_y$ are used to track the changes in length of the two tangential springs whose initial values are 0}, and their derivatives with respect to the normal impulse $\Lambda_z$ are 
\begin{align}
\label{eqn:u_dot}
 \begin{bmatrix} G_x' \\ G_y'\end{bmatrix}=\left\{\begin{array}{ll} 
        G_x' = \frac{\dot u}{\sqrt{E_z}},  G_y' = \frac{\dot w}{\sqrt{E_z}}, \ \text{if compression,}\\
        G_x' = e\frac{\dot u}{\sqrt{E_z}}, G_y' = e\frac{\dot w}{\sqrt{E_z}}, \ \text{if restitution}. 
        \end{array} \right.
\end{align}

According to~\eqref{eqn:impact_decompositon_impulse} and~\eqref{eqn:impact_derivatives}, the impulse along normal and tangential directions are updated in $(i+1)^{th}$ iteration as: 
\begin{equation}\label{eqn:impulse_iteration}
    \bm \Lambda_{i+1} = \bm \Lambda_{i} + (\Lambda_x' \hat{\bm x} + \Lambda_y' \hat{\bm y} + \hat{\bm z}) \Delta \Lambda_z
\end{equation}

\subsubsection{Impact state transition}
As shown in Fig.~\ref{fig:impact_model}, the impact starts with compression phase, then it will switch to restitution stage when the normal contact velocity decrease to zero and the strain energy $E_n$ reaches its maximum value. During the restitution stage, the strain energy will decease to zero till the impact ends. At the end of the impact the velocity of the object can be updated according to~\eqref{eqn:impact_dynamics_linear} and \eqref{eqn:impact_dynamics_angular} as:
% \begin{equation}\label{eqn:linear_velocity_variation}
% \Delta \bm v = \frac{\bm \Lambda}{M}  ,
% \end{equation}
% \begin{equation}\label{eqn:angular_velocity_variation}
% \Delta \bm \omega= \bm I^{-1}\bm r^\times \bm \Lambda.
% \end{equation}

\begin{minipage}{.45\linewidth}
\begin{equation}\label{eqn:linear_velocity_variation}
    \Delta \bm v = \frac{\bm \Lambda}{M},
    \end{equation}
\end{minipage}%
\begin{minipage}{.5\linewidth}
\begin{equation}\label{eqn:angular_velocity_variation}
    \Delta \bm \omega= \bm I^{-1}\bm r^\times \bm \Lambda,
    \end{equation}
\end{minipage}
\vspace{1mm}
where the manipulator is assumed to be fixed (not moving) at the moment of making contact with the object. 
\vspace{-1mm}

\subsection{Indirect Force Control}\label{appendix:indirect_control}
\textcolor{black}{In indirect force control (IFC) schemes~\cite{lutscher2017hierarchical}}, the physical robot is coupled with a virtual robot through a virtual mechanical relationship, such as mass-spring-damping system. By modulating the set point of the virtual robot, the desired contact
force and interaction behavior of the physical robot can be achieved. Here, instead of using a configuration space IFC controller~\cite{lutscher2017hierarchical}, we design a Cartersian IFC controller. 
% However, instead of implementing IFC controller in configuration space~\cite{lutscher2017hierarchical}, here, we design a IFC controller in Cartersian space. 

Throughout our experiments, the two \textit{KUKA-iiwa} robots are position-controlled in Cartesian impedance mode, which means that we send the desired joint positions and the end-effector stiffness to the robot controller. In addition, the relationship between the interaction force and end-effector displacement can be written as:
\begin{equation}\label{eqn:Cartesian_stiffness}
    \bm F = \bm K \left( \bm x_{v} - \bm x \right) - \bm D \dot{\bm x}
\end{equation}
where $\bm K$ and $\bm D$ are respectively the stiffness and damping matrices of the end-effector of robot, $\bm x_{v}$ and $\bm x$ are respectively the end-effector positions of virtual robot and physical robot, $\bm F$ is the contact force at the end-effector.

As the penetration speed $\dot{\bm x}$ is relatively small when the robot's end-effector is in contact with the object (after the compression stage), based on~\eqref{eqn:Cartesian_stiffness} we can obtain the desired end-effector motion $\bm x_d$ from the following set-point generator:
\vspace{-1mm}
\textcolor{black}{
\begin{equation}\label{eqn:set_point_generator}
   \bm x_{d} = \bm x + \Delta \bm x = \bm x + \frac{\bm F}{\bm K}.
\end{equation}
In summary, we will first generate the nominal end-effector motion $\bm x$, the desired contact force $\bm F$ and end-effector stiffness $\bm K$ from our MMTO algorithm. Then we use the set-point generator~\eqref{eqn:set_point_generator} to transform the contact force $\bm F$ to the penetration depth $\Delta \bm x$ of the robot end-effector.}
% Correspondingly, the desired joint position command for each robot can be computed through inverse kinematics. 
% With all these above elements, 
In this way, we can track the desired force trajectory with specific end-effector stiffness based on joint position control with Cartesian impedance mode.  \vspace{-2mm}

% \subsection{Details of single contact searching QP}
% \label{app:QP-relevant}

% \subsubsection{Normal smoothing}
% As shown in Fig.~\ref{fig:contact_searching} 1), in order to search the optimal contact location across different surfaces, the normal vector on the surfaces of the object is smoothed~\cite{murooka2020optimization} according to the following equation:
% \begin{equation}
%     \bm n_c^k = normalize \left( \frac{\bm n_{\mathcal{S}_c}+\sum_{\mathcal{S}_{\bar c} \in \mathcal{S}_{adj}} \omega_{\mathcal{S}_{\bar c}} \bm n_{\mathcal{S}_{\bar c}}}{1+ \sum_{\mathcal{S}_{\bar c} \in \mathcal{S}_{adj}} \omega_{\mathcal{S}_{\bar c}} }\right)
% \end{equation}
% where $n_{\mathcal{S}_c}$ is the normal vector of the contact surface $\mathcal{S}_c$; $n_{\mathcal{S}_{\bar c}}$ is the normal vector of the adjacent contact surface $\mathcal{S}_{\bar c}$, while $\omega_{\mathcal{S}_{\bar c}} \in [0,1]$ is the corresponding weight for adjacent contact surface which is calculated as follows:
% \begin{equation}
%     \omega_{\mathcal{S}_{\bar c}} = \left\{ \begin{array}{cc}
%         -2(\frac{d_{\mathcal{S}_{\bar c}}}{R})^2 +1 , & 0 \leq d_{\mathcal{S}_{\bar c}} \leq \frac{R}{2};\\
%         2(\frac{d_{\mathcal{S}_{\bar c}}}{R}-1)^2, & \frac{R}{2} \leq d_{\mathcal{S}_{\bar c}} \leq R.
%     \end{array}\right.
% \end{equation}
% where $d_{\mathcal{S}_{\bar c}}$ is the distance from contact location to the adjacent contact surface $\mathcal{S}_{\bar c}$; $R$ is the maximum distance within which the object surface will be considered as an adjacent contact surface.
\end{appendices}

% \section{A list of papers that we may cite, but don't know yet where}
% \input{sections/list_of_papers_4_reference.tex}

%\section*{ACKNOWLEDGMENT}

\bibliographystyle{IEEEtran}
\bibliography{reference}  % .bib



% \bibitem{ams}
% {\it{Mathematics into Type}}, American Mathematical Society. Online available: 

% \bibitem{oxford}
% T.W. Chaundy, P.R. Barrett and C. Batey, {\it{The Printing of Mathematics}}, Oxford University Press. London, 1954.

% \bibitem{lacomp}{\it{The \LaTeX Companion}}, by F. Mittelbach and M. Goossens

% \bibitem{mmt}{\it{More Math into LaTeX}}, by G. Gr\"atzer

% \bibitem{amstyle}{\it{AMS-StyleGuide-online.pdf,}} published by the American Mathematical Society

% \bibitem{Sira3}
% H. Sira-Ramirez. ``On the sliding mode control of nonlinear systems,'' \textit{Systems \& Control Letters}, vol. 19, pp. 303--312, 1992.

% \bibitem{Levant}
% A. Levant. ``Exact differentiation of signals with unbounded higher derivatives,''  in \textit{Proceedings of the 45th IEEE Conference on Decision and Control}, San Diego, California, USA, pp. 5585--5590, 2006.

% \bibitem{Cedric}
% M. Fliess, C. Join, and H. Sira-Ramirez. ``Non-linear estimation is easy,'' \textit{International Journal of Modelling, Identification and Control}, vol. 4, no. 1, pp. 12--27, 2008.

% \bibitem{Ortega}
% R. Ortega, A. Astolfi, G. Bastin, and H. Rodriguez. ``Stabilization of food-chain systems using a port-controlled Hamiltonian description,'' in \textit{Proceedings of the American Control Conference}, Chicago, Illinois, USA, pp. 2245--2249, 2000.

% \end{thebibliography}
\vspace{-5mm}

\begin{IEEEbiography}[{\includegraphics[width=1in,height=1.25in,clip,keepaspectratio]{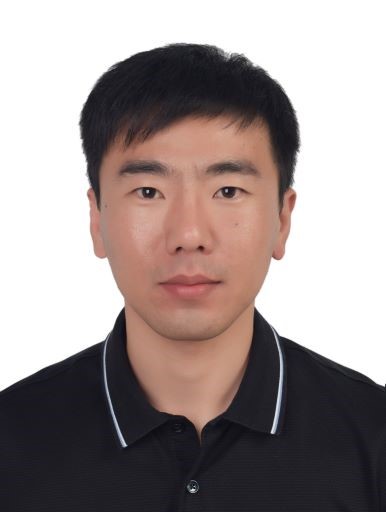}}]{Lei Yan} received his Ph.D. in Mechanical Engineering from Harbin Institute of Technology, China in 2019. From 2019 to 2021, he was a Research Associate with the School of Informatics, The University of Edinburgh, Edinburgh, UK.

He is currently an Assistant Professor with the School of Mechanical Engineering and Automation, Harbin Institute of Technology, Shenzhen, China. His research interests include dexterous dynamic manipulation, multi-robot collaboration and physical human-robot interaction. 
\end{IEEEbiography}\vspace{-5mm}
\begin{IEEEbiography}[{\includegraphics[width=1in,height=1.25in,clip,keepaspectratio]{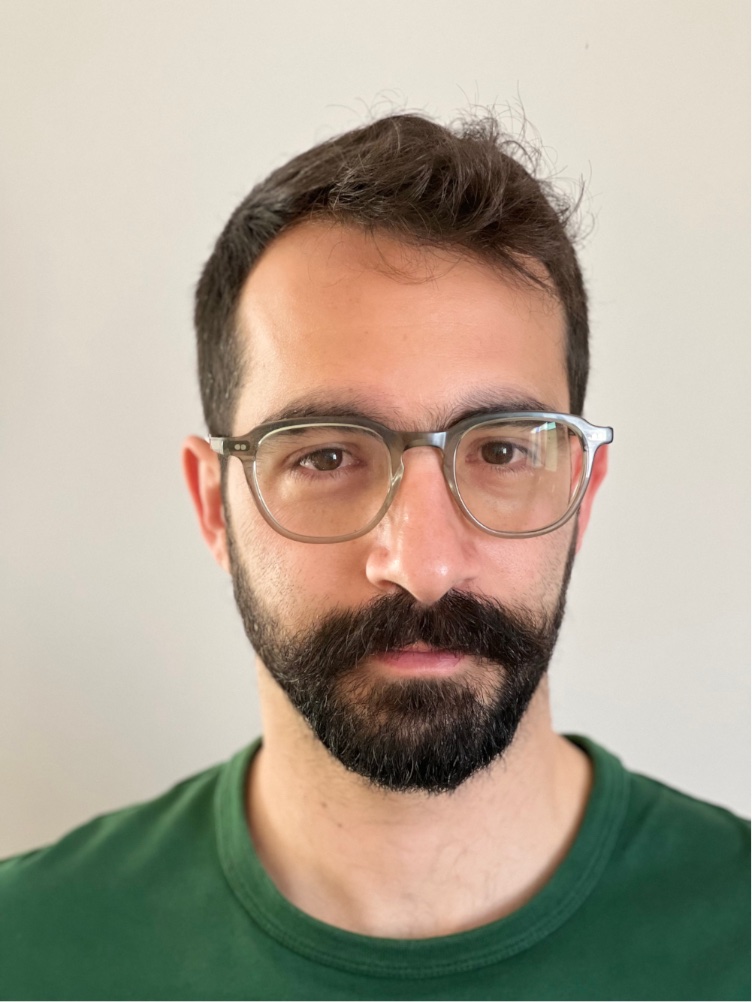}}]{Theodoros Stouraitis}
received his Ph.D. in Robotics from the University of Edinburgh in UK and in collaboration with the Honda Research Institute Europe in Germany. 
From 2012 to 2015, he was a research assistant at the Institute of Robotics and Mechatronics of the German Aerospace Center (DLR), in 2022, he was a Visiting Scientist at Massachusetts Institute of Technology (MIT), USA and from 2022 till 2023, he was a Guest Scientist at the Honda Research Institute Europe in Germany. 

He is currently a Control Scientist working on robotics and autonomous vehicles projects. His research interests include motion planning and control, non-linear optimization, contacts and human-robot interaction. 
\end{IEEEbiography}
\vspace{-5mm}
\begin{IEEEbiography}[{\includegraphics[width=1in,height=1.25in,clip,keepaspectratio]{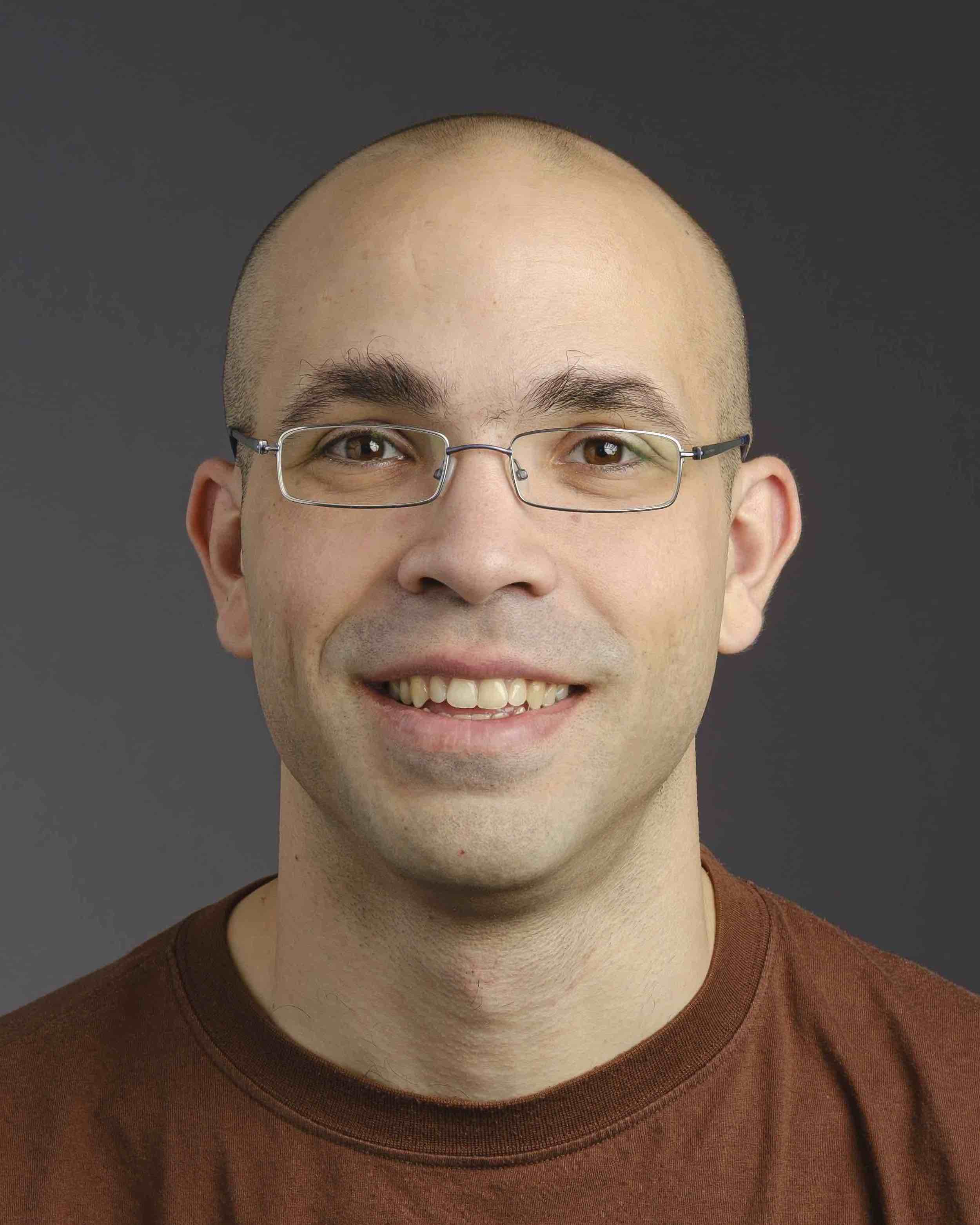}}]{Jo\~{a}o Moura}
received his Ph.D. in Robotics and Autonomous Systems jointly awarded by Heriot-Watt University and The University of Edinburgh, UK, in 2021.
He is currently a Research Associate in the School of Informatics at The University of Edinburgh, and affiliated with The Alan Turing Institute.
His research interests include contact-rich and non-prehensile manipulation, trajectory optimization, model predictive control, and imitation learning.
\end{IEEEbiography}
\vspace{-5mm}
\begin{IEEEbiography}[{\includegraphics[width=1in,height=1.25in,clip,keepaspectratio]{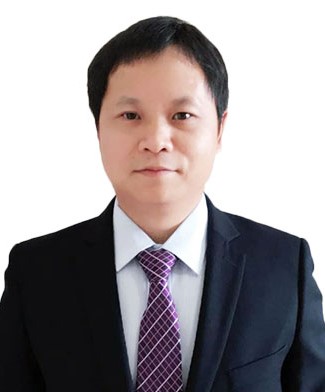}}]{Wenfu Xu} (Senior Member, IEEE) received the B.E. and M.E. degrees in Control Engineering from the Hefei University of Technology, Hefei, China, in 2001 and 2003, respectively, and the Ph.D. degrees in Control Science and Engineering from the Harbin Institute of Technology, Harbin, China, in 2007.

He was a Research Associate with the Department of Mechanical and Automation Engineering, The Chinese University of Hong Kong, Hong Kong, China. He is currently a Professor with the Department of Mechanical and Automation Engineering, Harbin Institute of Technology, Shenzhen, China. His research interests include space robotics, bionic robotics, and cable-driven manipulators. 
\end{IEEEbiography}
\vspace{-5mm}
\begin{IEEEbiography}[{\includegraphics[width=1in,height=1.25in,clip,keepaspectratio]{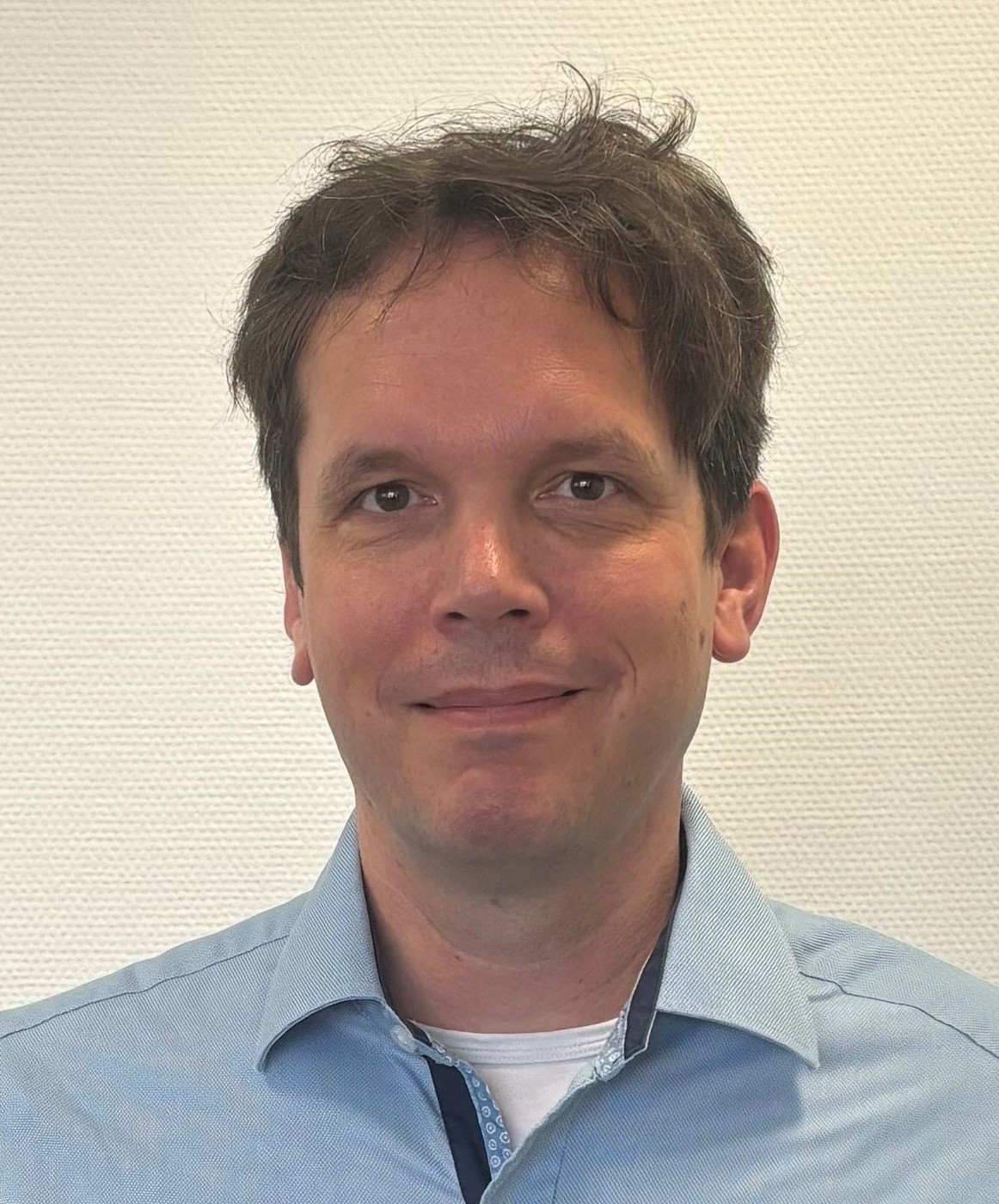}}]{Michael Gienger}
received the diploma degree in Mechanical Engineering from the Technical University of Munich, Germany, in 1998. From 1998 to 2003, he was a research assistant at the Institute of Applied Mechanics of the TUM and received his Ph.D. degree with a dissertation on “Design and Realization of a Biped Walking Robot”. After this, Michael Gienger joined the Honda Research Institute Europe in Germany in 2003. Currently he works as a Chief Scientist and Competence Group Leader in the field of robotics. His research interests include mechatronics, robotics, whole-body control, imitation learning, and human-robot interaction. 
\end{IEEEbiography}
\vspace{-5mm}
\begin{IEEEbiography}[{\includegraphics[width=1in,height=1.25in,clip,keepaspectratio]{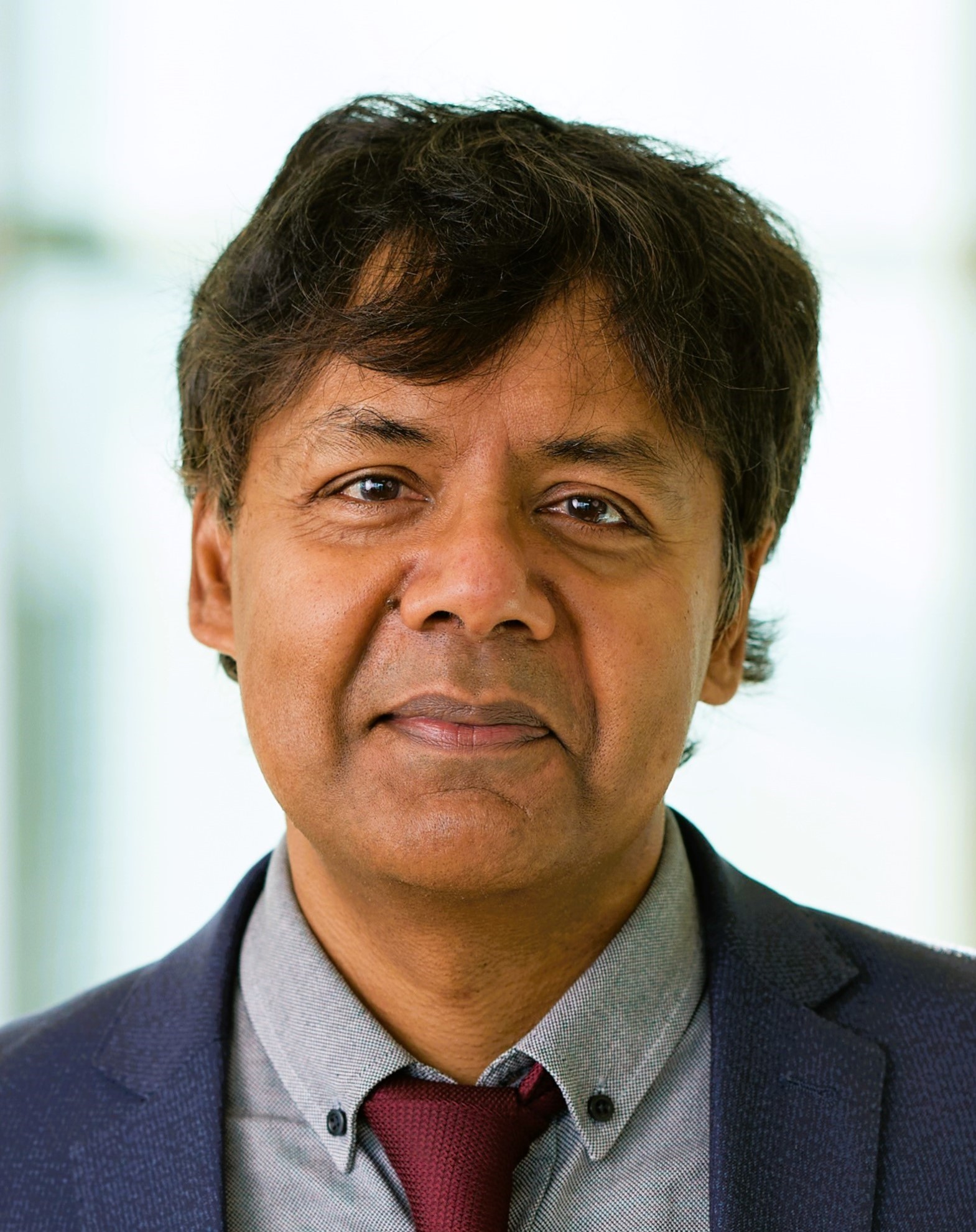}}]{Sethu Vijayakumar}
received his Ph.D. in Computer Science and Engineering from the Tokyo Institute of Technology, Japan in 1998. He is the Professor of Robotics at the University of Edinburgh, an adjunct faculty of the University of Southern California, Los Angeles and the founding Director of the Edinburgh Centre for Robotics. His research interests include statistical machine learning, anthropomorphic robotics, multi objective optimisation and optimal control in autonomous systems as well as the study of human motor control. %Professor Vijayakumar helps shape and drive the national Robotics and Autonomous Systems (RAS) agenda in his recent role as
He is the Programme co-Director for Artificial Intelligence (AI) at The Alan Turing Institute, the UK’s national institute for data science and AI. He is a Fellow of the Royal Society of Edinburgh,  and winner of the 2015 Tam Dalyell Prize for excellence in engaging the public with science.
\end{IEEEbiography}
% \balance
\end{document}